\documentclass[11pt]{article}

\usepackage[colorlinks=true,
            linkcolor=red,
            urlcolor=blue,
            citecolor=blue,
            breaklinks=true]{hyperref}
\usepackage{breakurl} 
\usepackage[a4paper, left=3cm, right=3cm, top=3cm, bottom=3cm]{geometry}
\RequirePackage{amsmath,amsfonts,amssymb}
\usepackage{mathtools}
\usepackage{mathrsfs}
\usepackage{dsfont}
\usepackage{natbib}
\usepackage{setspace}
\usepackage{graphicx}
\usepackage{subfigure}
\usepackage[export]{adjustbox}
\usepackage{booktabs}
\usepackage{algorithm}
\usepackage{algorithmic}
\usepackage[title]{appendix}

\hypersetup{
  pdfauthor = {Simon Bussy, Mokhtar Z. Alaya, Agathe Guilloux},
  pdftitle = {},
  pdfcreator = {pdflatex},
  pdfproducer = {pdflatex}
}

\newtheorem{theorem}{Theorem}

\newtheorem{lemma}{Lemma}

\newtheorem{assumption}{Assumption}{\bf}{\rm}

\DeclareMathOperator{\argmin}{argmin}

\DeclareMathOperator{\prox}{{prox}}
\DeclareMathOperator{\sgn}{sign}

\DeclareMathOperator{\diag}{diag}

\DeclareMathOperator{\TV}{TV}

\DeclareMathOperator{\bina}{bina}

\newcommand{\XB}{{\boldsymbol X}^B}%
\newcommand{\bigO}{\mathcal{O}}
\newcommand{\bP}{\mathbb P}
\newcommand{\bX}{\boldsymbol X}%
\newcommand{\dd}{\mathrm{d}}

\newcommand{\ind}[1]{\mathds{1}_{#1}}
\newcommand{\norm}[1]{\|#1\|}
\newcommand{\R}{{\mathbb{R}}}
\newcommand{\N}{\mathbb N}
\newcommand{\cC}{\mathcal C}
\newcommand{\cP}{\mathcal P}
\newcommand{\cU}{\mathcal U}
\newcommand{\cN}{\mathcal N}
\newcommand{\cG}{\mathcal G}
\newcommand{\cS}{\mathcal S}
\newcommand{\cH}{\mathcal H}
\newcommand{\cM}{\mathcal M}

\newcommand{\scrL}{\mathscr L}

\newcommand{\E}{\mathds{E}}

\renewcommand{\P}{\mathds{P}}
\newcommand{\bSigma}{\textbf{$\Sigma$}}

\date{}

\begin{document}

\title{\vspace{-1cm} Binacox: automatic cut-point detection in high-dimensional Cox model with applications in genetics \vspace{-1.5cm}}

\maketitle

\begin{center}
{\setstretch{1.2}

\large Simon Bussy \\
\normalsize
LPSM, UMR 8001, CNRS, Sorbonne University, Paris, France\\
\emph{email}: \texttt{simon.bussy@gmail.com}\\

\vspace{0.2cm}

\large Mokhtar Z. Alaya \\ 
\normalsize
Modal'X, UPL, Univ Paris Nanterre, F92000 Nanterre, France\\
\emph{email}: \texttt{mokhtarzahdi.alaya@gmail.com}\\

\vspace{0.2cm}

\large Anne-Sophie Jannot \\
\normalsize
Biomedical Informatics and Public Health Department, EGPH, APHP \\
and INSERM, UMRS 1138, Centre de Recherche des Cordeliers, Paris, France \\ 
\emph{email}: \texttt{annesophie.jannot@aphp.fr}\\

\vspace{0.2cm}

\large Agathe Guilloux \\
\normalsize
LaMME, UEVE and UMR 8071, Paris Saclay University, Evry, France \\
\emph{email}: \texttt{agathe.guilloux@math.cnrs.fr}\\
}
\vspace{.8cm}
\end{center}

\begin{abstract}
We introduce the \textit{binacox}, a prognostic method to deal with the problem of detecting multiple cut-points per features in a multivariate setting where a large number of continuous features are available.
The method is based on the Cox model and combines one-hot encoding with the \emph{binarsity} penalty, which uses total-variation regularization together with an extra linear constraint, and enables feature selection. Original nonasymptotic oracle inequalities for prediction (in terms of Kullback-Leibler divergence) and estimation with a fast rate of convergence are established.
The statistical performance of the method is examined in an extensive Monte Carlo simulation study, and then illustrated on three publicly available genetic cancer datasets.
On these high-dimensional datasets, our proposed method significantly outperforms state-of-the-art survival models regarding risk prediction in terms of the C-index, with a computing time orders of magnitude faster. In addition, it provides powerful interpretability from a clinical perspective by automatically pinpointing significant cut-points in relevant variables.\\

\emph{Keywords.} Cox model; Cut-point; Feature binarization; Nonasymptotic oracle inequality; Proximal methods; Survival analysis; Total variation
\end{abstract}

\section{Introduction}

Determining significant prognostic biomarkers is of increasing importance in many areas of medicine.
Scores used in clinical practice often categorize continuous features into binary ones using expert-driven cut-points. 
For instance, the Wells score, which categorizes patients into low, moderate and high risk groups for pulmonary embolism~\citep{wells2000derivation}, is one of the most extensively validated predictive scores. One of the categorized feature used in this score is ``having a heart rate of over 100 beats per minute, or not''. 
When used in routine care, this type of threshold makes a score more interpretable from a clinical point of view. In this particular example, it means that experts consider that heart rate has a nonlinear effect: there is reasonable agreement that above this threshold, patients have higher risk of unfavourable outcome. Despite this choice of threshold, there is little agreement on the exact nature of the relationship between heart rate and prognosis. 

With the increasing availability of high-dimensional datasets, data-driven predictive scores are becoming increasingly important, e.g., in genetic oncology studies, where similar questions occur because the effect of certain genes' expression on survival times are often non-linear. 
Therefore, to develop such scores, one has to deal with a two-sided problem: first to select relevant features, and second to find relevant thresholds -- also called \emph{cut-off} values or \emph{cut-points} -- for these selected continuous features, without prior or expert knowledge. 

\paragraph{The cut-point detection problem.}

Solving this problem means applying non-linearities to feature effects that most models cannot detect. This also offers the ability to classify patients into several groups in terms of their continuous feature values relative to the cut-points. More importantly, this can also lead to a better understanding of the features' effects on the outcome of interest; this strategy might uncover biological thresholds as well as potential criteria for new prospective studies, help diagnose diseases, and make treatment recommendations. A convenient tool for finding optimal cut-points is therefore of high interest.

Indeed, good cut-point detection is a common issue in medical studies, and numerous methods have been proposed for determining a single cut-point for a given feature. 
This ranges from choosing the mean or median, to methods based on distribution of values, or association with clinical outcomes, e.g., the minimal $p$-value from multiple log-rank tests, see~\citet{camp2004x, moul2007age, rota2015optimal} among many others.
However, the choice of the actual cut-points is not a straightforward problem, even for a single cut-point~\citep{lausen1992maximally, klein2003discretizing,contal1999application}. Recently,~\citet{icuma2018determination} proposed a Bayesian approach with accelerated failure time modeling, but still only allowing one cut-point per feature.

Indeed, while many studies have been devoted to find one optimal cut-point, there is often need in medical settings to determine not only one but multiple cut-points. 
For instance, prognoses are generally worst at both ends of the body mass index, i.e., for obese and underweight individuals~\citep{oreopoulos2008body}.
Methods exist to deal with multiple cut-point detection for one-dimensional signals (see for instance~\citet{BleVer-11} and~\citet{HarLev-10} that use a group fused lasso or total-variation penalty, respectively), and for multivariate time series (see~\citet{cho2015multiple}).
Though cut-point detection is also a paramount issue in survival analysis~\citep{faraggi1996simulation}, methods that have been developed in this setting only look at a single feature at a time (e.g.,~\citet{motzer1999survival}~and~\citet{leblanc1993survival} which use  survival trees, or more recently~\citet{chang2019methods}).
To our knowledge, a multivariate survival analysis method well-suited to detect multiple cut-points per feature in a high-dimensional setting has not been previously proposed. 

\paragraph{General framework.}

Let us consider the usual survival analysis framework.
Following~\citet{andersen2012statistical}, let non-negative random variables $T$ and $C$ stand for the time of the event of interest and censoring time respectively, and $X$ denote the $p$-dimensional vector of features (e.g., patient characteristics, therapeutic strategy, omics features). The event of interest could be for instance survival time, re-hospitalization, relapse or disease progression.
Conditionally on $X$, $T$ and $C$ are assumed to be independent, which is classical in survival analysis \citep{klein2005survival}. 
We then denote $Z$ the right-censored time and $\Delta$ the censoring indicator, defined as 
\begin{equation*}
Z = T \wedge C \quad \text{and} \quad \Delta = \ind{}({T \leq C})
\end{equation*}
respectively, where $a \wedge b$ denotes the minimum between two numbers $a$ and $b$, and $\ind{}(\cdot)$ the indicator function taking the value $1$ if the condition in $(\cdot)$ is satisfied and $0$ otherwise.

The Cox proportional hazards model~\citep{Cox1972JRSS} is by far the most widely used in survival analysis.
It describes the relation between the hazard function and the features by 
\begin{equation*}
\lambda(t | X=x) = \lambda_0(t) e^{x^\top \beta^\text{cox}},
\end{equation*}
where $\lambda_0$ is a baseline hazard function describing how the event risk changes over time at baseline levels of features, and $\beta^\text{cox} \in \R^p$ a vector quantifying the multiplicative impact on the hazard ratio of each feature. 

\paragraph{High-dimensional survival analysis.}

High-dimensional settings are becoming increasingly frequent, in particular for genetic data applications where cut-point estimation is a common problem (see for instance~\citet{harvey1999estrogen,shirota2001ercc1, cheang2009ki67}), but also in other contexts where the number of available features to consider as potential risk factors is tremendous, particularly with the development of electronic health records.
A penalized version of the Cox model well-suited for such settings is proposed in~\citet{simon2011regularization}, but it cannot model nonlinearity. Theory for using lasso-type methods in the Cox model was developed in~\cite{huang2013}. Other methods have been put forward to deal with this problem in similar settings, like boosting Cox models~\citep{li2005boosting} and random survival forests~\citep{ishwaran2008random}. However, none of these identify cut-point values, which is of major interest for both interpretation and clinical benefit.

\paragraph{Main contribution.}

In this paper, we propose a method called \textit{binacox} that estimates multiple cut-points in a Cox model with high-dimensional features. 
First, the binacox one-hot encodes the continuous input features~\citep{WuCog2012} through a mapping to a new binarized space of much higher dimension, and then trains the Cox model in this space, regularized with the \emph{binarsity} penalty~\citep{alaya2016} which combines total-variation regularization with an extra sum-to-zero constraint, and enables feature selection. Cut-points of the initial continuous input features are then detected by the jumps in the regression coefficient vectors, which the binarsity penalty forces to be piecewise-constant.
The main contribution of this paper is twofold. First we introduce the idea of using a total-variation penalty with an extra linear constraint on the weights of a Cox model trained on a binarization of the raw continuous features. This leads to a procedure that automatically detects relevant features and allows multiple cut-points per feature. Secondly the oracle inequality in prediction of Section~\ref{sec:theoretical guarantees} (see Theorem~\ref{thm-faster-in-approx}) is stated in terms of Kullback-Leibler divergence, as opposed to the results in~\cite{huang2013} (for the lasso penalty) expressed in Breiman divergence, the arguments are consequently different.

\paragraph{Organization of the paper.}
 
A precise description of the model is given in Section~\ref{sec:model}. Section~\ref{sec:theoretical guarantees} highlights the good theoretical properties of the binacox by establishing fast oracle inequalities for prediction and for estimation. 
Section~\ref{sec:Performance evaluation} presents the simulation procedure used to evaluate the performance of our method and compares it with existing ones.
In Section~\ref{sec:application}, we apply our method to high-dimensional genetic datasets. Finally, we discuss the obtained results in Section~\ref{sec:conclusion}.

\paragraph*{Notation.}
Throughout the paper, for every $q > 0,$ we denote by $\norm{v}_q$ the usual $\ell_q$-quasi norm of a vector $v \in \R^m,$ namely $\norm{v}_q =(\sum_{k=1}^m|v_k|^q)^{1/q}$,  and $\norm{v}_\infty = \max_{1 \leq k \leq m}|v_k|$. 
We write $\mathbf{1}$ (resp. $\mathbf{0}$) the vector having all coordinates equal to one (resp. zero).
We also denote $|A|$ the cardinality of a finite set $A$.
If $I$ is an interval, $|I|$ stands for its Lebesgue measure.
Then, for any $u \in \R^m$ and any $L \subset \{1, \ldots, m\},$ we denote $u_L$ the vector of $\R^m$ satisfying $(u_L)_k = u_k$ for $k \in L$ and $(u_L)_k = 0$ for $ k \in L^\complement :=  \{1, \ldots, m\}\backslash L$. 
Finally,  for a matrix $M$ of size $k \times k^{'}$, $M_{j,\bullet}$ denotes its $j$th row and $M_{\bullet,l}$ its $l$th column.

\section{Model and method}
\label{sec:model}

\subsection{Cox model with cut-points.}
Consider an independent and identically distributed (i.i.d.) sample 
\begin{equation*}
(X_1,Z_1,\Delta_1), \dots , (X_n,Z_n,\Delta_n) \in [0, 1]^p \times \R_+ \times \{0,1\},
\end{equation*}
where the condition $X_i \in [0, 1]^p$ for all $i = 1, \ldots, n$ is always true after an appropriate rescaling preprocessing step, without loss of generality.
Let $\bX = [X_{i,j}]_{1 \leq i \leq n; 1 \leq j \leq p}$ be the $n \times p$ design matrix vertically stacking the $n$ samples of $p$ raw features so that $\bX_{i, \bullet} = X_i$.
In order to simplify the presentation of our results, we assume in the paper that the raw features $\bX_{\bullet,j}$ are continuous for all $j = 1, \ldots, p$, but this is not a limitation in practice. 
Assume that the hazard function for patient $i$ is given by 
\begin{equation*}
\lambda^\star (t|X_{i}) = \lambda_0^\star(t) e^{f^\star(X_{i})},
\end{equation*}
where $\lambda_0^\star(t)$ is the baseline hazard function, and 
\begin{equation}
\label{f-star}
  f^\star(X_i)= \sum_{j=1}^p f^\star_j(X_{i,j}) = \sum_{j=1}^p\sum_{k=1}^{K_j^\star+1} \beta^\star_{j,k} \ind{}(X_{i,j}\in I^\star_{j,k}),
\end{equation}
with $I^\star_{j,k} = (\mu^\star_{j,k-1}, \mu_{j,k}^\star]$ for $k = 1, \ldots, K^\star_j+1$ and where $\beta^\star_{j,k} \neq \beta^\star_{j,k+1}$ for $k = 1, \ldots, K^\star_j$.
We impose that
\begin{equation*}
  \sum_{i=1}^n f^\star_j(X_{i,j}) = 0 \quad \text{ for all } j = 1, \ldots, p
\end{equation*}
to ensure identifiability (see~\citep{meier2009high} for a similar constraint in generalized additive models), which can also be written as a sum-to-zero constraint in each $\beta^\star$'s block, that is:
\begin{equation}
\label{sum-to-zero-constraint}
   \sum_{k=1}^{K_j^\star+1} \beta^\star_{j,k} n^\star_{j,k} = 0 \quad \text{ for all } j = 1, \ldots, p
\end{equation} 
where $n^\star_{j,k} = |\{i=1, \ldots, n: X_{i,j} \in I^\star_{j,k}\}|$.
For each feature $j = 1, \ldots, p$, the $\mu^\star_{j, k}$s ($k = 1, \ldots, K^\star_{j}$) are the so-called cut-points, and are such that
\begin{equation*}
\mu^\star_{j, 1} < \mu^\star_{j, 2} < \cdots < \mu^\star_{j,K^\star_{j}},
\end{equation*}  
with the conventions $\mu^\star_{j,0} = 0$ and $\mu^\star_{j,K^\star_j +1} = 1$.
Denoting $K^\star = \sum_{j=1}^p K^\star_j$, the vector of regression coefficients $\beta^\star \in \R^{K^\star + p}$ is given by 
\begin{align*}
 \beta^\star &= ({\beta^\star_{1, \bullet}}^\top, \ldots, {\beta^\star_{p, \bullet}}^\top)^\top = (\beta^\star_{1, 1},\ldots, \beta^\star_{1, K^\star_1 +1}, \ldots, \beta^\star_{p, 1}, \ldots,\beta^\star_{p, K_p^\star+1})^\top,
\end{align*} 
and the cut-points vector $\mu^\star \in \R^{K^\star}$ by
 \begin{align*}
 \mu^\star &= ({\mu^\star_{1, \bullet}}^\top, \ldots, {\mu^\star_{p, \bullet}}^\top)^\top = (\mu^\star_{1,1}, \ldots, \mu^\star_{1,K^\star_1}, \ldots, \mu^\star_{p,1}, \ldots, \mu^\star_{p,K^\star_p})^\top.
 \end{align*}
Our goal is to  simultaneously estimate $\mu^\star$ and $\beta^\star$, which also requires estimation of the unknown $K^\star_j$ for all $j = 1, \ldots, p$. Towards this end, the first step of our proposed method is to map the feature space to a much higher space of binarized features.

\subsection{Binarization.}

Let $\XB$ be the sparse binarized matrix with an extended number $p+d$ of columns, typically with $d \gg p$, where continuous input features have been one-hot encoded~\citep{WuCog2012,LiuHussTanDas2002}. The $j$th column $\bX_{\bullet,j}$ is then replaced by $d_j+1 \geq 2$ columns $\bX^B_{\bullet,j, 1}, \ldots, \bX^B_{\bullet,j, d_j+1}$ containing only zeros and ones, where the $i$th row $X_i^B \in \R^{p+d}$ with $d = \sum_{j=1}^p d_j$ is written 
\begin{equation*}
X_i^B = (X^B_{i,1,1}, \ldots, X^B_{i,1,d_1+1}, \ldots, X^B_{i,p,1}, \ldots, X^B_{i,p, d_p+1})^\top.
\end{equation*}
We consider a partition of intervals $I_{j, 1}, \ldots, I_{j, d_j+1}$ 
such that 
\begin{equation*}
\bigcup_{k=1}^{d_j+1}I_{j,k} = [0, 1]
\end{equation*}
and $I_{j,k} \cup I_{j,k'} = \varnothing$ for all $k\neq k'$ with $k, k'= 1, \ldots, d_j+1$. 
Now for $i = 1, \ldots, n$ and $l = 1, \ldots, d_j+1$, we define
\begin{equation*}
  X_{i, j, l}^B =
  \begin{cases}
    1 &\text{ if } X_{i,j} \in I_{j, l}, \\
    0 & \text{ otherwise}.
  \end{cases}
\end{equation*}
We then denote $I_{j,l} = (\mu_{j,l-1}, \mu_{j, l}]$ for $l =1, \ldots, d_j + 1$, with the convention $\mu_{j,0} = 0$ and $\mu_{j,d_j +1} = 1$. A natural choice for the $\mu_{j, l}$ is given by the quantiles, namely $\mu_{j, l}=q_j\big(l / (d_j+1)\big)$, where $q_j(\alpha)$ denotes a quantile of order $\alpha \in [0, 1]$ for $\bX_{\bullet, j}$.
If training data also contains  unordered qualitative features, one-hot encoding with $\ell_1$-penalization can be used, for instance.

To each binarized feature $\bX^B_{\bullet,j, l}$ corresponds a parameter $\beta_{j, l}$, and the vectors associated with the binarization of the $j$th feature are naturally denoted $\beta_{j, \bullet} = (\beta_{j, 1}, \ldots, \beta_{j, d_j+1})^\top$ and $\mu_{j, \bullet} = (\mu_{j, 1}, \ldots, \mu_{j, d_j})^\top$.
Hence, we define a candidate for the estimation of $f^\star$ defined in~\eqref{f-star} as
\begin{equation}
\label{f-beta}
f_\beta(X_i) = \beta^\top{X_i^B} = \sum_{j=1}^p f_{\beta_{j,\bullet}}(X_{i,j})=\sum_{j=1}^p \sum_{l=1}^{d_j +1} \beta_{j,l} \ind{}( X_{i,j} \in I_{j,l}).
\end{equation}
The full parameter vectors of size $p+d$ and $d$ respectively are finally obtained by concatenation of the vectors $\beta_{j, \bullet}$ and $\mu_{j, \bullet}$, i.e.,
\begin{align*}
\beta &= (\beta_{1, \bullet}^\top, \ldots, \beta_{p,
\bullet}^\top)^\top = (\beta_{1,1}, \ldots, \beta_{1,d_1+1}, \ldots, \beta_{p,1}, \ldots, \beta_{p, d_p+1})^\top,
\end{align*}
and 
\begin{align*}
  \mu &= (\mu_{1, \bullet}^\top, \ldots, \mu_{p,\bullet}^\top)^\top = (\mu_{1,1}, \ldots, \mu_{1,d_1}, \ldots, \mu_{p,1}, \ldots,\mu_{p, d_p})^\top.
\end{align*}

\subsection{Estimation procedure.}

In the following, for a fixed vector $\mu$ of quantization, we define the binarized partial negative log-likelihood (rescaled by $1/n$) as follows:
\begin{align}
\label{binarized-model}
\ell_n(f_\beta) = - \frac{1}{n} \sum_{i=1}^n \Delta_i \Big\{f_\beta(X_i) 
 - \log\sum_{\substack{i':Z_{i'} \geq Z_i}} e^{f_\beta(X_{i'})}
\Big\}.
\end{align}
Our approach consists in minimizing the function $\ell_n$ plus the binarsity penalization term introduced in~\citet{alaya2016}. The resulting optimization problem is written
\begin{equation}
\label{problem-estim}
\hat \beta \in \argmin_{\beta \in \mathscr{B}_{p+d}(R)} \big\{ \ell_n(f_\beta) + \bina(\beta)\big\},
\end{equation}
where $\mathscr{B}_{p+d}(R) =\{\beta \in \R^{p+d}: \sum_{j=1}^p\norm{\beta_{j,\bullet}}_\infty \leq R\}$ and
\begin{equation}
  \label{eqn:binarsity}
   \bina(\beta) = \sum_{j=1}^p \Big( \sum_{l=2}^{d_j+1} \omega_{j,l}| 
    \beta_{j, l} -\beta_{j, {l-1}}| + \delta_j(\beta_{j, \bullet}) \Big),
\end{equation}
with 
\begin{equation*}
  \delta_j(u) = 
  \begin{cases}
    0 \quad &\text{ if } \quad n^\top_{j,\bullet} u = 0, \\
    \infty &\text{ otherwise,}
  \end{cases}
\end{equation*}
and where $n_{j,\bullet} = (n_{j,1}, \ldots, n_{j,d_j+1})^\top \in \mathbb{N}^{d_j+1}$ with $n_{j,l} = |\{i=1, \ldots, n: X_{i,j} \in I_{j,l}\}|$ for all $j=1, \ldots, p$ and $l=1, \ldots, d_j+1.$
The constraint over $\mathscr{B}_{p+d}(R)$ is standard in the literature for obtaining proofs of oracle inequalities for sparse generalized linear models~\citep{van2008high}, and is discussed in detail below.
The weights $\omega_{j,l}$ are of order
\begin{equation*}
\omega_{j,l} = \bigO\bigg(\sqrt{\frac{\log(p+d)}{n}}\bigg),
\end{equation*}
see Appendix~\ref{preliminaries-proofs} for their explicit form.

It turns out that the binarsity penalty is well-suited to our problem. First, it tackles the problem that $\XB$ is not full rank by construction, since $\sum_{l=1}^{d_j+1} X_{i, j, l}^B = 1$ for all $j = 1, \ldots, p$, which means that the columns in each block sum to  $\mathbf 1$. This problem is solved since the penalty imposes the linear constraint $\sum_{l=1}^{d_j+1} n_{j,l}\beta_{j,l} = 0$ in each block with the $\delta_j(\cdot)$ term. 
Note that if the $I_{j,l}$ are taken as the interquantiles intervals, we have that $n_{j,l}$ are all equal for $l=1, \ldots, d_j+1$, and we get the standard sum-to-zero constraint $\sum_{l=1}^{d_j+1}\beta_{j,l} = 0$. Then, the other term in the penalty consists of a within-block weighted total variation penalty: 
\begin{equation}
  \label{eq:tv_no_linear_constraint}
  \norm{\beta_{j,\bullet}}_{\TV,\omega_{j,\bullet}} = \sum_{l=2}^{d_j+1} \omega_{j,l} |\beta_{j, l} -  \beta_{j, l-1}|,
\end{equation}
that takes advantage of the fact that within each block, binarized features are ordered. The effect is then to keep the number of different values taken by $\beta_{j, \bullet}$ to a minimum, which makes significant cut-points appear, as detailed hereafter.

For all $\beta \in \R^{p+d},$ let $\mathcal{A}(\beta)=  \big[\mathcal{A}_1 (\beta), \ldots, \mathcal{A}_ p (\beta)\big]$ be the concatenation of the support sets relative to the total-variation penalization, namely 
\begin{equation*}
\mathcal{A}_j(\beta)= \big\{l: \beta_{j,l} \neq \beta_{j,l-1}, \textrm{ for } l =2, \ldots, d_j+1\big\}
\end{equation*} 
for all $j = 1, \ldots, p$.
Similarly, we denote $\mathcal{A}^\complement(\beta)= \big[\mathcal{A}_1^\complement(\beta), \ldots, \mathcal{A}_p^\complement(\beta)\big]$ the complementary set of $\mathcal{A}(\beta).$ 
We then write 
\begin{equation}
\label{active-set-hat-beta}
{\mathcal{A}}_j(\hat \beta) = \{\hat l_{j,1}, \ldots,  \hat l_{j,s_j}\},
\end{equation}
where $\hat l_{j,1} < \cdots <  \hat l_{j,s_j}$ and $s_j = |{\mathcal{A}}_j(\hat \beta)|$.
Finally, we obtain the following $\mu^\star_{j,\bullet}$'s estimator
\begin{equation}
\label{cut-points-estimator}
\widehat{\mu}_{j,\bullet}= (\mu_{j,\hat l_{j,1}}, \ldots, \mu_{j,\hat l_{j,s_j}})^\top
\end{equation}
for all $j = 1, \ldots, p$. By construction, $K^\star_j$ is estimated by $\widehat{K}_j = s_j$. Some details on the algorithm used to solve the regularization problem~\eqref{problem-estim} are given in Appendix~\ref{sec:Algorithm}.

\section{Theoretical guarantees}
\label{sec:theoretical guarantees}

\subsection{Oracle inequality for prediction}
This section is devoted to a first theoretical result. In order to evaluate the prediction error, we first define the (empirical) Kullback-Leibler divergence~\citep{senoussi1990probleme} $KL_n$ between the true function $f^\star$ and any candidate $f$ as
\begin{equation}
\label{eq:def-KLn}
KL_n(f^\star,f) = \frac{1}{n} \sum_{i=1}^n \int_0^\tau  \log \bigg\{ \frac{e^{f^\star(X_i)}\sum_{i=1}^nY_i(t)e^{f(X_i)}}{e^{f(X_i)}\sum_{i=1}^nY_i(t)e^{f^\star(X_i)}} \bigg\}
Y_i(t) \lambda_0^\star(t) e^{f^\star(X_i)} \dd t,
\end{equation}
where we denote $Y_i(t) = \ind{}(Z_i\geq t)$ the at-risk process, and $\tau > 0$ is to be defined later.

We seek to establish an oracle inequality expressed in terms of a compatibility factor~\citep{vandegeer2009} satisfied by the following non-negative symmetric matrix: 
\begin{equation}
\label{matrix-Sigma-star}
  \Sigma_n(f^\star, \tau) = \frac{1}{n}\sum_{i=1}^n \int_0^{\tau} \big(X_i^B - \bar{X}_n(s)\big)\big(X_i^B - \bar{X}_n(s)\big)^\top y_i(s)e^{f^\star(X_i)} \lambda_0^\star(s)\dd s,
\end{equation}
where 
\begin{equation*}
  \bar{X}_n(s) = \frac{\sum_{i=1}^n X_i^By_i(s) e^{f^\star(X_i)}}{\sum_{i=1}^ny_i(s) e^{f^\star(X_i)}}
\end{equation*}
and 
\begin{equation*}
y_i(s) = \E[Y_i(s) | X_i] \text{ for all } 0 \leq s \leq t \text{ and all } i = 1, \ldots, n.
\end{equation*}

For any concatenation of index subsets $L=[L_1, \ldots, L_p]$, we define the compatibility factor
\begin{equation}
  \label{compatibility-condition}
  \kappa_{\tau}(L) = \inf\limits_{\beta \in \mathscr{C}_{\TV,\omega}(L) \backslash \{\bold 0 \}}\frac{\sqrt{\beta^\top\Sigma_n(f^\star, \tau)\beta}}{\norm{\beta_{L}}_2},
\end{equation}
where
\begin{equation*}
\label{C-AGG}
\mathscr{C}_{\TV,\omega}(L) \stackrel{}{=} \Big\{\beta \in \mathscr{B}_{p+d}(R): \sum_{j=1}^p\norm{(\beta_{j, \bullet})_{L^\complement_j}}_{\TV,\omega_{j,\bullet}} \leq 3\sum_{j=1}^p \norm{(\beta_{j, \bullet})_{L_j}}_{\TV,\omega_{j,\bullet}} \Big\}
\end{equation*}
is a cone composed of all vectors with similar support $L$.
\begin{assumption}\label{assump:tau}
$\tau$ is hereafter assumed to satisfy
\begin{align*}
\max_{1 \leq i \leq n} \int_0^\tau  \lambda^\star (t|X_{i}) \dd t < \infty \quad \text{ and } \quad \min_{1 \leq i \leq n} \mathbb P( C_i > \tau | X_i) > 0.
\end{align*}
\end{assumption}
Such assumptions on $\tau$ are common in survival analysis, see e.g.,~\cite{andersen2012statistical} and~\cite{lemler2016oracle}. We refer the reader to~\cite{gill1983large} for a discussion on the role of $\tau.$ In addition, we define $c_Z := \min_{1 \leq i \leq n} y_i(\tau)$ and remark that
\begin{equation*}
  c_Z \geq \exp\big(-\max_{1 \leq i \leq n} \int_0^\tau  \lambda^\star (t|X_{i}) \dd t\big) \min_{1 \leq i \leq n} \mathbb P( C_i > \tau | X_i) > 0.
\end{equation*}
For the sake of simplicity, we introduce the additional notation:
\begin{align*}
f^\star_\infty&= \max_{1 \leq i \leq n}|f^\star(X_i)|, \quad s^{(0)}(\tau) = n^{-1}\sum_{i=1}^n y_i(\tau)e^{f^\star(X_i)}, \quad \text{ and } \quad
\Lambda^\star_0(\tau) = \int_0^{\tau} \lambda_0^\star(s)\dd s.
\end{align*}

\begin{assumption}\label{assump:compatibility}
Let $\varepsilon\in (0,1)$ and define $t_{n,p,d,\varepsilon}$ as the solution of $$2.221(p+d)^2 \exp\{-nt^2_{n,p,d,\varepsilon} / (2 + 2t_{n,p,d,\varepsilon}/3)\} = \varepsilon.$$ 
For any concatenation set $L=[L_1, \ldots, L_p]$ such that $\sum_{j=1}^p |L_j| \leq K^\star$, assume that 
\begin{equation*}
  \kappa^2_{\tau}(L) > \Xi_{\tau}(L),
\end{equation*}
where 
\begin{align*}
  \Xi_{\tau}(L) = 4|L|\Big(\frac{8\max_{j}(d_j+1) \max_{j,l}\omega_{jl}}{\min_{j,l}\omega_{j,l}}\Big)^2 \Big\{&\big(1 + e^{2f^\star_\infty}\Lambda^\star_0(\tau)\big)\sqrt{(2/n)\log\big(2(p+d)^2/\varepsilon\big)}\\
  &+ \big(2e^{2f^\star_\infty}\Lambda^\star_0(\tau)/s^{(0)}(\tau)\big)t^2_{n,p,d,\varepsilon} \Big\}.
\end{align*}
\end{assumption}
Note that $\kappa^2_{\tau}(L)$ is the smallest eigenvalue of a population integrated covariance matrix defined in~\eqref{matrix-Sigma-star}, so it is reasonable to treat it as a constant.
Moreover, $t^2_{n,p,d,\varepsilon}$ is of order 
\begin{equation*}
  \dfrac1n \log \dfrac{(p+d)^2}{\varepsilon},
\end{equation*}
so if $|L|\log(p+d)/n$ is sufficiently small, Assumption~\ref{assump:compatibility} is verified.
With these preparations made, let us now state the oracle inequality for prediction satisfied by our estimator of $f^\star$ which is, by construction, given by $\hat f = f_{\hat \beta}$ (see~\eqref{f-beta}).

\begin{theorem}
\label{thm-faster-in-approx}
The  inequality
\begin{align}
\label{OI-faster}
KL_n(f^\star,f_{\hat \beta}) 
\leq  &\inf_{\beta}\bigg\{3KL_n(f^\star,f_{\beta}) 
+ \frac{1024(f^\star_\infty + R + 2)}{{{\kappa}^2_{\tau}\big(\mathcal{A}({\beta})\big) - \Xi_{\tau}\big(\mathcal{A}({\beta})\big)}}|\mathcal A(\beta)|\max_{1 \leq j \leq p}\norm{(\omega_{j,\bullet})_{\mathcal{A}_j(\beta)}}^2_\infty \bigg\} 
\end{align}
holds with a probability greater than $1 - 28.55 e^{-c} - e^{-ns^{(0)}(\tau)^2/8e^{2f^\star_\infty}} - 3\varepsilon$ for some $c>0$,  where the infimum is over the set of vectors $\beta \in \mathscr{B}_{p+d}(R)$ such that $n_{j,\bullet}^\top\beta_{j,\bullet}=0$ for all $j=1, \ldots,p$, and such that $|\mathcal A(\beta)| \leq K^\star$. 
\end{theorem}
The proof of Theorem~\ref{thm-faster-in-approx} is postponed to Appendix~\ref{proof-theorem-faster-OI}.
The second term in the right-hand side of~\eqref{OI-faster} can be viewed as a ``variance'' (or ``complexity'') term, and its dominant term satisfies
\begin{equation*}
\frac{|\mathcal{A}({\beta})|\max_j\norm{(\omega_{j,\bullet})_{\mathcal{A}_j(\beta)}}^2_\infty}{\kappa^2_\tau\big(\mathcal A(\beta)\big) - \Xi_\tau\big(\mathcal A(\beta)\big)} \lesssim \frac{|\mathcal{A}({\beta})|}{\kappa^2_\tau\big(\mathcal A(\beta)\big) - \Xi_\tau\big(\mathcal A(\beta)\big)} \frac{\log(p+d)}{n},
\end{equation*}
where the symbol $\lesssim$ means that the inequality holds up to a multiplicative constant. 
Then, one obtains the expected fast convergence rate $\bigO\big(\log(p+d)/n\big)$ for the estimator $\hat f$. Note that, in the proof of Theorem~\ref{thm-faster-in-approx}, the fact that the true $f^\star$  lies in the true Cox model with cut-points is not necessary. Hence Theorem~\ref{thm-faster-in-approx} can be applied to any $f^\star$. 

The value $|\mathcal{A}({\beta})|$ characterizes the sparsity of the vector $\beta$, since it counts the number of non-equal consecutive values of $\beta$. 
If $\beta$ is block-sparse, namely whenever $|\mathscr{A}(\beta)| \ll p$ where $\mathscr{A}(\beta) = \{j = 1, \ldots, p : \beta_{j, \bullet} \neq \textbf{0} \}$ (meaning that few raw features are useful for prediction), then $|\mathcal{A}(\beta)| \leq |\mathscr{A}(\beta)| \max_{j \in \mathscr{A}(\beta)} |\mathcal{A}_j(\beta)|$, which means that $|\mathcal{A}(\beta)|$ is controlled by the block sparsity $|\mathscr{A}(\beta)|$.
Also, the oracle inequality still holds for vectors such that $n_{j,\bullet}^\top\beta_{j,\bullet}=0$, which is natural since the binarsity penalization imposes these extra linear constraints.

The assumption  $\beta \in \mathscr{B}_{p+d}(R)$ is a technical one,  allowing a connection, via the notion of self-concordance~\citep{bach2010selfconcordance}, between the empirical squared $\ell_2$-norm and the empirical Kullback-Leibler (see Lemma~\ref{lemma:self-concordance}).
Also, note that
\begin{equation}
\label{eq:control_inner_ball}
\max_{1 \leq i \leq n} | \beta^\top X_i^B | \leq \sum_{j=1}^p 
\norm{\beta_{j , \bullet}}_\infty \leq |\mathscr{A}(\beta)|
\times \norm{\beta}_\infty,
\end{equation}
where $\norm{\beta}_\infty = \max_{1 \leq j \leq p} \norm{\beta_{j, \bullet}}_\infty$.
The first inequality in~\eqref{eq:control_inner_ball} comes from the fact that the entries of $\XB$ are in $\{0, 1\}$, and entails that $ \max_{1 \leq i \leq n} | \beta^\top X_i^B | \leq R$ whenever $\beta \in \mathscr{B}_{p+d}(R)$.

The second inequality in~\eqref{eq:control_inner_ball} shows that $R$ can be upper bounded by $|\mathscr{A}(\beta)| \times \norm{\beta}_\infty$, and therefore the constraint $\beta \in \mathscr{B}_{p+d}(R)$ becomes merely a box constraint on $\beta$, which depends on the dimensionality of the features through $|\mathscr{A}(\beta)|$ only.
The fact that the procedure depends on $R$, and that the oracle inequality stated in Theorem~\ref{thm-faster-in-approx} depends linearly on $R$, is commonly found in the literature on sparse generalized linear models, see~\cite{van2008high,bach2010selfconcordance,ivanoff2016adaptive}.
However, the constraint $\mathscr{B}_{p+d}(R)$ is a technicality which is not used in the numerical experiments  in Sections~\ref{sec:Performance evaluation} and~\ref{sec:application}.

Notice in addition that our proof is different from that  of~\cite{huang2013} and could be applied in their setting (lasso in the Cox model with time-dependent covariates). Alternative oracle inequalities, in terms of the Kullback-Leibler divergence instead of the symmetric Bregman divergence, could hence be proven.

\subsection{Oracle inequality for estimation}

\paragraph{Approximation of $f^\star$.}
\label{pragraph-1-section-theory}

Since $\beta ^\star \in \R^{p+K^\star}$ and $\hat \beta \in \R^{p+d}$, we define in this section an approximation of $f^\star$ denoted $f_{b^\star}$ with $b^\star \in \R^{p+d}$. 
We choose $d_j$ such that 
\begin{equation*}
\min_{1 \leq k \leq  K^\star_{j}+1}|I^\star_{j,k}| \geq  \max_{1 \leq l \leq d_j+1}|I_{j,l}| \;\text{for all }j = 1,\ldots ,p.
\end{equation*} This choice ensures that for all features $j = 1,\ldots,p$, there exists a unique interval $I_{j,l}$  containing cut-point $\mu^\star_{j,k}$, which we denote 
\begin{equation}
\label{eq:def_l_star}
I_{j, l^\star_{j,k}} = (\mu_{j, l^\star_{j,k}-1}, \mu_{j,l^\star_{j,k}}]
\end{equation}
for all $k = 1,\ldots,K_j^\star$. Note that in practice, this requirement is met by increasing $d_j$.
For each single $j$th block, let us recall that as defined in~\eqref{f-star}, we associate with $\beta^\star_{j,\bullet}$ the $\mu_{j,\bullet}^\star$-piecewise constant function 
\begin{equation*}
f^\star_{j}: x \mapsto \sum_{k=1}^{K^\star_j +1} \beta^\star_{j,k}\ind{}(x\in I^\star_{j,k})
\end{equation*}
defined for all $x\in [0,1]$.
Now, let us define the $\mu_{j,\bullet}$-piecewise constant function
\begin{equation}
\label{eq:approx-beta-star}
\tilde{f}_j: x \mapsto \sum_{k=1}^{K^\star_j+1} \beta^\star_{j,k} \sum_{l=l^\star_{j,k-1} + 1}^{l^\star_{j,k}} \ind{}(x \in I_{j,l}),
\end{equation}
for $x\in [0,1]$, where $l^\star_{j,k}$ is defined in~\eqref{eq:def_l_star}, and with the conventions $l^\star_{j,0} = 0$ and $l^\star_{j,K^\star_j + 1} = d_j+1$ for all $j =1, \ldots, p$. With this definition, $\tilde{f}_j$ has the same number of jumps and amplitudes  thereof as $f^\star_{j}$. The only difference between these two functions is the location of the jumps: $f^\star_j$ jumps once for each cut-point $\mu^\star_{j,k}$ for all $k = 1,\ldots, K^\star_j+1$, while $\tilde{f}_j$ jumps once for each $\mu_{j,l}$  closest (on the right hand side) to $\mu^\star_{j,k}$ for all $k = 1,\ldots, K^\star_j+1$. This choice of approximation  is discussed at the beginning of Appendix~\ref{proof-thm-estim}.

In the $j$th block, the vector associated with $\tilde{f}_j$ now lives in $\R^{d_j+1}$ as expected, but the extra linear constraint required to apply Theorem~\ref{thm-faster-in-approx} is not fulfilled. We then define
\begin{equation}
\label{definition-of-b-star}
  f_{b^\star_{j,\bullet}} : x \mapsto \tilde{f}_{j}(x) - \frac 1n\sum_{i=1}^n \tilde{f}_j(X_{i,j})
\end{equation}
for $x\in [0,1]$, which gives rise to
$n_{j,\bullet}^\top  b^\star_{j,\bullet}=0$ for all $j = 1, \ldots, p$, where $b^\star_{j,\bullet} \in \R^{d_j+1}$ is the vector associated with $f_{b^\star_{j,\bullet}}$.

\begin{figure}[!htp]
\centering
\includegraphics[width=0.9\textwidth]{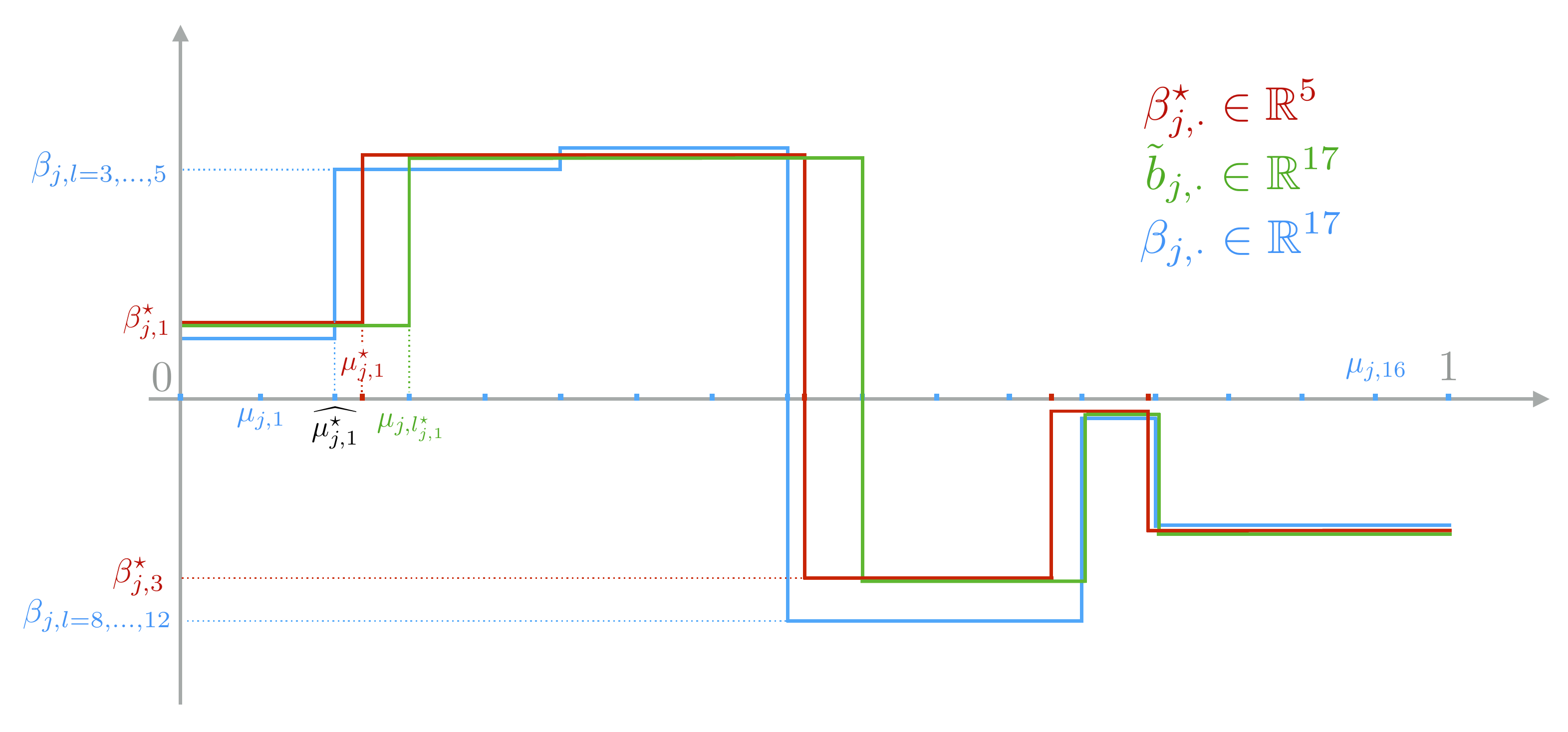}
\caption{Illustration of the different vectors for the $j$th block, with $d_j=17$. In this scenario, the algorithm detects an extra cut-point and $\widehat{K}_j = 5 = s_j$, while $K^\star_j = 4$.}
\label{fig:illustration}
\end{figure}

Denoting $b^\star = \big((b^\star_{1, \bullet})^\top, \ldots, (b^\star_{p, \bullet})^\top\big)^\top$, our approach to prove the oracle inequality for estimation relies on the application of Theorem~\ref{thm-faster-in-approx} to the approximate candidate $b^\star \in \R^{p+d}$ of $\beta^\star$. Figure~\ref{fig:illustration} gives a clearer view of the different quantities involved so far in the estimation procedure on a toy example. See also the upper part of Figure~\ref{fig:estimation} in Section~\ref{sec:results-simulation}. 
Note that, in addition, if $\beta^\star$ is block-sparse, then it is also the case for $b^\star$, and the following holds:
\begin{equation*}
|\mathscr{A}(b^\star)| \leq |\mathscr{A}(\beta^\star)|.
\end{equation*}

Let us introduce some further notation. We define 
\begin{equation}
\label{eq:def-pi_n}
\pi_n = \frac{|\{i=1, \ldots, n: N_i(\tau) = 1\}|}{n},
\end{equation} 
and let in addition 
\begin{equation*}
R^\star = \sum_{j \in \mathscr A(\beta^\star)}\norm{b^\star_{j,\bullet}}_\infty,
\end{equation*}
\begin{equation*}
{\bf{I}} = 2 \big( | \mathscr A (\beta^\star) |+ K^\star \big) \Big(1 + 3\frac{ \psi(f_\infty^\star + R^\star +2 )}{f_\infty^\star + R^\star+2}\Big)
\pi_n \max_{j \in \mathscr A(\beta^\star)} \| \beta_{j,\bullet} \|^2_{\infty} \max_{j \in \mathscr A(\beta^\star)} \| n_{j,\bullet}/n\|^2_\infty  \bigg(1+ \frac{4e^{2f^\star_\infty}}{c_Z} \bigg),
\end{equation*}
where $\psi(x) = e^{x} - x - 1$, and 
\begin{align*}
{\bf{II}} = \frac{2048(f^\star_\infty + R^\star + 2)^2K^\star\max_{1 \leq j \leq p}\norm{(\omega_{j,\bullet})_{\mathcal{A}_j(b^\star)}}^2_\infty}{{{\kappa}^2_{\tau}\big(\mathcal{A}({b^\star})\big) - \Xi_{\tau}\big(\mathcal{A}({b^\star})\big)}}.
\end{align*}

\begin{theorem}
\label{thm-oracle-estimation}
The  inequality
\begin{align}
\label{OI-estimation}
\norm{(\hat\beta - b^\star)_{\mathcal{A}(b^\star)}}_1 \leq 
\frac{\sqrt{{K^\star(\bf{I}} + {\bf{II})}}}{\kappa_\tau\big(\mathcal{A}(b^\star)\big)}
\end{align}
holds with probability greater than $1 - 28.55 e^{-c} - e^{-ns^{(0)}(\tau)^2/8e^{2f^\star_\infty}} - 3\varepsilon - 2 e^{-nc_Z^2/2}$ for some $c>0$. 
\end{theorem}

A proof of Theorem~\ref{thm-oracle-estimation} is presented in Appendix~\ref{proof-thm-estim}. The term $\bf{I}$ is a bias term and, if all $d_j \to \infty$ as $n \to \infty$ and under mild conditions on the distributions of the $X_{i,j}$, it goes to $0$ as $n \to \infty$. The order of magnitude in the inequality of Theorem~\ref{thm-oracle-estimation} is then given, for $n$ and $d_j$ large enough, by
\begin{equation*}
\frac{\sqrt{{K^\star(\bf{I}} + {\bf{II})}}}{\kappa_\tau\big(\mathcal{A}(b^\star)\big)}
\lesssim \frac{K^\star\sqrt{\log(p+d)/n}}{\kappa_\tau\big(\mathcal{A}(b^\star)\big)\sqrt{\kappa^2_\tau\big(\mathcal A(b^\star)\big) - \Xi_\tau\big(\mathcal A(b^\star)\big)}},
\end{equation*}
which is the expected fast rate in oracle inequalities for estimation, see for instance~\cite{bickel2009simultaneous}.

\section{Performance evaluation}
\label{sec:Performance evaluation}

\subsection{Practical details}
\label{sec:practical}
Let us now give some details about the binacox's use in practice. First, as already mentioned, we naturally choose the estimated quantiles for the $\mu_{j,l}$. This choice provides two major practical advantages: $i$) the resulting grid is data-driven and follows the distribution of $\bX_{\bullet, j}$, and $ii$) there is no need to tune hyper-parameters $d_j$ (number of bins for the one-hot encoding of raw feature $j$). Indeed, if $d_j$ is ``large enough'' (we take $d_j = 50$ for all $j =1, \ldots, p$ in practice), increasing $d_j$ barely changes the results since the cut-points selected by the penalization no longer change, and the size of each block automatically adapts itself to the data; depending on the distribution of $\bX_{\bullet, j}$, ties may appear in the corresponding empirical quantiles (for more details on this last point, see~\citet{alaya2016}).

\begin{figure}[!htp]
\centering
\includegraphics[width=0.8\textwidth]{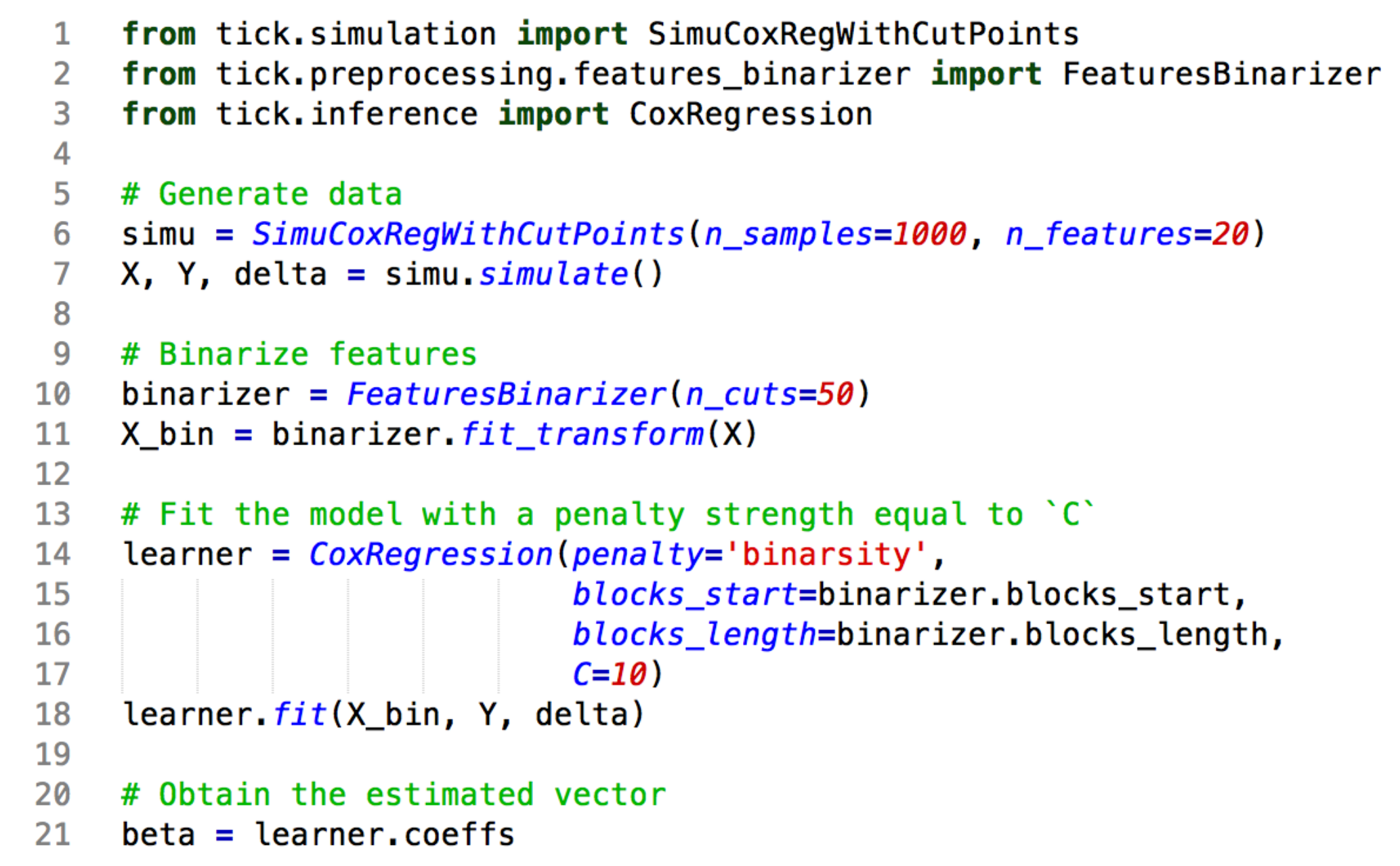}
\caption{Sample python code for the use of the binacox in the \texttt{tick} library, using the \texttt{FeaturesBinarizer} transformer for feature binarization.}
\label{fig:code}
\end{figure}

Note also that the binacox is proposed in the~\texttt{tick} library~\citep{2017arXiv170703003B}, and that all the code used in this paper is open-sourced at \url{https://github.com/SimonBussy/binacox} ; we provide sample code for its use in Figure~\ref{fig:code}. For practical convenience, we take all weights $\omega_{j,l} = \gamma$ and select the hyper-parameter $\gamma$ using a $V$-fold cross-validation procedure with $V = 10$, taking the negative partial log-likelihood defined in~\eqref{binarized-model} as a score computed after a refit of the model on the binary space obtained by the estimated cut-points, and with the sum-to-zero constraint only (without the TV penalty, which actually gives a fair estimate of $\beta^\star$ in practice), which intuitively makes sense. Figure~\ref{fig:cv} in Appendix~\ref{sec:Implementation} gives the learning curves obtained with this cross-validation procedure on an example.

We also add a simple de-noising step in the cut-point detection phase, which is useful in practice. Indeed, it is usual to observe two consecutive $\hat \beta$'s jumps in the neighbourhood of a true cut-point, leading to an over-estimation of $K^\star$. This can be viewed as a clustering problem. We tried different clustering methods but in practice, nothing works better than this simple routine: if $\hat \beta$ has three consecutive different coefficients within a block, then only the largest jump is considered as a ``true'' jump. Figure~\ref{fig:groups} in Appendix~\ref{sec:Implementation} illustrates this routine.

\subsection{Simulation}
\label{sec:simulation}

In order to assess the methods, we run an extensive Monte Carlo
simulation study. Let us first present the design used in the following.

\subsubsection{Design.}

\begin{figure}[!htb]
\centering
\subfigure{
	\includegraphics[width=.5\textwidth,valign=c]{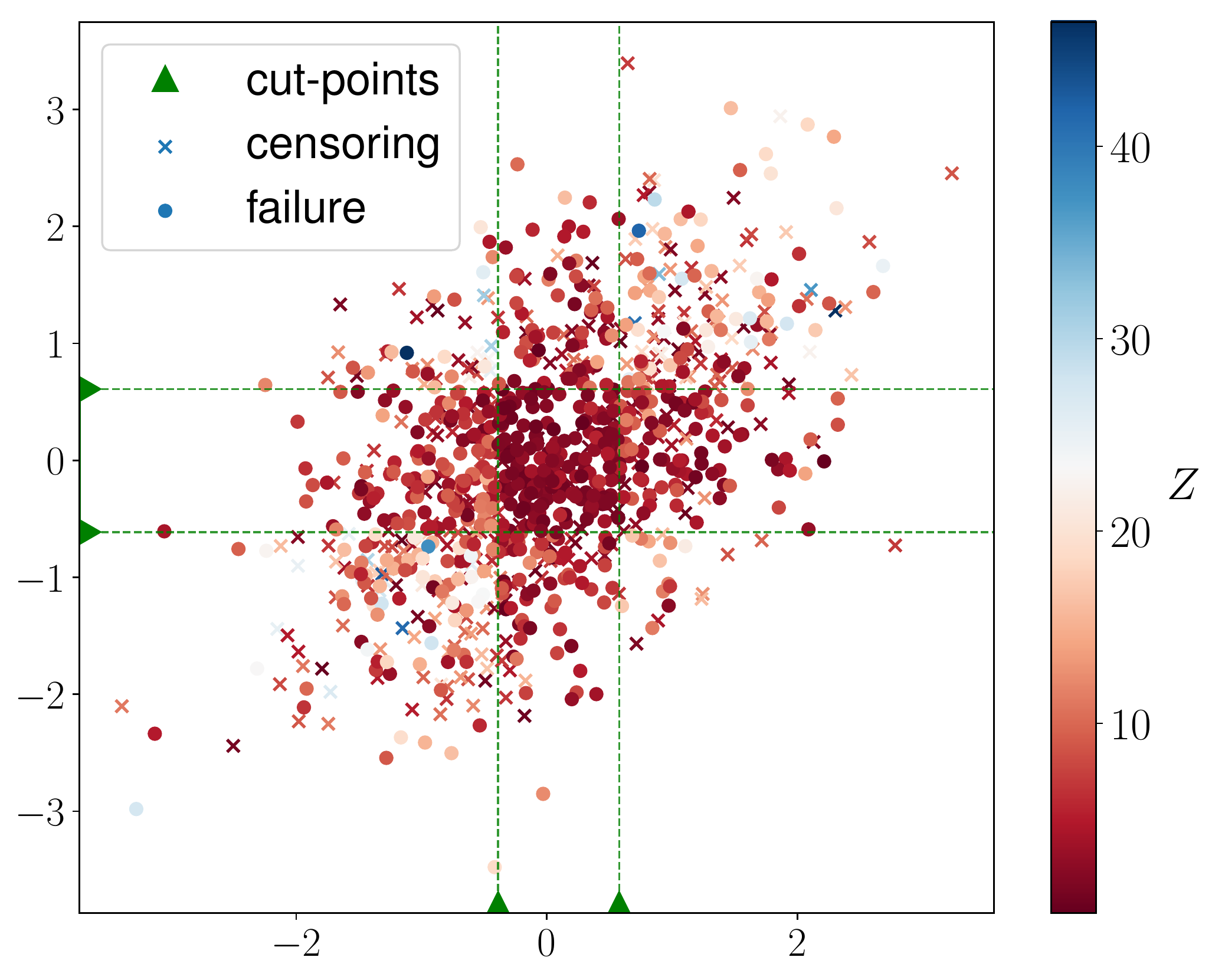}
	\includegraphics[width=.5\textwidth,valign=c]{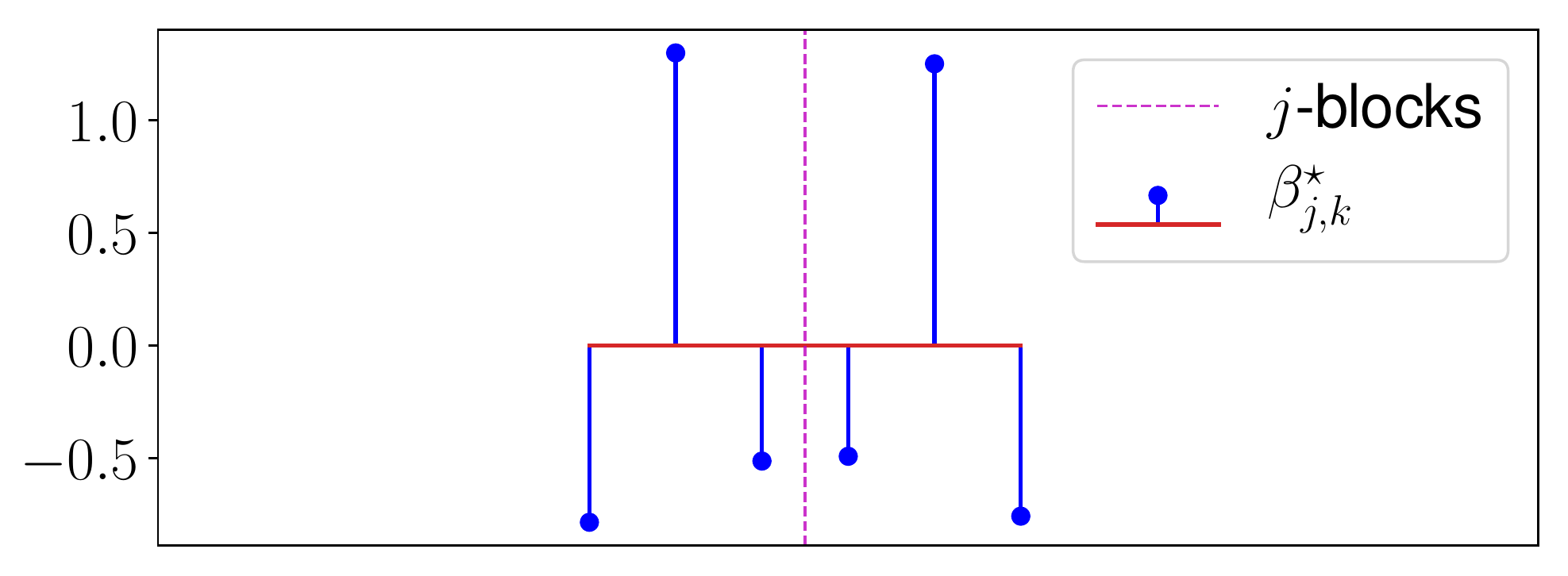}
}
\caption{Left: illustration of data simulated with $p=2$, $K^\star_1=K^\star_2=2$, and $n=1000$. Dots represent failure times $(z_i = t_i)$ while crosses represent censoring times $(z_i = c_i)$, and the colour gradient represents the $z_i$ values (red for low and blue for high). Right: $\beta^\star$ is plotted, with a dotted line to demarcate the two blocks (since $p=2$).}
\label{fig:example-data}
\end{figure}

We first take $[X_{i,j}] \in \R^{n \times p} \sim \cN \big(0,\bSigma(\rho)\big)$, with $\bSigma(\rho)$ a $(p \times p)$ Toeplitz covariance matrix \citep{mukherjee1988some} with correlation $\rho \in (0, 1)$. For each feature $j = 1, \ldots, p$, we sample the cut-points $\mu^\star_{jk}$ uniformly without replacement from the estimated quantiles $q_j(u / 10)$ for $u =1, \ldots, 9$ and $k =1, \ldots, K^\star_j$. In this way, we avoid having undetectable cut-points (with very few examples above the cut-point value) or pairs of overly close together indissociable cut-points. We choose the same $K^\star_j$ values for all $j=1, \ldots, p$. Now that the true cut-points vector $\mu^\star$ has been generated, one can compute the corresponding binarized version of the features, which we denote $x_i^{B^\star}$ for the $i$th example. 
Then, we generate 
\begin{equation*}
  c_{jk} \sim (-1)^k|\cN(1, 0.5)|
\end{equation*}
for all $k = 1, \ldots, K^\star_j + 1$ and $j =1, \ldots, p$ to make sure we create ``real'' cut-points, and take 
\begin{equation*}
  \beta_{jk}^\star = c_{jk} - (K^\star_j + 1)^{-1}\sum_{k=1}^{K^\star_j + 1} c_{jk}
\end{equation*}
in order to impose the sum-to-zero constraint of the true coefficients in each block. We also induce a sparsity aspect by uniformly selecting a proportion $r_s$ of features $j \in \cS$ with no cut-point effect, i.e.,  features for which we enforce $\beta_{jk}^\star = 0$ for all $k = 1, \ldots, K^\star_j + 1$. Lastly, we generate survival times using Weibull distributions, which is a common choice in survival analysis~\citep{klein2005survival}:  
\begin{equation*}
  T_i \sim \nu^{-1} \big[- \log(U_i)\exp\big(-(x_i^{B^\star})^\top \beta_i^\star \big)\big]^{1/\varsigma}
\end{equation*}
with $\nu > 0$ and $\varsigma > 0$ the scale and shape parameters respectively, and $U_i \sim \cU([0,1])$, where $\cU([a,b])$ stands for the uniform distribution on a segment $[a,b]$. The distribution of the censoring variable $C_i$ is the geometric distribution $\cG(\alpha_c)$, where $\alpha_c \in (0, 1)$ is empirically tuned to maintain a desired censoring rate $r_c \in [0,1]$. The choice of all hyper-parameters is driven by the applications on real data presented in Section~\ref{sec:application}, and summarized in Table~\ref{table:parameters choice}. Figure~\ref{fig:example-data} gives an example of data generated according to the design we have just described.
\begin{table}
\caption{Hyper-parameter choices for simulation.\label{table:parameters choice}}
\centering
\begin{tabular}{cccccccc}
\toprule
$n$ & $p$ & $\rho$ & $K^\star_j$ & $\nu$ & $\varsigma$ & $r_c$ & $r_s$ \\
\midrule
(200, 4000) & 50 & 0.5 & $\{1, 2, 3\}$ & 2 & 0.1 & 0.3 & 0.2 \\
\bottomrule
\end{tabular}
\end{table}

\subsubsection{Metrics.} 

We evaluate the methods being analysed using two metrics. The first assesses the estimation of the cut-points values by 
\begin{equation*}
  m_1 = |\cS'|^{-1} \sum_{j\in \cS'} \ \cH(\cM^\star_j, \widehat\cM_j),
\end{equation*}
where $\cM^\star_j = \{\mu^\star_{j,1}, \ldots, \mu^\star_{j,K^\star_j}\}$ (resp. $\widehat\cM_j = \{\hat\mu_{j,1}, \ldots, \hat\mu_{j,\widehat K_j}\}$) is the set of true (resp. estimated) cut-points for feature $j$,\, $\cS' = \big\{j,\ j \notin \cS \cap \{l,\ \widehat\cM_l = \emptyset\} \big\}$ the indexes corresponding to features with at least one true cut-point and one detected cut-point, and $\cH(A, B)$  the Hausdorff distance between the sets $A$ and $B$, defined as
\begin{equation*}
  \cH(A,B) = \max \big(\mathcal{E}(A||B), \mathcal{E}(B||A)\big),
\end{equation*}
where $\mathcal{E}(A||B) = \sup_{b\in B}\inf_{a \in A} |a - b|$.
This is inspired by~\citet{HarLev-10}, except that in our case, both $\cM^\star_j$ and $\widehat\cM_j$ can be empty, which explains the use of $\cS'$.
The second metric we use is precisely focused on the sparsity aspect; it assesses the ability for each method to detect features with no cut-points, and is defined by 
\begin{equation*}
  m_2 = |\cS|^{-1} \sum_{j\in \cS} \widehat K_j.
\end{equation*}

\subsection{Competing methods}
\label{sec:competing methods}

To the best of our knowledge, all existing algorithms and methods are based on multiple log-rank tests in univariate models. These methods are widely used, and recent implementations include the web applications \texttt{Cutoff Finder} and \texttt{Findcutoffs} described in~\cite{budczies2012cutoff} and~\cite{chang2017determining} respectively.

We describe in what follows the principle of these univariate log-rank tests. Consider one of the initial variables $\bX_{\bullet, j} = (x_{1,j}, \ldots,x_{n,j})^\top$, and denote its 10th and 90th quantiles
as $x_{10th,j}$ and $x_{90th,j}$. Then, define a grid
$\{g_{j,1}, \ldots, g_{j,\kappa_j}\}$. In most implementations, the $g_{j,k}$'s are chosen at the original observation points and are such that $x_{10th,j} \leq g_{j,k} \leq x_{90th,j}$.
For each $g_{j,k}$, the $p$-value $\text{pv}_{j,k}$ of the log-rank test associated with the univariate Cox model defined by
\begin{equation*}
\lambda_0(t) \exp\big( \beta^j \ind{}(x \leq g_{j,k})\big)
\end{equation*}
is computed (via the \texttt{python} package \texttt{lifelines} in our implementation). For each initial variable $\bX_{\bullet, j}$, $\kappa_j$ $p$-values are available at this stage. The choice of the size $\kappa_j$ of the grid depends on the implementation, and ranges for several dozen to all observed values between $x_{10th,j}$  and $x_{90th,j}$. 

In Figure~\ref{fig:estimation}, the values $-\log(\text{pv}_{j,k})$ for $k=1, \ldots, \kappa_j$ (denoted by ``MT'' for ``Multiple Testing'') are represented, for the simulated example illustrated in Figure~\ref{fig:example-data}. Notice that the level $-\log(\alpha)=-\log(0.05)$ is exceeded for numerous $g_{j,k}$'s values, and of course this procedure allows us to detect only a single cut-point per feature.
A common approach is to consider the maximal value $-\log(\text{pv}_{j,\hat k})$ and then define the cut-point for variable $j$ as $g_{j,\hat k}$. 
As argued in~\cite{altman1994dangers}, this is obviously ``associated with an inflation of type I error'', and for this reason we do not consider this approach.

To cope with the multiple testing (MT) problem at hand, multiple testing corrections have to be applied, of which we consider two. The first is the well-known Bonferroni $p$-value correction, referred to as MT-B in the following. We insist on the fact that although commonly used, this method is not correct in this situation since the $p$-values are correlated. Note also that in this context, the Benjamini–Hochberg (BH) procedure would result in the same cut-points being detected as MT-B (with FDR=$\alpha$), since we only consider as a cut-point candidate the points with minimal $p$-value. Indeed, applying the classical BH procedure would select far too many cut-points.
The second correction, denoted MT-LS, is the correction proposed in~\citet{lausen1992maximally}, based on asymptotic theoretical considerations. Figure~\ref{fig:estimation} also illustrates how these corrections behave on the simulated example illustrated in Figure~\ref{fig:example-data}.
A third correction we could imagine would be a bootstrap-based MaxT procedure (or MinP) as proposed in~\citet{dudoit2007multiple} or~\citet{westfall1993adjusting}, but this would be intractable in our high-dimensional setting (see Figure~\ref{fig:computing-times-comparison} that compares the computing times for a single feature only; a bootstrap procedure based on MT would dramatically increase the required computing time).

\subsection{Simulation results}
\label{sec:results-simulation}

\subsubsection{Example.} 
\begin{figure}[!htb]
\centering
\includegraphics[width=\textwidth]{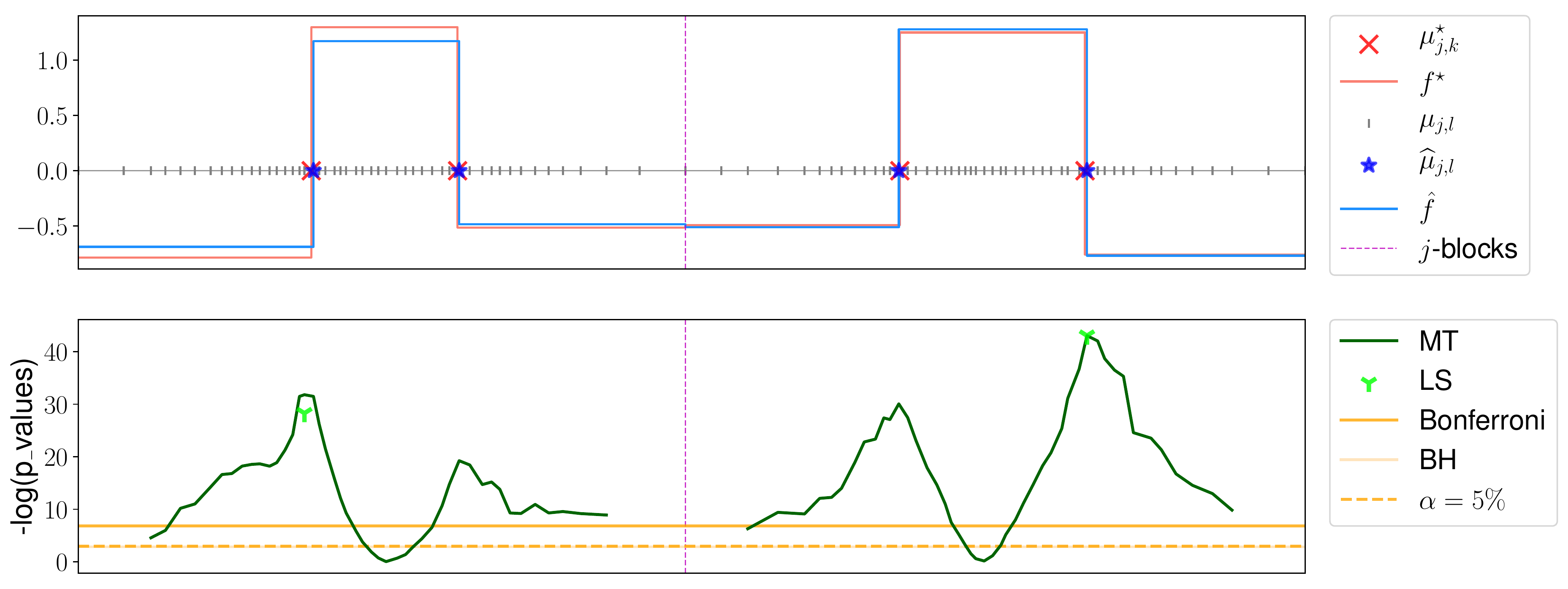}\hfill
\caption{Top: Illustration of the main quantities involved in the binacox, with estimations obtained for the data represented in Figure~\ref{fig:example-data}. Our algorithm detects the correct number of cut-points $\widehat K_j = 2$, and estimates their positions accurately, as well as their amplitudes. Bottom: results obtained using the multiple testing-related methods introduced in Section~\ref{sec:competing methods}. Here the BH threshold lines overlap that corresponding to $\alpha=5\%$. The BH procedure would consider as cut-points all  $\mu_{j,l}$ values for which the corresponding dark green (MT) line's values are above this, thus detecting far too many cut-points.}
\label{fig:estimation}
\end{figure}

Figure~\ref{fig:estimation} illustrates how the  methods considered behave on the data shown in Figure~\ref{fig:example-data}. With the help of this example, we can clearly see the good performance of the binacox method: the position, strength and number of cut-points are well estimated.
The MT-B and MT-LS methods can only detect one cut-point by construction. Both methods detect ``the most significant'' cut-point for each of the 2 features, namely those corresponding to the highest jumps in $\beta^\star_{j,\bullet}$ (see Figure~\ref{fig:example-data}): $\mu^\star_{1,1}$ and $\mu^\star_{2,2}$. 

With regards to the shape of the ``$p$-value curves'', one can see that for each of the two features, the two ``main'' local maxima correspond to the true cut-points. One could then imagine creating  a method for detecting such maxima, but this is beyond the scope of this paper (plus it would still be based on MT methods, which have high computational costs, as detailed hereafter).

\subsubsection{Computing times.}

\begin{figure}[!htb]
\centering
\subfigure[Average computing times in seconds (with the black lines representing $\pm$ the standard deviation) obtained on 100 simulated datasets (according to Section~\ref{sec:simulation} with $p=1$ and $K^\star=2$) for training the binacox versus the multiple testing methods, where cut-point candidates are either all $x_i$ values between the 10th and 90th empirical quantiles of $X$ (``MT all''), or the same candidates as the grid considered by the binacox (``MT grid'').]{\includegraphics[width=.58\textwidth,valign=c]{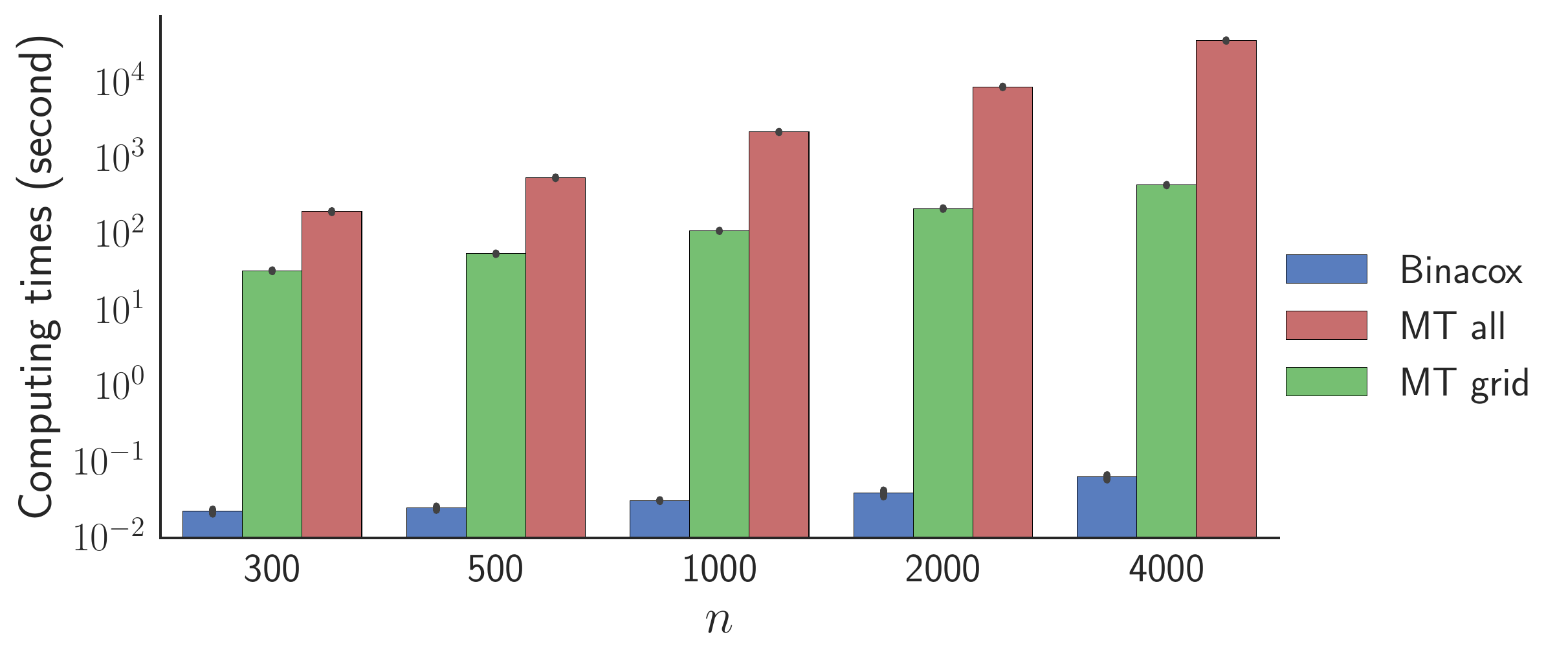}\label{fig:computing-times-comparison}}
\hspace{.02\textwidth}
\subfigure[Average (bold) computing times in seconds and standard deviation (bands) obtained on 100 simulated datasets (according to Section~\ref{sec:simulation} with $K^\star_j=2$) for training the binacox when increasing the dimension $p$ up to 100. The method remains very fast in high-dimensional settings.]{\includegraphics[width=.38\textwidth,valign=c]{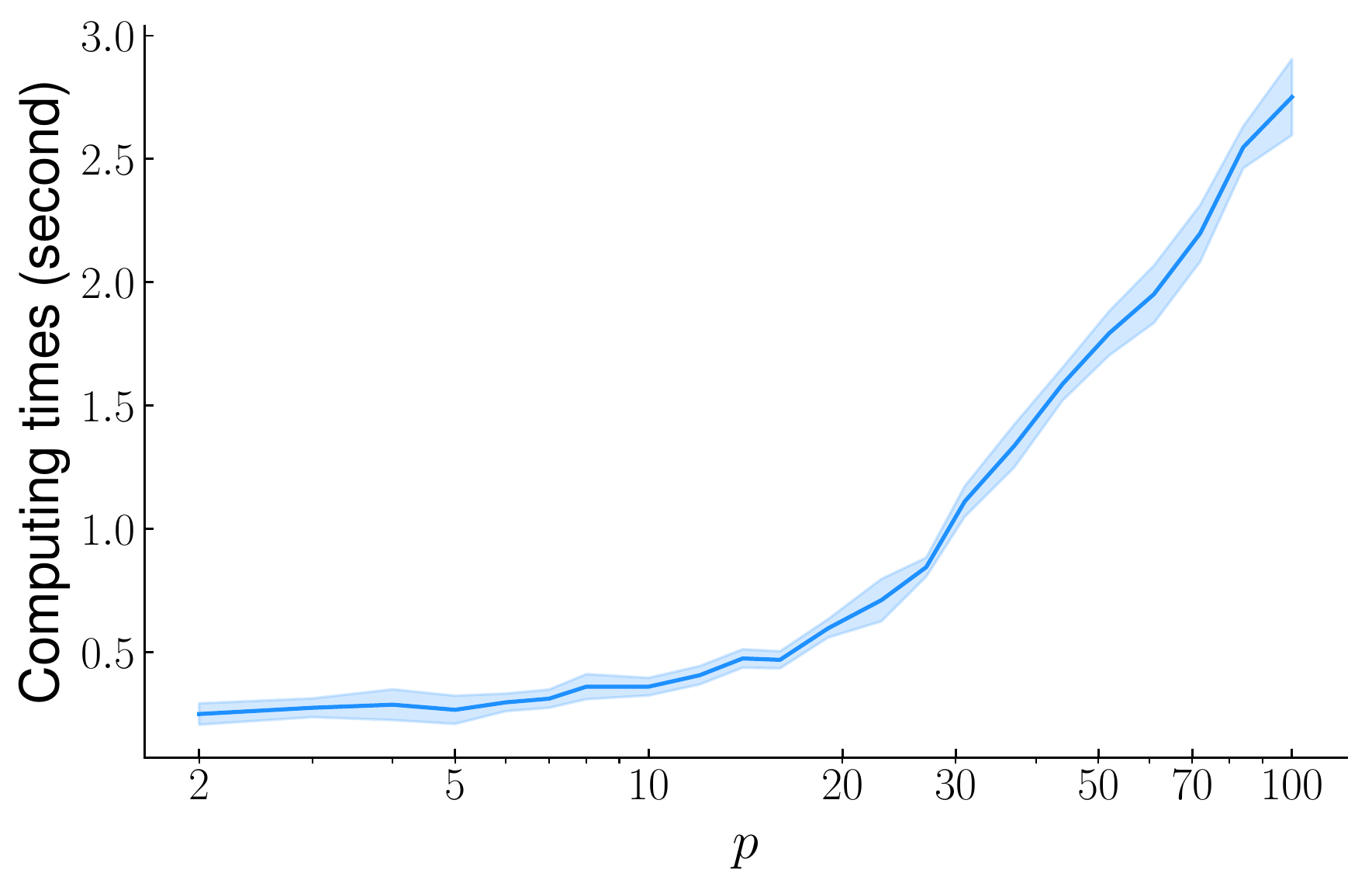}\label{fig:computing-times-binacox}}\\
\caption{Computing time for the methods considered.}
\label{3figs}
\end{figure}

Now let us look at the computing time required for the  methods considered. As the multiple testing-related methods are univariate, we can directly parallelize their computations across dimensions (which is what we did in the applications), so let us consider here a single feature $X$ ($p=1$). Following the method explained in Section~\ref{sec:competing methods}, we have to compute all log-rank test $p$-values computed on the  populations $\{y_i : x_i > \mu\}$ and $\{y_i : x_i \leq \mu\}$ for $i = 1, \ldots, n$, for $\mu$ taking all $x_i$ values between the 10th and 90th empirical quantiles of $X$. We denote ``MT all'' this method in Figure~\ref{fig:computing-times-comparison}, and compare its computing times with the binacox for various values of $n$. We  also show the ``MT grid'' method that only computes the $p$-values for candidates $\mu_{j,l}$ used in the binacox method.

Since the number of candidates does not change with $n$ for the ``MT grid'' method, the computing time ratio between ``MT all'' and ``MT grid'' naturally increases, going roughly from one to two orders of magnitude higher when $n$ goes from 300 to 4000. Hence to make computations much faster, we will use the ``MT grid'' for all multiple testing-related methods in the following.
The resulting loss of precision in the MT-related methods is negligible for a high enough $d_j$ ($= 50$ in practice).

Next, we emphasize the fact that the binacox is still roughly 5 times faster than the ``MT grid'' method, and it remains very fast when we increase the dimension, as shown in Figure~\ref{fig:computing-times-binacox}. It turns out that the computational time grows roughly logarithmically with $p$.

\subsubsection{Performance comparison.}

\begin{figure}[!htb]
\centering
\includegraphics[width=\textwidth]{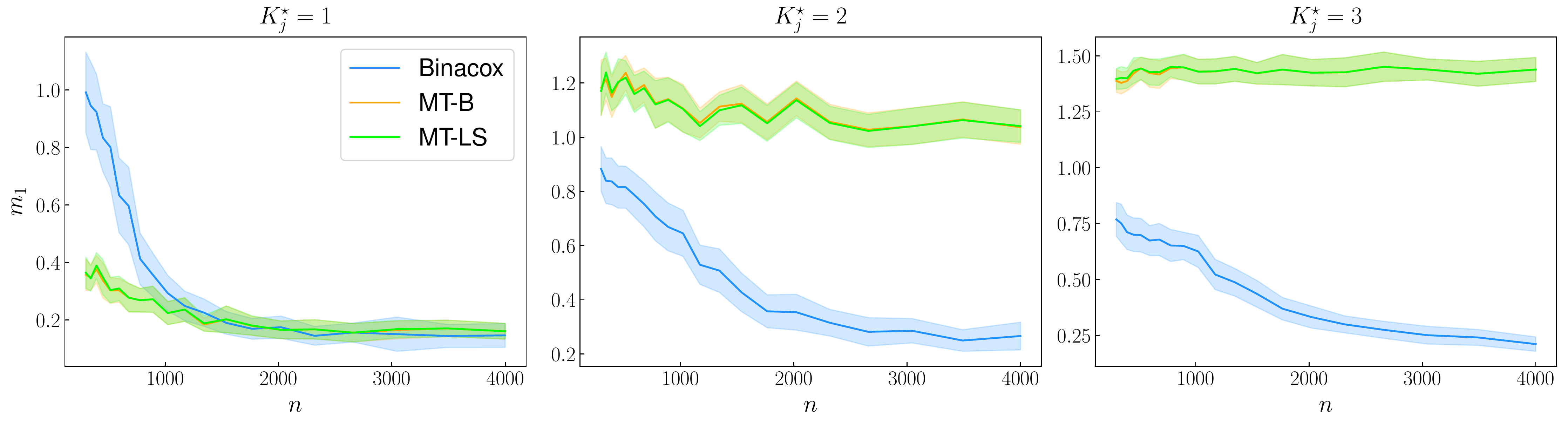}\hfill
\caption{Average (bold) $m_1$ scores and standard deviation (bands) obtained on 100  datasets simulated according to Section~\ref{sec:simulation} with $p = 50$ and $K^\star_j$ equal to 1, 2 and 3 (for all $j = 1, \ldots, p$) for the left, center and right sub-figures respectively) for varying $n$. The lower the value of $m_1$, the better the result; the binacox clearly outperforms the other methods when there is more than one cut-point, and is competitive with other methods when there is only one cut-point, but performs worse when $n$ is small because it overestimates $K^\star_j$.}
\label{fig:m_1}
\end{figure}

\begin{figure}[!htb]
\centering
\includegraphics[width=.75\textwidth]{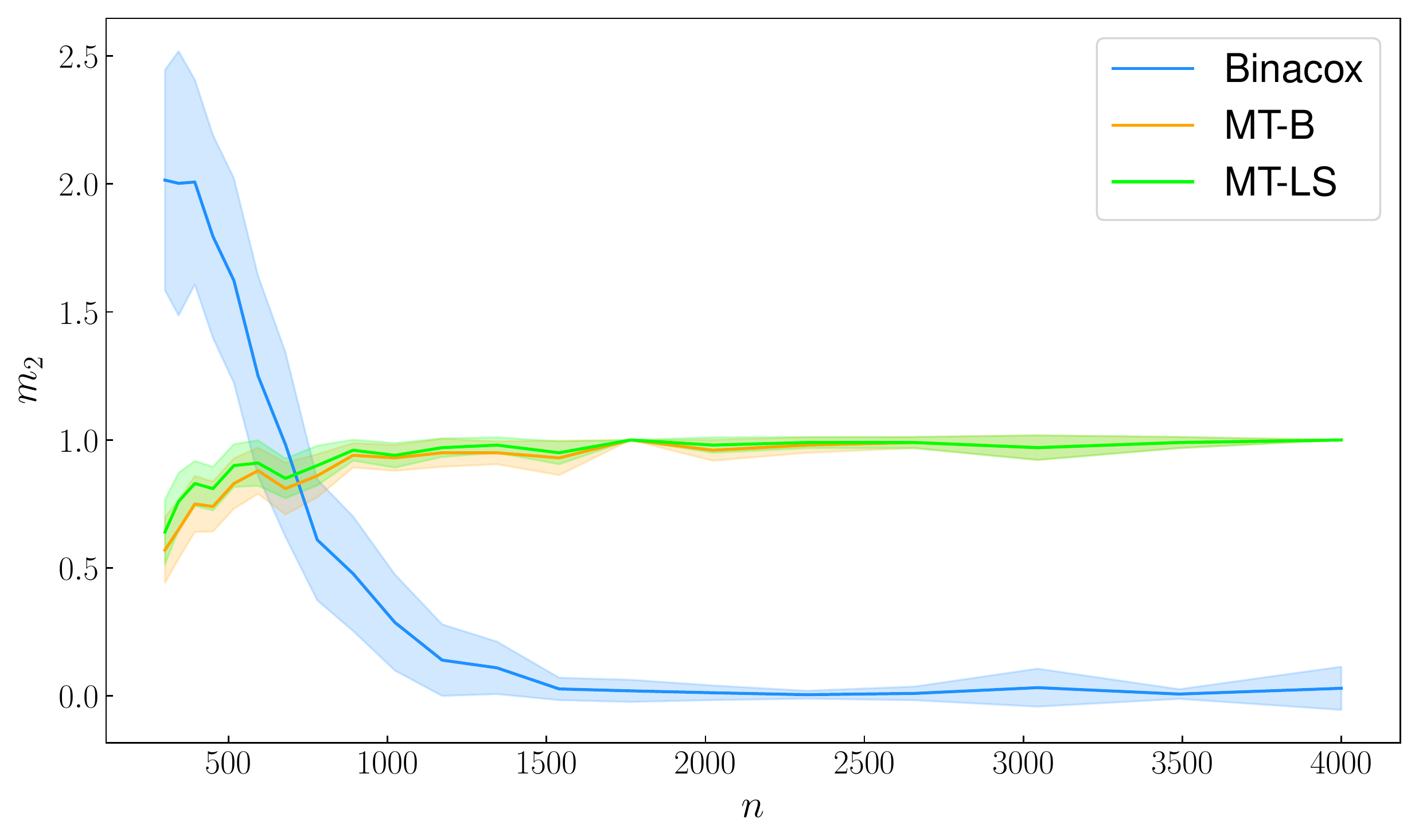}
\caption{Average (bold) $m_2$ scores and standard deviation (bands) obtained on 100  datasets simulated according to Section~\ref{sec:simulation} with $p = 50$ for varying $n$.  MT-B and MT-LS tend to detect a cut-point when there is none (no matter the value of $n$), while binacox overestimates the number of cut-points for small values of $n$ but detects  $\cS$ well for $p=50$ on the simulated data when $n>1000$.}
\label{fig:m_2}
\end{figure}

Let us compare now the results of simulations in terms of the $m_1$ and $m_2$ metrics introduced in Section~\ref{sec:simulation}. Figure~\ref{fig:m_1} gives a comparison of the  methods considered for the cut-point estimation aspect, i.e., in terms of the $m_1$ score. It appears that the binacox outperforms the MT-related methods when $K^\star_j>1$, and is competitive when $K^\star_j=1$ except for small values of $n$. This is due to an overestimation in the number of cut-points by the binacox (see Figure~\ref{fig:m_2}), especially when $p$ is high and $n$ is small, which gives higher $m_1$ values, even if the ``true'' cut-point is actually well-estimated.
Note that for such values of $p$, the binacox runs much faster than the MT-related methods.

Figure~\ref{fig:m_2}, on the other hand, assesses the ability of each method to detect features with no cut-points using the $m_2$ metric, i.e., the ability to estimate $\hat K^\star_j = 0$ for $j \in \cS$. The binacox appears to be quite effective at detecting features with no cut-point when $n$ takes a high enough value compared to $p$, which is not the case for the MT-related methods.

\section{Application on genetic data}
\label{sec:application}

In this section, we apply our method to three biomedical datasets. We extracted normalized expression data and survival times $Z$ in days from breast invasive carcinoma (BRCA, $n=1211$), glioblastoma multiforme (GBM, $n=168$) and kidney renal clear cell carcinoma (KIRC, $n=605$). 
These datasets are available on \emph{The Cancer Genome Atlas} (TCGA) platform, which aims to accelerate the understanding of the molecular basis of cancer with the help of genomic technology, including large-scale genome sequencing. 
For each patient, 20,531 features corresponding to normalized gene expression values are available. 

As we saw in Section~\ref{sec:results-simulation}, the MT-related methods are intractable in such high-dimensional cases.
We therefore include a screening step to select the portion of features most relevant to our problem from the 20,531 available.
To do so, we fit the binacox on each $j$th block separately and  take the resulting $\norm{\hat\beta_{j,\bullet}}_{\TV}$ as a score that roughly assess the propensity for feature $j$ to have one (or more) relevant cut-point(s). We then select the features corresponding to the top $P$ values with $P=50$, this choice being suggested by the distribution of the obtained scores given in Figure~\ref{fig:screening} of Appendix~\ref{sec:screening}. 

\subsection{Estimation results.}
\begin{figure}[!htb]
\centering  
\includegraphics[width=1\columnwidth,valign=c]{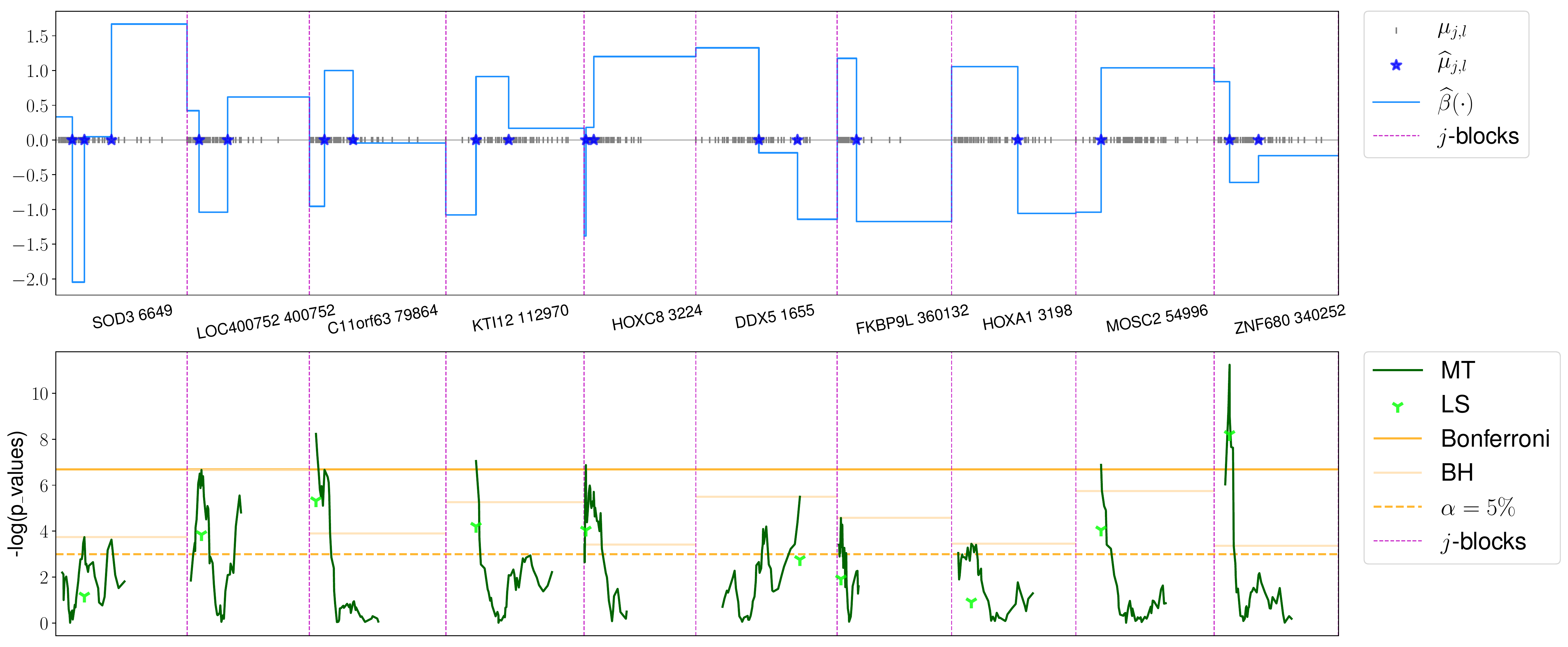}
\caption{Illustration of the results obtained on the top 10 features ordered according to the binacox $\norm{\hat\beta_{j,\bullet}}_{\TV}$ values on the GBM dataset. The binacox detects multiple cut-points and sheds light on non-linear effects for various genes. The BH thresholds are shown, but are unusable in practice.}
\label{fig:gbm-results}
\end{figure}

In Figure~\ref{fig:gbm-results} we present the results obtained by the  methods considered on the GBM cancer dataset for the top 10  features ordered according to the binacox $\norm{\hat\beta_{j,\bullet}}_{\TV}$ values. We observe that all cut-points detected by the univariate multiple testing methods with Bonferroni (MT-B) or Lausen and Schumacher (MT-LS) corrections are also detected by the multivariate binacox (which detects more cut-points); see Table~\ref{table:GBM cut-points}. The binacox identifies many more cut-points than the univariate MT-B and MT-LS methods. Further, all cut-points detected by these two methods are also detected by the binacox. Furthermore, it turns out that these top 10 genes (from the original 20,531) are quite relevant to GBM, the  most aggressive cancer that begins in the brain. 

For instance, the first gene, SOD3, is relevant from a physiopathological point of view since its polymorphisms are already known as GBM risk factors~\citep{rajaraman2008oxidative}.
Other genes in the top 10 (C11orf63 or the HOX genes) are also known to be directly related to brain development~\citep{canu2009hoxa1}, and are already known as potential GBM prognosis marker~\citep{duan2015hoxa13,guan2019overexpression}.
\begin{table}
\caption[\textwidth]{Estimated cut-point values for each method on the top 10  genes presented in Figure~\ref{fig:gbm-results} for GBM. Dots ($\cdot$) mean ``no cut-point detected''.\label{table:GBM cut-points}}
\centering
\begin{tabular}{cccc}
\toprule
Genes & Binacox & MT-B & MT-LS \\
\midrule
    SOD3 6649     &  200.87, 326.40, 606.48  &  $\cdot$   &  $\cdot$  \\
 LOC 400752       &     31.46, 62.50         &  $\cdot$   &   34.04  \\
  C11orf63 79864  &     40.30, 109.67        &   19.65    &   19.65  \\
   KTI12 112970   &     219.60, 305.70       &   219.60   &   219.60  \\
    HOXC8 3224    &     3.30, 15.75          &   3.30     &   3.30  \\
    DDX5 1655     &      10630.11, 13094.89  &  $\cdot$   &  $\cdot$  \\
  FKBP9L 360132   &        111.72            &  $\cdot$   &  $\cdot$  \\
    HOXA1 3198    &        67.28             &  $\cdot$   &  $\cdot$  \\
   MOSC2 54996    &        107.53            &   107.53   &   107.53  \\
  ZNF680 340252   &     385.85, 638.06       &   385.85   &   385.85  \\
\bottomrule
\end{tabular}
\end{table}

Relevant results were also obtained on the KIRC and BRCA datasets; these are postponed to Appendix~\ref{sec:KIRC and BRCA results}.

\subsection{Risk prediction.}
Let us now investigate how performances are impacted in terms of risk prediction when detected cut-points are taken into account; namely, comparing predictions when training a Cox model on the original continuous feature space versus on the $\widehat\mu$-binarized space constructed with the cut-point estimates. 

In a classical Cox model, $R_i=\exp(X_i^\top \hat \beta)$ is known as the predicted risk for patient $i$ measured at $t = 0$.
A common metric to evaluate risk prediction performances in this type of survival setting is the C-index~\citep{heagerty2005survival}, which is defined by
\begin{equation*}
  \cC_\tau =\bP[R_i > R_j | Z_i < Z_j , Z_i < \tau],
\end{equation*}
with $i \neq j$ two independent patients and $\tau$ the follow-up period. A Kaplan-Meier estimator for the censoring distribution leads to a nonparametric and consistent estimator of $\cC_\tau$ \citep{uno2011c}, which is already implemented in the \texttt{python} package \texttt{lifelines}.

We randomly split the three datasets 100 times into training and validation sets (30\% for testing) and compare the average C-index on the validation sets in Table~\ref{table:C-index tcga} when the $\widehat\mu$-binarized space is constructed based on the $\widehat\mu$'s obtained either from the binacox, MT-B, or MT-LS. 
We also compare performances obtained by two nonlinear multivariate methods known to perform well in high-dimensional settings:
boosted Cox (CoxBoost)~\citep{li2005boosting} used with 300 boosting steps (this number being fine-tuned by cross-validation), and random survival forests (RSF)~\citep{ishwaran2008random} used with 200 trees (also cross-validated), respectively implemented in the \texttt{R} packages \texttt{CoxBoost} and \texttt{randomForestSRC}. Note that for a fair comparison, and to avoid selection bias~\citep{ambroise2002selection}, the screening step is re-run on each training set, using the C-index obtained by
univariate Cox models (not to confer advantage to our method), namely Cox PH models fitted on each covariate separately.

\begin{table}[!htb]
\label{table:C-index tcga}
\caption{Comparison of average C-indexes (and standard deviation in parentheses) on 100 random train/test splits for the Cox model trained on continuous features versus on its binarized version constructed using the considered methods' cut-point estimates, and the CoxBoost and RSF methods. On the three datasets, the binacox method gives by far the best results (in bold).}
\centering
\resizebox{\textwidth}{!}{
\begin{tabular}{ccccccc}
\toprule
Cancer & Continuous & Binacox & MT-B & MT-LS & CoxBoost & RSF \\
\midrule
 GBM & 0.563 (0.037) & \textbf{0.603 (0.048)} & 0.579 (0.049) & 0.577 (0.043) & 0.569 (0.037) & 0.564 (0.036) \\
 KIRC & 0.675 (0.028) & \textbf{0.709 (0.022)} & 0.682 (0.022) & 0.682 (0.022) & 0.683 (0.029) & 0.695 (0.026) \\
 BRCA & 0.592 (0.050) & \textbf{0.669 (0.047)} & 0.626 (0.055) & 0.621 (0.061) & 0.598 (0.053) & 0.659 (0.037) \\
\bottomrule
\end{tabular}}
\end{table}
\begin{figure}[!htb]
\centering  
\includegraphics[width=\textwidth,valign=c]{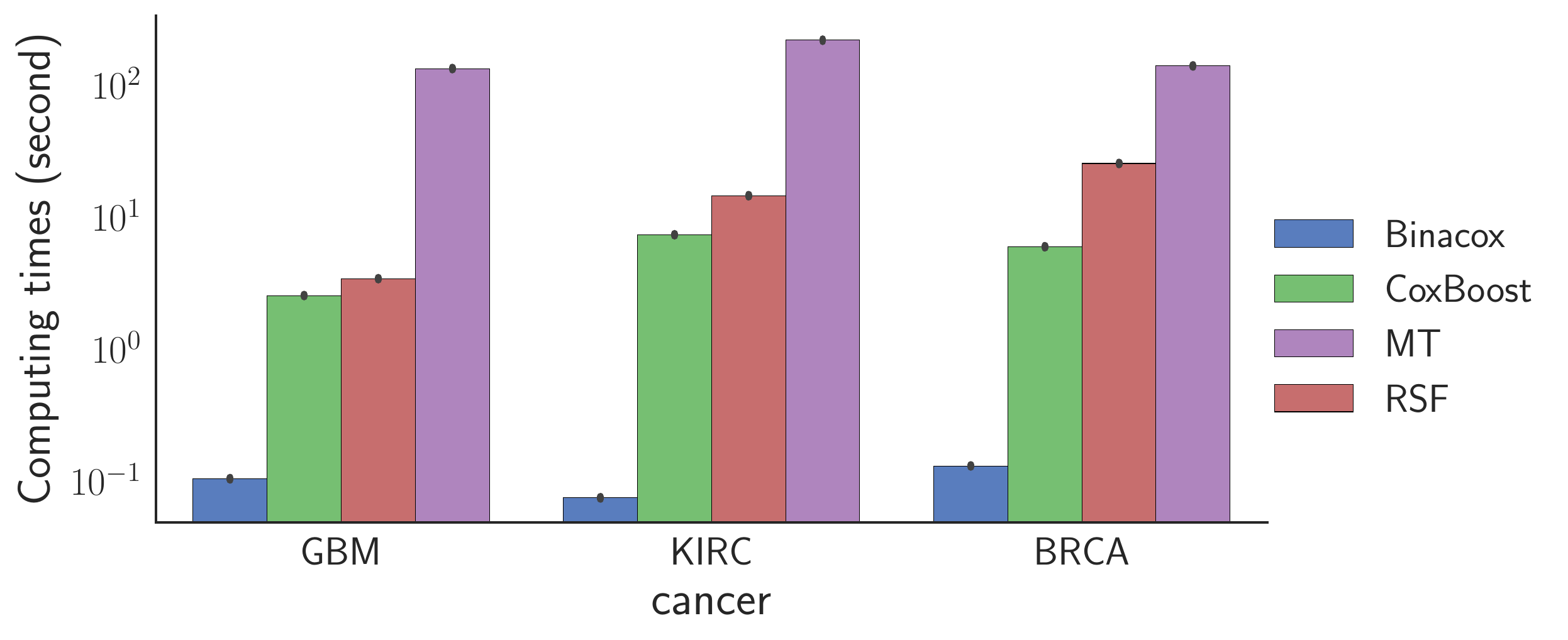}
\caption{Average computing times (in seconds) required by  each method on the three datasets (with the black lines representing $\pm$ the standard deviation) obtained on 100 random train/test split. The binacox method is at least one and up to several orders of magnitude faster.}
\label{fig:computing-times-tcga}
\end{figure}

The binacox method clearly improves risk prediction compare to classical Cox, as well as with respect to the MT-B and MT-LS methods. Moreover, it also significantly outperforms both CoxBoost and RSF. 
To the best of our knowledge, no better performances have been achieved on this data in the literature~\citep{yousefi2017predicting}.
Figure~\ref{fig:computing-times-tcga} compares the computing times of the methods. Clearly the binacox is by far the most computationally efficient.

\section{Conclusion}\label{sec:conclusion}

In this paper, we introduced the binacox method, designed for estimating multiple cut-points in a Cox model with high-dimensional features.
We illustrated the good theoretical properties of the model by establishing nonasymptotic oracle inequalities for prediction and estimation.
An extensive Monte Carlo simulation study was then carried out to evaluate the method's performance. It showed that our approach outperforms existing methods, with computing times orders of magnitude faster. Moreover, in addition to the raw feature selection ability of the binacox, it succeeds in detecting multiple cut-points per feature.
We also applied the binacox to three publicly available high-dimensional genetics datasets. Furthermore, several genes pinpointed by the model turn out to be biologically relevant (e.g., the gene SOD3 for GBM), whilst others require further investigation in the genetics research community.
More importantly, our method provides powerful interpretation aspects that could be useful in both clinical research and daily practice. Indeed, the estimated cut-points could be directly considered in clinical practice. 
Thus, the method could be an interesting alternative to more classical methods found in the medical literature to deal with prognosis studies in high-dimensional frameworks, providing a new way to model nonlinear feature associations, and giving rise to new data-driven risk scores. 
Our study lays the groundwork for the development of powerful methods which could one day help provide improved personalized care.

\section*{Acknowledgments}
Mokhtar Z. Alaya is grateful for a grant from DIM Math Innov Région Ile-de-France \url{http://www.dim-mathinnov.fr}. Agathe Guilloux's work has been supported by the INCA-DGOS grant PTR-K 2014.
The results shown in this paper are based upon data generated by the TCGA Research Network and freely available from \url{http://cancergenome.nih.gov}. 
\textit{Conflict of Interest}: None declared.

\section*{Software}
All methodology discussed in the paper is implemented in \texttt{Python/C++} and \texttt{R}. The code that generates all figures is available from \url{https://github.com/SimonBussy/binacox} in the form of annotated programs, together with notebook tutorials.

\newpage

\begin{appendices}

\section{Additional details}

\subsection{Algorithm.}
\label{sec:Algorithm}

To solve regularization problem \eqref{problem-estim}, we first look at the proximal operator of the binarsity penalty~\citep{alaya2016}.  
It turns out that it can be computed very efficiently, using an algorithm introduced in~\citet{Cond-13} that we modify in order to include the weights $\omega_{j, k}$. It basically applies -- in each block -- the proximal operator of the total variation (since the binarsity penalty is block separable), followed by a centering within each block to satisfy the constraint, see Algorithm~\ref{algorithm-primal-computation} below. We refer to~\citet{alaya2014} for the weighted total variation proximal operator.
\begin{algorithm}[]
   \caption{Proximal operator of $\bina(\beta)$, see~\citep{alaya2016}}
   \label{algorithm-primal-computation}
\begin{algorithmic}
   \STATE {\bfseries Input:}  vector $\beta \in \mathscr{B}_{p+d}(R)$ and weights $\omega_{j, l}$ for $j=1, \ldots, p$ and $l=1, \ldots, d_j+1$   
   \STATE {\bfseries Output:} vector $\eta = \prox_{\bina}(\beta)$   
   \FOR{$j=1$ {\bfseries to} $p$}   
   \STATE $\theta_{j,\bullet}  \gets \prox_{\norm{\cdot}_{\TV,\omega_{j,\bullet}}}(\beta_{j, \bullet})$ (TV-weighted in block $j$, see~\eqref{eq:tv_no_linear_constraint})
   \STATE $\eta_{j,\bullet} \gets  \theta_{j,\bullet} - \frac{n_{j,\bullet}^\top \theta_{j,\bullet}}{\norm{n_{j,\bullet}}_2^2} n_{j,\bullet}$ (projection onto $\text{span}(n_{j,\bullet})^\perp$)
   \ENDFOR
   \STATE {\bfseries Return:} $\eta$
\end{algorithmic}
\end{algorithm} 

\subsection{Implementation}
\label{sec:Implementation}

\begin{figure}[!htb]
\centering
\includegraphics[width=.7\columnwidth,valign=c]{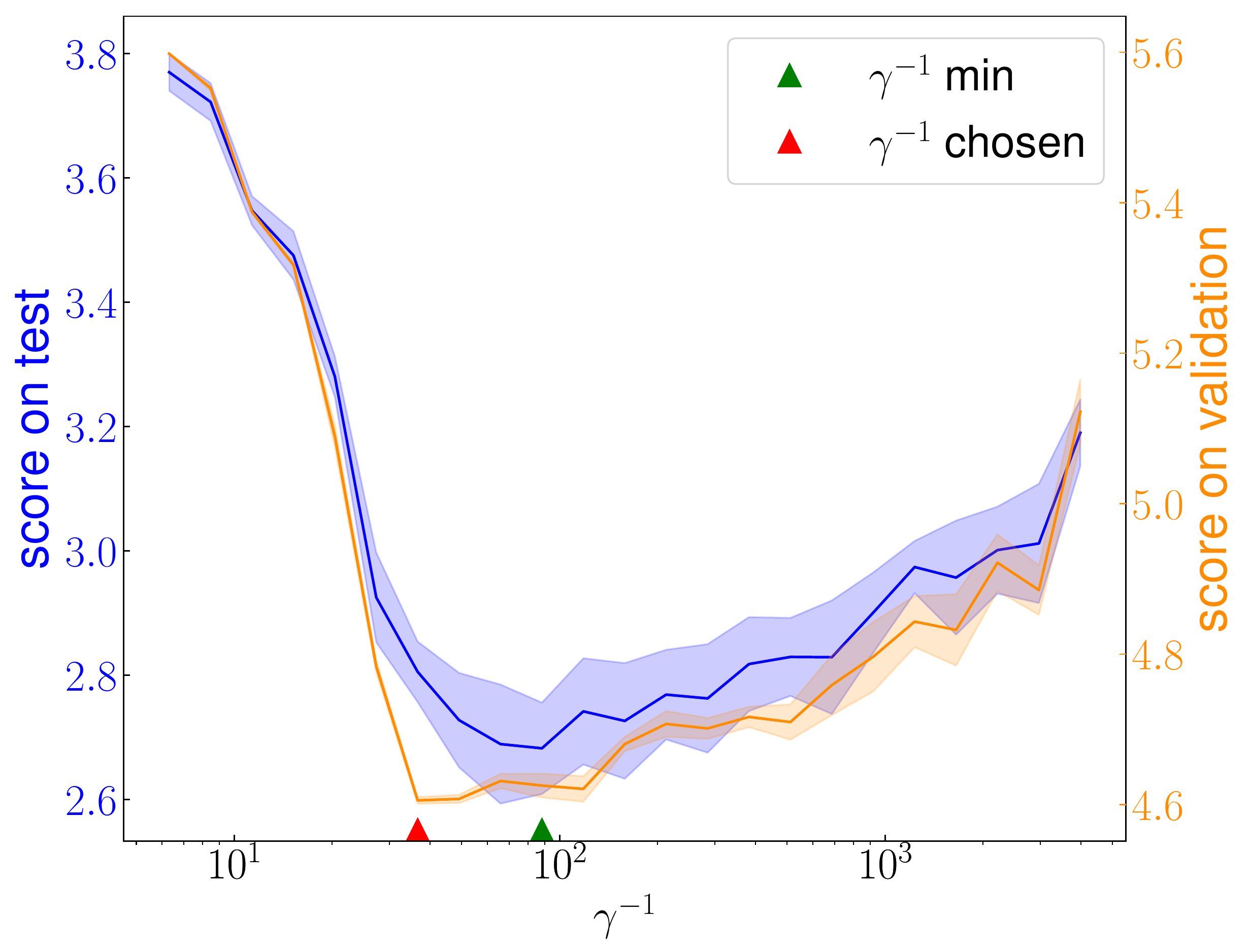}
\caption{Learning curves obtained for various $\gamma$, in blue on the different test sets during cross-validation, and in orange on the validation set. Bold lines represent average scores on the folds, and bands represent 95\% Gaussian confidence intervals. The green triangle points out the value of $\gamma^{-1}$ that gives the minimum score (best training score), while the $\gamma^{-1}$ value we automatically select (the red triangle) is the smallest value such that the score is within one standard error of the minimum, which is a classical trick~\citep{simon2011regularization} that favors a slightly higher penalty strength (smaller $\gamma^{-1}$) to avoid  over-estimation of $K^\star$ in our case.}
\label{fig:cv}
\end{figure}

Figure~\ref{fig:cv} gives the learning curves obtained during the $V$-fold cross-validation procedure presented in Section~\ref{sec:competing methods} with $V = 10$ for the fine-tuning of parameter $\gamma$, which is the strength of the binarsity penalty. 
We randomly split the data into training and validation sets (30\% for validation, cross-validation being done on the training).
Recall that the score we use is the negative partial log-likelihood defined in~\eqref{binarized-model} computed after a refit of the model on the binary space obtained by the estimated cut-points, with the sum-to-zero constraint in each block but without the TV penalty.

\begin{figure}[!htb]
\centering  
\includegraphics[width=1\columnwidth,valign=c]{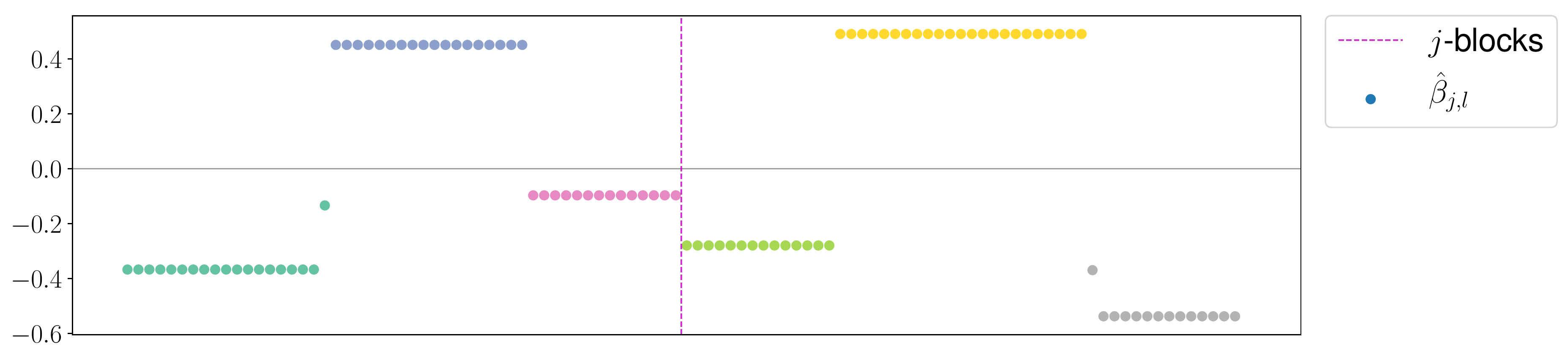}
\caption{Illustration of the de-noising step in the cut-point detection phase on the simulated date of Figure~\ref{fig:example-data}. Within each block (separated with the dotted pink line), the different colors represent $\hat \beta_{j,l}$ with corresponding $\mu_{j,l}$ in distinct estimated $I^\star_{j,k}$. The following rule is applied: when a $\hat \beta_{j,l}$ is ``isolated'', it is assigned to its ``closest'' group.}
\label{fig:groups}
\end{figure}

Figure~\ref{fig:groups} illustrates the de-noising step for the cut-point detection when looking at the $\hat \beta$ support relative to the TV norm. The $\hat \beta$ vector plotted here corresponds to the data generated in Figure~\ref{fig:example-data} of Section~\ref{sec:simulation},  where the final estimation results were presented in Figure~\ref{fig:estimation} of Section~\ref{sec:results-simulation}.
Since it is usual to observe three consecutive $\hat \beta$'s jumps in the neighbourhood of a true cut-point, which is the case in Figure~\ref{fig:groups} for the first and the last jumps, this could lead to an over-estimation of $K^\star$. 
To bypass this problem, we then use the following rule: if $\hat \beta$ has three consecutive different coefficients within a block, then only the largest jump is considered as a ``true'' one.  

\subsection{TCGA gene screening}
\label{sec:screening}

\begin{figure}[!htb]
\centering  
\includegraphics[width=1\columnwidth,valign=c]{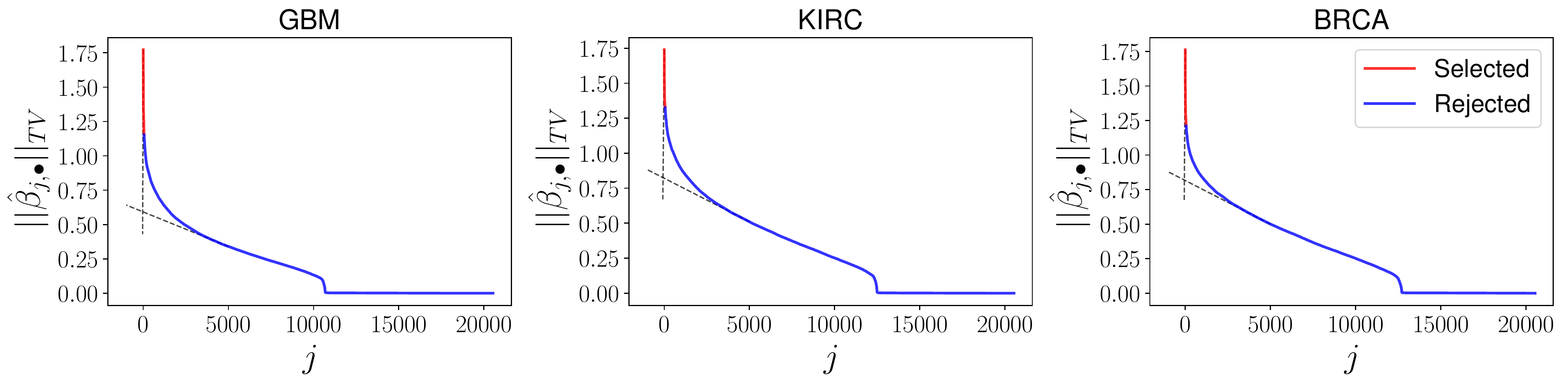}
\caption{$\norm{\hat\beta_{j,\bullet}}_{\TV}$ obtained for univariate binacox fits for the three datasets considered. The top $P$ selected features appear in red, and it turns out that taking $P = 50$ coincides with the elbow (represented with the dotted grey lines) in each of the three curves.}
\label{fig:screening}
\end{figure}

Figure~\ref{fig:screening} illustrates the screening procedure followed to reduce the high-dimensionality of the TCGA datasets to make the multiple testing related methods tractable. 
We then fit a univariate binacox on each block $j$ separately and compute the resulting $\norm{\hat\beta_{j,\bullet}}_{\TV}$ to assess the propensity for feature $j$ to obtain one (or more) relevant cut-point(s). It appears that taking the top $P$ features with $P = 50$ is a reasonable choice for each dataset considered.

\subsection{Results on BRCA and KIRC data}
\label{sec:KIRC and BRCA results}

\begin{figure}[!htb]
\centering  
\includegraphics[width=1\columnwidth,valign=c]{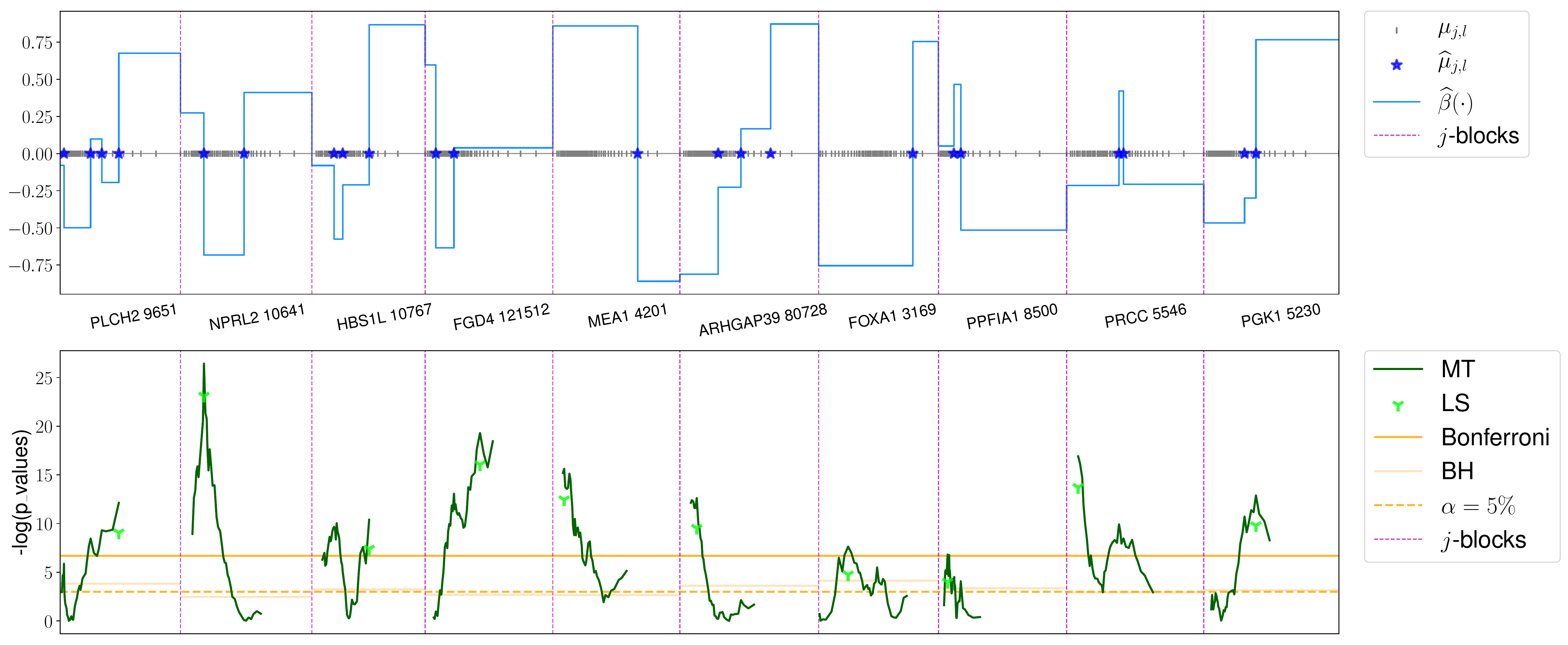}
\caption{Illustration of the results obtained on the top 10 features ordered according to the binacox $\norm{\hat\beta_{j,\bullet}}_{\TV}$ values on the BRCA dataset.}
\label{fig:brca-results}
\end{figure}

Figure~\ref{fig:brca-results} illustrates the results obtained by all  methods we consider on the BRCA cancer dataset for the top 10  features ordered according to the binacox $\norm{\hat\beta_{j,\bullet}}_{\TV}$ values. Table~\ref{table:BRCA cut-points} summarizes the detected cut-point values for each method.
It turns out that the selected genes are quite relevant from a clinical point of view (for instance, NPRL2 is a tumor suppressor gene~\citep{huang2016downregulation}), and in  particular for BRCA (breast) cancer. For instance, HBS1L expression is known for being predictive of breast cancer survival~\citep{antonov2014ppisurv,antonov2011bioprofiling,HBS1L}, while FOXA1 and PPFIA1 are highly related to breast cancer, see~\citet{badve2007foxa1} and~\citet{dancau2010ppfia1} respectively.

\begin{table}
\caption{Estimated cut-point values for each method on the top 10 genes presented in Figure~\ref{fig:brca-results} for BRCA.\label{table:BRCA cut-points}}
\centering
\begin{tabular}{cccc}
\toprule
Genes & Binacox & MT-B & MT-LS \\
\midrule
PLCH2 9651  & 28.43, 200.74, 273.04, 382.87 & 382.87   & 382.87   \\
NPRL2 10641 & 330.64, 568.06                & 330.64   & 330.64   \\
HBS1L 10767 &  1023.91, 1212.54, 1782.77    & 1782.77  & 1782.77  \\
FGD4 121512 & 163.59, 309.24                &  517.90  & 517.90   \\
MEA1 4201   &  2199.21                      & 786.29   & 786.29   \\
ARHGAP39 80728 & 493.01, 734.37, 1049.04    & 265.26   & 265.26   \\
FOXA1 3169  & 11442.32                      & 3586.03  & 3586.03  \\
PPFIA1 8500 & 1500.02, 1885.27              & 1152.98  & 1152.98  \\
PRCC 5546   & 2091.16, 2194.08              & 1165.49  & 1165.49  \\
PGK1 5230  & 10205.72, 12036.29             & 12036.29 & 12036.29 \\
\bottomrule
\end{tabular}
\end{table}

\begin{figure}[!htb]
\centering  
\includegraphics[width=1\columnwidth,valign=c]{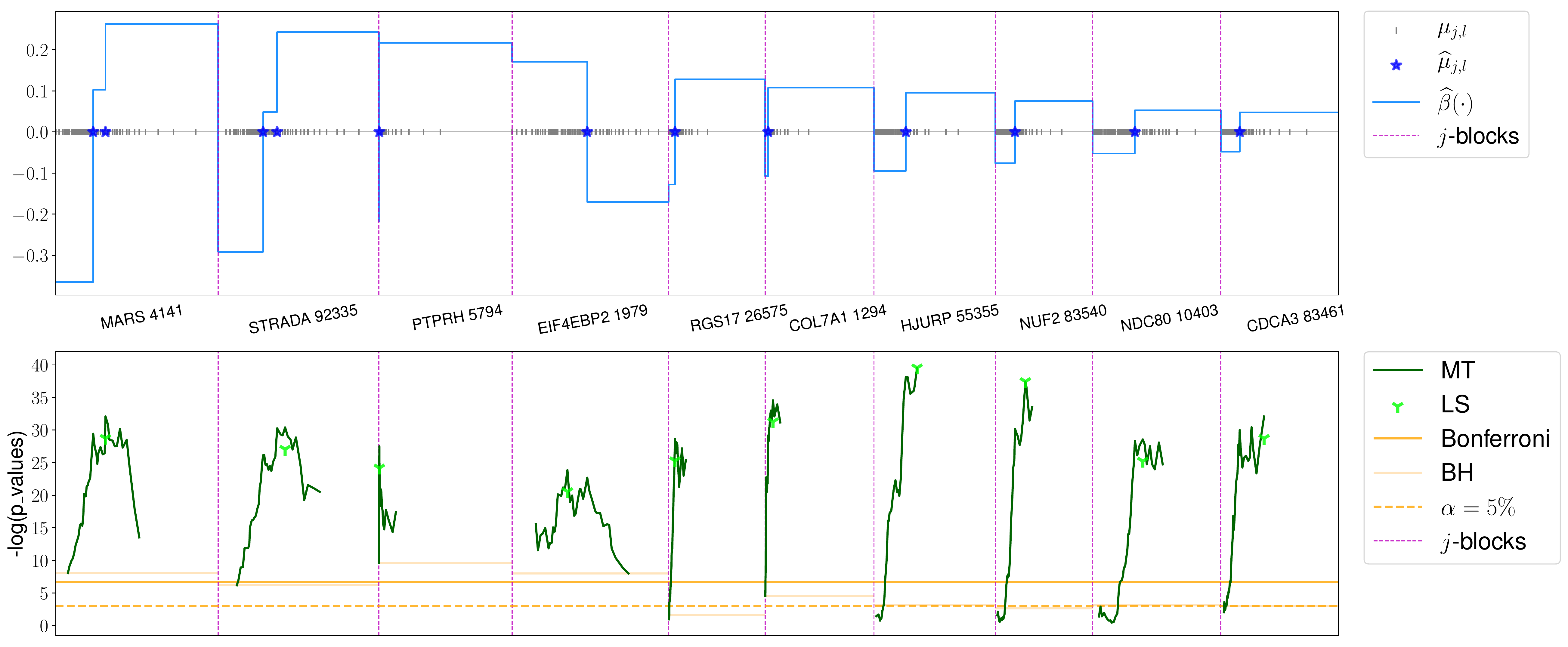}
\caption{Illustration of the results obtained on the top 10 features ordered according to the binacox $\norm{\hat\beta_{j,\bullet}}_{\TV}$ values on the KIRC dataset.}
\label{fig:kirc-results}
\end{figure}

Lastly, Figure~\ref{fig:kirc-results} gives the results obtained by the various methods on the KIRC cancer dataset for the top 10 features ordered according to the binacox $\norm{\hat\beta_{j,\bullet}}_{\TV}$ values, and Table~\ref{table:KIRC cut-points} summarizes the detected cut-point values for each method.
Once again, the selected genes are relevant for cancer studies including KIRC. For instance, EIF4EBP2 is related to cancer proliferation~\citep{mizutani2016oncofetal}), RGS17 is known to be overexpressed in various cancers~\citep{james2009rgs17}, and both COL7A1 and NUF2 are known to be related to renal cell carcinoma (see~\citep{csikos2003dystrophic} and~\citep{kulkarni2012cancer} respectively). Moreover, the first two genes MARS 4141 and STRADA 92335 already appear as relevant KIRC prognosis markers in~\citet{bussy2019c} .
\begin{table}
\caption{Estimated cut-point values for each method on the top 10 genes illustrated in Figure~\ref{fig:kirc-results} for KIRC.\label{table:KIRC cut-points}}
\centering
\begin{tabular}{cccc}
\toprule
Genes & Binacox & MT-B & MT-LS \\
\midrule
MARS 4141      & 1196.21, 1350.00  &  1350.00   &  1350.00   \\
STRADA 92335   & 495.24, 553.73    &  586.88    &  586.88   \\
PTPRH 5794     &   3.32            &  3.32      &  3.32  \\
EIF4EBP2 1979  &  6504.80          &  5455.59   &  5455.59  \\
RGS17 26575    &   4.30            &  4.30      &  4.30  \\
COL7A1 1294    &   44.19           &  113.08    &  113.08  \\
HJURP 55355    &   99.83           &  134.31    &  134.31   \\
NUF2 83540     &   42.18           &  63.09     &  63.09   \\
NDC80 10403    &   91.39           &  107.53    &  107.53   \\
CDCA3 83461    &   52.03           &  110.18    &  110.18   \\
\bottomrule
\end{tabular}
\end{table}

\section{Proof of Theorem~\ref{thm-faster-in-approx}}
In this section, we provide the proof of Theorem~\ref{thm-faster-in-approx}.
First, we derive some preliminary results which will be required in the following. 

\subsection{Preliminary results}
\label{preliminaries-proofs}

\paragraph{Additional notation.}
For $u, v \in \R^m$, we denote by $u \odot v$ the Hadamard product defined by $u\odot v =(u_1v_1, \ldots, u_mv_m)^\top.$ 
We denote by $\sgn(u)$ the subdifferential of the function $u\mapsto |u|$, i.e., 
\begin{equation*}
  \sgn(u) =
  \begin{cases}
    \{ 1\} & \text{ if } u > 0, \\
    [-1, 1] & \text{ if } u = 0, \\
    \{ -1\} & \text{ if } u < 0.
  \end{cases}
\end{equation*}
We write $\partial(\phi)$ for the subdifferential mapping of a convex functional $\phi$.
We adopt in the proofs counting process notation. We then define the observed-failure counting process  $ N_i(t)=  \ind{}(Z_i \leq t, \Delta_i = 1),$ the at-risk process $Y_i(t) = \ind{}(Z_i\geq t),$ and $\bar{N}(t) = \dfrac 1n \sum_{i=1}^nN_i(t).$
For every vector $v$, let us denote $v^{\otimes0}= 1$, $v^{\otimes1}= v,$ and  $v^{\otimes2}=vv^\top$ (outer product). Recall finally that $\tau>0$ denotes the finite study duration.

\paragraph{Weights.}
For a given numerical constant $c>0$, the weights $\omega_{j,l}$ have an explicit form given by
\begin{equation}
\label{weights}
  \omega_{j,l} = 5.64\sqrt{\frac{c + \log(p+d)+ \scrL_{n,c}}{n} } + 18.62  \frac{(c + \log(p+d)+ 1+\scrL_{n,c})}{n},
\end{equation}
where $\scrL_{n,c} = 2 \log\log\big((2en + 24 ec)\vee e\big)$.

\paragraph{Properties of the binarsity penalty.}
We define $\omega = (\omega_{1,\bullet}, \ldots, \omega_{p,\bullet})$ the weights vector, with $\omega_{j,1}=0$ for all $j = 1, \ldots, p$.
Then, we rewrite the total variation part in the binarsity penalty as follows. Let us define the $(d_j+1) \times (d_j+1)$ matrix $D_j$ by
\begin{equation*}
  {D}_j=
  \begin{bmatrix}
  1&0& & 0\\
  -1&1&\\
  &  \ddots&\ddots\\
  0& &  -1&1
  \end{bmatrix} \in \R^{d_j+1}\times \R^{d_j+1}.
\end{equation*}
We then remark that for all $\beta_{j,\bullet} \in \R^{d_j+1}$, one has $\norm{\beta_{j,\bullet}}_{\TV, \omega_{j,\bullet}}  = \norm{\omega_{j,\bullet} \odot {D}_j \beta_{j,\bullet}}_1$.
Moreover, note that the matrix ${D}_j$ is invertible. 
We denote its inverse ${{T}_{j}}$, which is defined by the $(d_j+1)\times (d_j+1)$ lower triangular matrix with entries $({{T}_{j}})_{r,s} = 0$ if $r < s$ and $({{T}_{j}})_{r,s} = 1$ otherwise.
We set
\begin{equation}
  \label{differenceMat-D-inverse-T}
  {{\bf{D}}} =\diag({D}_{{1}}, \ldots, {D}_{{p}}) \quad \textrm{ and } \quad {{\bf{T}}} =\diag({T}_{{1}}, \ldots, {T}_{{p}}).
\end{equation}
Lemma~\ref{lemma-subadditive-Bina} then states that binarsity is a sub-additive penalty~\citep{kutateladze2013fundamentals}.
\begin{lemma}
\label{lemma-subadditive-Bina}
For all $\beta, \beta' \in \R^{p+d}$, we have that
\begin{equation*}
\bina(\beta+ \beta') \leq \bina(\beta) + \bina(\beta')
\quad  \text{and} \quad \bina(-\beta) \leq \bina(\beta).
\end{equation*}
\end{lemma}
\noindent{{\it Proof of Lemma~\ref{lemma-subadditive-Bina}.}}
The hyperplane $\text{span}\{u \in \R^{d_j+1}: n_{j,\bullet}^\top u = 0\}$ is a convex cone, then the indicator function $\delta_{j}$ is sublinear (i.e., positively homogeneous and sub-additive~\citep{kutateladze2013fundamentals}). Furthermore, the total variation penalization satisfies the triangle inequality, which gives the first statement of Lemma~\ref{lemma-subadditive-Bina}.
To prove the second, we use the fact that $\delta_{j}(\beta_{j,\bullet}) + \delta_{j}(- \beta_{j,\bullet}) \geq 0$ to obtain: 
\begin{equation*}
  \begin{split}
    \bina(-\beta) &= 
    \sum_{j=1}^p \Big(\norm{\beta_{j,\bullet}}_{\TV,\omega_{j,\bullet}} + \delta_{j}(- \beta_{j,\bullet})\Big)
    \leq \sum_{j=1}^p \Big(\norm{\beta_{j,\bullet}}_{\TV, \omega_{j,\bullet}} +  \delta_{j}(\beta_{j,\bullet}))\Big)=\bina(\beta),
  \end{split}
\end{equation*}
which concludes the proof of Lemma~\ref{lemma-subadditive-Bina}.
$\hfill\square$

\paragraph{Additional useful quantities.}
The Doob-Meyer decomposition~\citep{Aal-78} implies that, for all $i = 1, \ldots, n$ and all $t \geq 0$,
\begin{equation*}
\dd N_i(t) = Y_i(t) \lambda_0^\star(t) e^{f^\star(X_i)}\dd t + \dd M_i(t),
\end{equation*}
where the martingales $M_i$ are square integrable and orthogonal.
With this notation, we define, for all $t\geq 0$ and any $f$, the process
\begin{equation*}
  S_n^{(r)}(f, t) = \sum_{i=1}^n Y_i(t) e^{f(X_i)}(X_i^B)^{\otimes r}
\end{equation*}
for $r\in \{0, 1, 2\}$,  where $X_i^B$ is the $i$th row of the binarized matrix $\bX^B$.
The empirical loss $\ell_n$ can then be rewritten as
\begin{align*}
\ell_n(f) = - \frac{1}{n} \sum_{i=1}^n \int_0^\tau  \big\{f(X_i) 
 - \log \big( S_n^{(0)}(f, t) \big) 
\big\}\dd N_i(t).
\end{align*}
Together with this loss, we introduce the loss
\begin{align*}
\ell(f) &= - \frac{1}{n} \sum_{i=1}^n \int_0^\tau \big\{f(X_i) 
 - \log \big( S_n^{(0)}(f, t) \big) 
\big\}Y_i(t) \lambda_0^\star(t) e^{f^\star(X_i)} \dd t \nonumber \\
&= -\frac{1}{n} \sum_{i=1}^n \int_0^\tau  \log \Big(\dfrac{e^{f(X_i)}}{S_n^{(0)}(f, t)} \Big) Y_i(t) \lambda_0^\star(t) e^{f^\star(X_i)} \dd t.
\end{align*}
We will use the fact that for a function $f_\beta$ of the form
$f_\beta(X_i) = \beta^\top{X_i^B} = \sum_{j=1}^p f_{\beta_{j,\bullet}}(X_i)$, the Doob-Meyer decomposition implies that
\begin{align}
\label{eqn:gradients}
\nabla \ell_n(f_\beta) &= - \frac{1}{n} \sum_{i=1}^n \int_0^\tau  \Big\{X_i^B - \frac{S_n^{(1)}(f_\beta, t)}{S_n^{(0)}(f_\beta, t)}
\Big\}\dd N_i(t) \nonumber \\
&= \nabla \ell(f_\beta) + H_n(f_\beta),
\end{align}
where $H_n(f_\beta)$ is an error term defined by
\begin{equation}
\label{eqn:procH_n}
H_n(f_\beta)= - \frac{1}{n} \sum_{i=1}^n \int_0^\tau  \Big\{ X_i^B - \dfrac{S_n^{(1)}(f_\beta,t)}{S_n^{(0)}(f_\beta,t)} \Big\}\dd M_i(t).
\end{equation}
We also introduce the empirical $\ell_2$-norm defined for any function $f$ as
\begin{equation}
\label{eq:empirical-l2-norm}
\norm{f}_{n}^2 = \int_0^\tau \sum_{i=1}^n \big(f(X_i) - \bar{f}(t)\big)^2\frac{Y_i(t)e^{f^\star(X_i)}}{S_n^{(0)}(f^\star,t)} \dd \bar{N}(t),
\end{equation}
with 
\begin{equation*}
\bar{f}(t) = \sum_{i=1}^n \dfrac{Y_i(t) e^{f^\star(X_i)}}{S_n^{(0)}(f^\star,t)} f(X_i).
\end{equation*}

In the following section, we state some lemmas required for proving  our theorems. Their proofs are postponed to Section~\ref{proof-lemmas}.

\subsection{Lemmas}\label{sec:lemmas}

First, Lemma~\ref{lemma:KKT_monotonicy} is a consequence of the Karush-Kuhn-Tucker (KKT) optimality conditions~\citep{boyd2004convex} for a convex optimization and the monotony of subdifferential mappings.
\begin{lemma}
\label{lemma:KKT_monotonicy}
Let $\beta \in \mathscr{B}_{p+d}(R)$ such that $n_{j,\bullet}^\top \beta_{j, \bullet} = 0$, and $h = (h_{1, \bullet}^\top, \ldots, h_{p,\bullet}^\top)^\top$ with $h_{j, \bullet} \in \partial\big(\norm{\beta_{j,\bullet}}_{\TV,\omega_{j, \bullet}}\big)$ for all $j = 1, \ldots, p$. Then the following holds:
\begin{equation*}
(\hat \beta - \beta)^\top \nabla{\ell}(f_{\hat \beta}) \leq - (\hat \beta - \beta)^\top H_n(f_{\hat \beta}) - (\hat \beta - \beta)^\top h.
\end{equation*}
\end{lemma}
Next, Lemma~\ref{lemma:self-concordance} is derived from the self-concordance definition and Lemma 1 in~\cite{bach2010selfconcordance}. It connects the empirical $\ell_2$-norm defined in~\eqref{eq:empirical-l2-norm} to our empirical divergence defined in~\eqref{eq:def-KLn}.

\begin{lemma}
\label{lemma:self-concordance}
Let $\hat \beta$ be defined by Equation~\eqref{problem-estim} and $\beta \in \mathscr{B}_{p+d}(R)$. Then the following inequalities hold almost surely:
\begin{equation}
\label{eqn:self_concord_1}
KL_n(f^\star,f_{\beta}) - KL_n(f^\star,f_{\hat \beta})  
+ (\hat \beta - \beta)^\top \nabla \ell(f_{\hat \beta}) \geq 0, 
\end{equation}
and
\begin{equation}
\label{eqn:self_concord_2}
\norm{f^\star - f_\beta}_{n}^2\frac{ \psi(-\norm{f^\star-f_\beta}_\infty)}{\norm{f^\star-f_\beta}_\infty^2} \leq KL_n(f^\star,f_{\beta}) \leq \norm{f^\star - f_\beta}_{n}^2 \frac{ \psi(\norm{f^\star-f_\beta}_\infty)}{\norm{f^\star-f_\beta}_\infty^2},
\end{equation}
where we recall that $\psi(x) = e^{x} - x - 1.$
\end{lemma}
Let us now define the non-negative definite matrix 
\begin{equation*}
  \widehat\Sigma_n(f^\star, \tau) = \sum_{i=1}^n\int_0^\tau \big(X_i^B -\breve{X}_n(t)\big)^{\otimes2}\frac{Y_i(t)e^{f^\star(X_i)}}{S_n^{(0)}(f^\star,t)} \dd \bar{N}(t),
\end{equation*}
where
\begin{equation*}
  \breve{X}_n(t) = \frac{S_n^{(1)}(f^\star,t)}{S_n^{(0)}(f^\star,t)}.
\end{equation*}
This matrix is linked to our empirical norm via the relation $\norm{f_\beta}_{n}^2 = \beta^\top \widehat\Sigma_n(f^\star, \tau) \beta.$
The proof of Theorem~\ref{thm-faster-in-approx} requires the matrix $\widehat\Sigma_n(f^\star, \tau)$ to fulfill a compatibility condition. The following lemma shows that such a condition is true with large probability as long as Assumption~\ref{assump:compatibility} holds.

\begin{lemma}
\label{compatibility-Sigma-T}
Let $\zeta \in \R^{p+d}_+$ be a given vector of non-negative weights and $L = [L_1, \ldots, L_p]$ a concatenation of index subsets.
Set for all $j = 1, \ldots, p$,
\begin{equation}
\label{index-notation}
  L_j = \{a_j^1, \ldots, a_j^{b_j}\} \subset \{1, \ldots, d_j+1\}, 
\end{equation}
with the convention that  $a_j^{0} = 0$  and $a_j^{b_j+1} = d_j +2.$
Then, with a probability greater than $1 - e^{-ns^{(0)}(\tau)^2/8e^{2f^\star_\infty}} - 3\varepsilon$, one has
\begin{equation*}
\label{compatibility-condition-SigmaT}
 \inf\limits_{u \in \mathscr{C}_{1,\omega}(L) \backslash \{\bold 0 \}}\frac{({\bf{T}}u)^\top \widehat \Sigma_n(f^\star, \tau){\bf{T}}u}{|\norm{u_L\odot\zeta_L}_1 - \norm{u_{L^\complement}\odot\zeta_{L^\complement}}_1|^2} \geq \big({\kappa}^2_{\tau}(L) -  \Xi_{\tau}(L)\big) \kappa^2_{\bf{T}, \zeta}(L),
\end{equation*}
where 
\begin{align*}
 \Xi_{\tau}(L) = 4|L|\Big(\frac{8\max_{j}(d_j+1) \max_{j,l}\omega_{jl}}{\min_{j,l}\omega_{j,l}}\Big)^2 \Big\{\big(1 &+ e^{2f^\star_\infty}\Lambda^\star_0(\tau)\big)\sqrt{2/n\log(2(p+d)^2/\varepsilon)} \\
 &+ (2e^{2f^\star_\infty}\Lambda^\star_0(\tau)/s^{(0)}(\tau))t^2_{n,p,d,\varepsilon} \Big\},
\end{align*}
\begin{equation*}
  \kappa_{\bf{T}, \zeta}(L) = \Big(32\sum_{j=1}^p\sum_{l=1}^{d_j+1} |\zeta_{j,l+1} - \zeta_{j,l}|^2 + (b_j +1) \norm{\zeta_{j,\bullet}}_\infty^2 \big\{\min_{1\leq b\leq b^j}|a_j^{b} - a_j^{b-1}|\big\}^{-1}\Big)^{-\frac 1 2 },
 \end{equation*}
and
  \begin{equation*}
\label{C-1-hat-omega}
\mathscr{C}_{1,\omega}(L) \stackrel{}{=} \Big\{u \in \mathscr{B}_{p+d}(R): \sum_{j=1}^p\norm{(u_{j, \bullet})_{L^\complement_j}}_{1,\omega_{j,\bullet}} \leq 3\sum_{j=1}^p \norm{(u_{j, \bullet})_{L_j}}_{1,\omega_{j,\bullet}}  \Big\}.
\end{equation*}
\end{lemma}
We now state a technical result connecting the norms $\norm{\cdot}_1$ and $\norm{\cdot}_2$ on $\mathscr{C}_{\TV,\omega}(L)$.

\begin{lemma}
\label{lemma:technical_l1_TV} 
Let $\Sigma$ and $\tilde \Sigma$ be two non-negative matrices of the same size.
For any concatenation $L = [L_1, \ldots, L_p]$  of index subsets, one has 
\begin{align*}
 \inf\limits_{\beta \in \mathscr{C}_{\TV,\omega}(L) \backslash \{\bold 0 \}}\frac{\beta^\top \tilde \Sigma \beta}{\norm{\beta_{L}}^2_2} \geq \inf\limits_{\beta \in \mathscr{C}_{\TV,\omega}(L) \backslash \{\bold 0 \}}&\frac{\beta^\top\Sigma  \beta}{\norm{\beta_{L}}^2_2}\\
 & - |L|\Big(\frac{8\max_{j}(d_j+1) \max_{j,l}\omega_{jl}}{\min_{j,l}\omega_{j,l}}\Big)^2\max_{j,l} |\Sigma_{j,l} -\tilde \Sigma_{j,l}|.
\end{align*}
\end{lemma}

\subsection{Proof of Theorem~\ref{thm-faster-in-approx}}
\label{proof-theorem-faster-OI}

Combining Lemmas~\ref{lemma:KKT_monotonicy} and~\ref{lemma:self-concordance}, we get
\begin{align*}
KL_n(f^\star,f_{\hat \beta}) &\leq KL_n(f^\star,f_{\beta})  + (\hat \beta - \beta)^\top \nabla \ell(f_{\hat \beta}) \\
&\leq KL_n(f^\star,f_{\beta})  - (\hat \beta - \beta)^\top H_n(f_{\hat \beta}) - (\hat \beta - \beta)^\top h.
\end{align*}
Then, if $- (\hat \beta - \beta)^\top H_n(f_{\hat \beta}) - (\hat \beta - \beta)^\top h < 0$, the theorem holds. 
Let us assume for now that $- (\hat \beta - \beta)^\top H_n(f_{\hat \beta}) - (\hat \beta - \beta)^\top h \geq 0$.

\paragraph{Bound for $- (\hat \beta - \beta)^\top H_n(f_{\hat \beta}) - (\hat \beta - \beta)^\top h$.}

From the definition of the sub-gradient $ \hat h = (\hat h _{1, \bullet}^\top, \ldots, \hat h _{p, \bullet}^\top)^\top \in \partial\big( \norm{\hat \beta}_{\TV,\omega}\big)$,
one can choose $h$ such that
 \begin{equation*}
 h_{j,l} = \left\{
\begin{split}
&2 D_{j}^\top \big(\omega_{j,\bullet}\odot\sgn(D_{{j}}\beta_{j,\bullet})\big) &\text{ if } {l \in \mathcal{A}_j(\beta) },\\
&2 D_{j}^\top \big(\omega_{j,\bullet}\odot\sgn \big( D_{j}(\hat \beta_{j, \bullet} - \beta_{j,\bullet}) \big)\big) &\text{ if } {l \in \mathcal{A}_j^\complement(\beta)}.
\end{split}
\right.
\end{equation*}
This gives
\begin{equation*}
\begin{split}
- (\hat \beta - \beta)^\top h &= -\sum_{j=1}^p (\hat \beta_{j, \bullet} -\beta_{j, \bullet})^\top h_{j, \bullet}  \\
 &= \sum_{j=1}^p \big((-h_{j, \bullet})_{\mathcal{A}_j(\beta)}\big)^\top (\hat \beta_{j, \bullet} -\beta_{j, \bullet})_{  \mathcal{A}_j(\beta)} -\sum_{j=1}^p \big((h_{j, \bullet})_{ \mathcal{A}^\complement_j(\beta)}\big)^\top (\hat \beta_{j, \bullet} -\beta_{j, \bullet})_{ \mathcal{A}^\complement_j(\beta)}\\
&= 2 \sum_{j=1}^p \big((-\omega_{j,\bullet}\odot\sgn(D_{j}\beta_{j,\bullet}))_{ \mathcal{A}_j(\beta)}\big)^\top D_{j}(\hat \beta_{j, \bullet} -\beta_{j, \bullet})_{\mathcal{A}_j(\beta)} \\
& \hspace{.5cm} - 2\sum_{j=1}^p \big((\omega_{j,\bullet}\odot\sgn \big( D_{j}(\hat \beta_{j, \bullet} - \beta_{j,\bullet}))_{\mathcal{A}^\complement_j(\beta)}\big)^\top D_{j} ({\hat \beta}_{j, \bullet} - \beta_{j, \bullet})_{ \mathcal{A}^\complement_j(\beta)}.
\end{split}
\end{equation*}
Using the fact that $u^\top \sgn(u)= \norm{u}_1$, we have that
\begin{align}
\label{ineq2-proof-thm11}
- (\hat \beta - \beta)^\top h & \leq 2\sum_{j=1}^p  \norm{(\omega_{j,\bullet})_{ \mathcal{A}_j(\beta)}\odot D_{{j}}(\hat \beta_{j, \bullet} -\beta_{j, \bullet})_{ \mathcal{A}_j(\beta)}}_1 \nonumber \\
& \quad - 2\sum_{j=1}^p \norm{ (\omega_{j,\bullet})_{ \mathcal{A}^\complement_j(\beta)}\odot D_{{j}}(\hat \beta_{j, \bullet} -\beta_{j, \bullet})_{\mathcal{A}^\complement_j(\beta)}}_1 \nonumber \\
& = 2\sum_{j=1}^p \norm{(\hat \beta_{j, \bullet} -\beta_{j, \bullet})_{\mathcal{A}_j(\beta)}}_{\TV,\omega_{j,\bullet}} - 2\sum_{j=1}^p \norm{ (\hat \beta_{j, \bullet} -\beta_{j, \bullet})_{\mathcal{A}^\complement_j(\beta)}}_{\TV,\omega_{j,\bullet}}.
\end{align}
Inequality~\eqref{ineq2-proof-thm11} therefore gives
\begin{align*}
KL_n(f^\star,f_{\hat \beta}) & \leq KL_n(f^\star,f_{\beta}) - (\hat \beta - \beta)^\top H_n(f_{\hat \beta}) + 2\sum_{j=1}^p  \norm{(\hat \beta_{j, \bullet} -\beta_{j, \bullet})_{ \mathcal{A}_j(\beta)}}_{\TV,\omega_{j,\bullet}} \\
& \quad - 2\sum_{j=1}^p \norm{ (\hat \beta_{j, \bullet} -\beta_{j, \bullet})_{\mathcal{A}^\complement_j(\beta)}}_{\TV,\omega_{j,\bullet}}.
\end{align*}
Using the fact that ${\bf{T}}{\bf{D}} = \bf{I}$ (see their definitions in Equation~\eqref{differenceMat-D-inverse-T}), we get 
\begin{align*}
KL_n(f^\star,f_{\hat \beta}) 
& \leq KL_n(f^\star,f_{\beta}) - \big({\bf{D}}(\hat \beta - \beta)\big)^\top \mathbf{T}^\top H_n(f_{\hat \beta}) \\
& \quad + 2\sum_{j=1}^p  \norm{(\hat \beta_{j, \bullet} -\beta_{j, \bullet})_{ \mathcal{A}_j(\beta)}}_{\TV,\omega_{j,\bullet}}  - 2\sum_{j=1}^p \norm{ (\hat \beta_{j, \bullet} -\beta_{j, \bullet})_{\mathcal{A}^\complement_j(\beta)}}_{\TV,\omega_{j,\bullet}}.
\end{align*}
On the event 
\begin{equation}
  \label{event-En}
  \mathscr{E}_n := \Big\{ |{\bf T}^\top H_n(f_{\hat \beta})| \leq (\omega_{1,1}, \ldots, \omega_{p,d_{p}+1}) \Big\} 
\end{equation}(the vector comparison has to be understood elementwise),
we have 
\begin{align*}
KL_n(f^\star,f_{\hat \beta}) 
&\leq KL_n(f^\star,f_{\beta}) + \sum_{j=1}^p\sum_{l=1}^{d_j+1}\omega_{j,l}|\big({\bf{D}}(\hat \beta - \beta)\big)_{j,l}|\\
&\quad + 2\sum_{j=1}^p  \norm{(\hat \beta_{j, \bullet} -\beta_{j, \bullet})_{ \mathcal{A}_j(\beta)}}_{\TV,\omega_{j,\bullet}} - 2\sum_{j=1}^p \norm{ (\hat \beta_{j, \bullet} -\beta_{j, \bullet})_{\mathcal{A}^\complement_j(\beta)}}_{\TV,\omega_{j,\bullet}}.
\end{align*}
Hence,
\begin{align*}
KL_n(f^\star,f_{\hat \beta}) 
&\leq KL_n(f^\star,f_{\beta}) +  \sum_{j=1}^p  \norm{(\hat \beta_{j, \bullet} -\beta_{j, \bullet})_{ \mathcal{A}_j(\beta)}}_{\TV,\omega_{j,\bullet}} +  \sum_{j=1}^p \norm{ (\hat \beta_{j, \bullet} -\beta_{j, \bullet})_{\mathcal{A}^\complement_j(\beta)}}_{\TV,\omega_{j,\bullet}}\\
 &\quad + 2\sum_{j=1}^p \norm{(\hat \beta_{j, \bullet} -\beta_{j, \bullet})_{ \mathcal{A}_j(\beta)}}_{\TV,\omega_{j,\bullet}}  - 2\sum_{j=1}^p \norm{ (\hat \beta_{j, \bullet} -\beta_{j, \bullet})_{\mathcal{A}^\complement_j(\beta)}}_{\TV,\omega_{j,\bullet}}\\
 &\leq KL_n(f^\star,f_{\beta}) + 3 \sum_{j=1}^p  \norm{(\hat \beta_{j, \bullet} -\beta_{j, \bullet})_{ \mathcal{A}_j(\beta)}}_{\TV,\omega_{j,\bullet}} - \sum_{j=1}^p \norm{ (\hat \beta_{j, \bullet} -\beta_{j, \bullet})_{\mathcal{A}^\complement_j(\beta)}}_{\TV,\omega_{j,\bullet}}.
\end{align*}
One therefore has 
\begin{align}
\label{KL-TV}
KL_n(f^\star,f_{\hat \beta}) 
&\leq KL_n(f^\star,f_{\beta}) + 3 \sum_{j=1}^p  \norm{(\hat \beta_{j, \bullet} -\beta_{j, \bullet})_{ \mathcal{A}_j(\beta)}}_{\TV,\omega_{j,\bullet}}.
\end{align}
On the event $\mathscr{E}_n$, the following also holds
\begin{align*}
  \sum_{j=1}^p \norm{ (\hat \beta_{j, \bullet} -\beta_{j, \bullet})_{\mathcal{A}^\complement_j(\beta)}}_{\TV,\omega_{j,\bullet}} \leq 3 \sum_{j=1}^p  \norm{(\hat \beta_{j, \bullet} -\beta_{j, \bullet})_{ \mathcal{A}_j(\beta)}}_{\TV,\omega_{j,\bullet}},
\end{align*}
which means that $\hat \beta - \beta \in \mathscr{C}_{\TV,\omega}\big(\mathcal{A}({\beta})\big)$ and ${\bf{D}}(\hat \beta - \beta) \in \mathscr{C}_{1,\omega}\big(\mathcal{A}({\beta})\big)$.
Now returning to~\eqref{KL-TV}, by Lemma~\ref{compatibility-Sigma-T} and under Assumption~\ref{assump:compatibility}, we get 
\begin{align}
\label{eqn:inequ_comp1}
KL_n(f^\star,f_{\hat \beta}) 
&\leq KL_n(f^\star,f_{\beta}) + \frac{\norm{f_{\hat \beta} - f_\beta}_{n}}{\sqrt{{\kappa}^2_{\tau}\big(\mathcal{A}({\beta})\big) - \Xi_{\tau}\big(\mathcal{A}({\beta})\big)}\kappa_{\bf{T}, \hat\zeta}\big(\mathcal{A}({\beta})\big)},
\end{align}
where 
\begin{equation*}
\hat \zeta_{j,l} =
  \begin{cases}
    3 \omega_{j,l} &\text{ if } l \in \mathcal{A}({\beta}), \\
    0 &\text{ if } l \in \mathcal{A}^\complement({\beta}).
  \end{cases}
\end{equation*}
The second term in the right-hand side of~\eqref{eqn:inequ_comp1} fulfills 
\begin{align*}
\frac{\norm{f_{\hat \beta} - f_\beta}_{n}}{\sqrt{{\kappa}^2_{\tau}\big(\mathcal{A}({\beta})\big) - \Xi_{\tau}\big(\mathcal{A}({\beta})\big)}\kappa_{\bf{T}, \hat\zeta}\big(\mathcal{A}({\beta})\big)}
\leq \frac{\norm{f^\star - f_{\hat \beta}}_{n} + \norm{f^\star - f_{\beta} }_{n}}{\sqrt{{\kappa}^2_{\tau}\big(\mathcal{A}({\beta})\big) - \Xi_{\tau}\big(\mathcal{A}({\beta})\big)}\kappa_{\bf{T}, \hat\zeta}\big(\mathcal{A}({\beta})\big)}.
\end{align*}
By~\eqref{eqn:self_concord_2} in Lemma~\ref{lemma:self-concordance}, we get that
\begin{align*}
&\norm{f^\star - f_\beta}_{n} \leq \sqrt{\frac{\norm{f^\star-f_\beta}_\infty^2}{ \psi(-\norm{f^\star-f_\beta}_\infty)} KL_n(f^\star,f_{\beta})}.
\end{align*}
Introducing $g(x) = x^2/\psi(-x) = x^2/(e^{-x} + x +1)$, we note that 
\begin{equation}\label{eq:ineq_g}
g(x) \leq x+2 \text{ for any } x > 0.
\end{equation}  Then 
\begin{align*}
&\norm{f^\star - f_\beta}_{n} \leq \sqrt{(\norm{f^\star-f_\beta}_\infty + 2) KL_n(f^\star,f_{\beta})}.
\end{align*}
In addition, one can easily check that $\max_{1 \leq i \leq n} \sup_{\beta \in \mathscr{B}_{p+d}(R)} |f_\beta(X_i)| \leq R$. 
Hence,
\begin{align*}
\norm{f^\star - f_\beta}_\infty &\leq \max_{1 \leq i \leq n}\big\{|f^\star(X_i)| + |f_\beta(X_i)|\big\} \leq f^\star_\infty + R.
\end{align*}
This implies that 
\begin{align*}
&\norm{f^\star - f_\beta}_{n} \leq \sqrt{(f^\star_\infty + R + 2) KL_n(f^\star,f_{\beta})}.
\end{align*}
With these bounds, inequality~\eqref{eqn:inequ_comp1} yields
\begin{align*}
&KL_n(f^\star,f_{\hat \beta}) 
\leq KL_n(f^\star,f_{\beta}) + \sqrt{(f^\star_\infty + R + 2)}
\frac{\sqrt{KL_n(f^\star,f_{\beta})} + \sqrt{KL_n(f^\star,f_{\hat\beta})}}{\sqrt{{\kappa}^2_{\tau}\big(\mathcal{A}({\beta})\big) - \Xi_{\tau}\big(\mathcal{A}({\beta})\big)}\kappa_{\bf{T}, \hat\zeta}\big(\mathcal{A}({\beta})\big)}.
\end{align*}
We now use the elementary inequality $2uv \leq \varrho u^2 + v^2 /\varrho$ with $\varrho > 0.$
We get 
\begin{align*}
&KL_n(f^\star,f_{\hat \beta}) \leq KL_n(f^\star,f_{\beta}) \\
& + \frac{\varrho(f^\star_\infty + R + 2)}{2\Big({\kappa}^2_{\tau}\big(\mathcal{A}({\beta})\big) - \Xi_{\tau}\big(\mathcal{A}({\beta})\big)\Big) \kappa^2_{\bf{T}, \hat\zeta}\big(\mathcal{A}({\beta})\big)} + \frac{1}{2\varrho}\big( \sqrt{KL_n(f^\star,f_{\beta})} + \sqrt{KL_n(f^\star,f_{\hat\beta})} \big)^2.
\end{align*}
Hence
\begin{align*}
\big(1 - \frac{1}{\varrho}\big) KL_n(f^\star,f_{\hat \beta}) 
&\leq \big(1 + \frac{1}{\varrho}\big) KL_n(f^\star,f_{\beta}) \\&\qquad+\frac{\varrho(f^\star_\infty + R + 2)}{2\Big({\kappa}^2_{\tau}\big(\mathcal{A}({\beta})\big) - \Xi_{\tau}\big(\mathcal{A}({\beta})\big)\Big) \kappa^2_{\bf{T}, \hat\zeta}\big(\mathcal{A}({\beta})\big)}.
\end{align*}
By choosing $\varrho = 2$, we obtain  
\begin{align*}
KL_n(f^\star,f_{\hat \beta}) 
&\leq 3KL_n(f^\star,f_{\beta}) + \frac{2(f^\star_\infty + R + 2)}{\Big({\kappa}^2_{\tau}\big(\mathcal{A}({\beta})\big) - \Xi_{\tau}\big(\mathcal{A}({\beta})\big)\Big) \kappa^2_{\bf{T}, \hat\zeta}\big(\mathcal{A}({\beta})\big)}.
\end{align*}
On the other hand, by definition of $\kappa^2_{\bf{T}, \zeta}$ (see Lemma~\ref{compatibility-Sigma-T}), we know that
\begin{equation*}
\frac{1}{\kappa^2_{\bf{T}, \hat\zeta}\big(\mathcal{A}({\beta})\big)}
\leq 512 |\mathcal{A}({\beta})| \max_{1 \leq j \leq p}\norm{(\omega_{j,\bullet})_{\mathcal{A}_j(\beta)}}^2_\infty.
\end{equation*}
Finally,
\begin{align*}
KL_n(f^\star,f_{\hat \beta}) 
&\leq 3 KL_n(f^\star,f_{\beta}) +  \frac{1024(f^\star_\infty + R + 2)|\mathcal{A}({\beta})|\max_{1 \leq j \leq p}\norm{(\omega_{j,\bullet})_{\mathcal{A}_j(\beta)}}^2_\infty}{{{\kappa}^2_{\tau}\big(\mathcal{A}({\beta})\big) - \Xi_{\tau}\big(\mathcal{A}({\beta})\big)}}.
\end{align*}
Therefore, on the event $\mathscr{E}_n$, we obtain the desired result.  
\paragraph{Computation of $\P[\mathscr{E}_n^\complement]$.}
From the definition of $H_n$ in Equation~\eqref{eqn:procH_n}, ${\bf{T}}^\top  H_n(f_{\hat \beta})$ is written:
\begin{equation*}
{\bf{T}}^\top  H_n(f_{\hat \beta}) = -\frac 1n \sum_{i=1}^n \int_0^\tau \Big\{{\bf{T}}^\top X_i^B - {\bf{T}}^\top\frac{S_n^{(1)}(f_{\hat \beta},t)}  {S_n^{(0)}(f_{\hat \beta}, t)}\Big\}\dd M_i(t).
\end{equation*}
Hence, each component of this vector has the form required to apply Theorem 3 from~\cite{GaiGui-12}. We recall that $
H_n$ and ${\bf{T}}^\top  H_n$ have a block structure: they are vectors of $p$ blocks of length $d_j+1$ for all $j = 1,\ldots,p$. We then denote by $\big({\bf{T}}^\top  H_n\big)_{j,l}$ the $l$th component of the $j$th block. 

In addition, due to the definition of $X_i^B$, we know that each coefficient of ${\bf{T}}^\top X_i^B$ takes a value lower than $1$. As a consequence, for all $t \leq \tau$, one has
\begin{equation*}
\left|\big({\bf{T}}^\top X_i^B - {\bf{T}}^\top\frac{S_n^{(1)}(f_{\hat \beta},t)}  {S_n^{(0)}(f_{\hat \beta}, t)}\big)_{j,k}\right|\leq \left|\big({\bf{T}}^\top X_i^B\big)_{j,k}\right| + \left|\big({\bf{T}}^\top\frac{S_n^{(1)}(f_{\hat \beta},t)}  {S_n^{(0)}(f_{\hat \beta}, t)}\big)_{j,k}\right|\leq 2.
\end{equation*}
We now use  Theorem 3 from~\cite{GaiGui-12} to obtain
\begin{align*}
\P\Big[\big|\big({\bf{T}}^\top H_n(f_{\hat \beta}, t)\big)_{j,l}\big| \geq 5.64&\sqrt{\frac{c + \scrL_{n,c}}{n} }+ 18.62  \frac{(c + 1+\scrL_{n,c})}{n} \Big]\leq 28.55 e^{-c},
\end{align*}
and by choosing the weights $\omega_{j,l}$ as defined in~\eqref{weights}, we conclude that $\P[\mathscr{E}_n^\complement]\leq 28.55 e^{-c}$ for some $c>0$.

$\hfill\square$

\subsection{Proofs of the lemmas}
\label{proof-lemmas}

\subsubsection{Proof of Lemma~\ref{lemma:KKT_monotonicy}} 
To characterize the solution of  Problem~\eqref{problem-estim}, the following result can be sraightforwardly obtained using the Karush-Kuhn-Tucker (KKT) optimality conditions~\citep{boyd2004convex} for a convex optimization problem.
A vector $\hat \beta\in \R^{p+d}$ is an optimum of the objective function in \eqref{problem-estim} if and only if there exists the following three sequences of subgradient: 
\begin{equation*}
\left\{
\begin{split}
\hat h &= (\hat h_{j, \bullet})_{j=1, \ldots, p} \text{ with } \hat h_{j, \bullet} \in \partial\big(\norm{\hat \beta_{j,\bullet}}_{\TV,\omega_{j, \bullet}}\big), \\
\hat g &= (\hat g_{j, \bullet}) _{j=1, \ldots, p} \text{ with } \hat g_{j, \bullet} \in \partial\big(\delta_{j}(\hat \beta_{j,\bullet})\big), \\
\hat k &\in \partial\big(\delta_{\mathscr{B}_{p+d}(R)}(\hat \beta)\big)
\end{split}
\right.
\end{equation*}
such that 
\begin{equation}
\label{kkt-result}
(\nabla \ell_n^{}(f_{\hat \beta}))_{j, \bullet} + \hat h_{j, \bullet} +  \hat g_{j, \bullet} +  \hat k_{j, \bullet} = \mathbf{0},
\end{equation}
for all $j=1, \ldots, p$, and where
\begin{equation*}
{\hat h}_{j,l} \left\{
\begin{array}{ll}
  = \Big(D_{{j}}^\top \big(\omega_{j,\bullet} \odot\sgn(D_{{j}}{ \hat\beta}_{j,\bullet})\big)\Big)_l & \mbox{if } l \in  \mathcal{A}_j(\hat \beta), \\
  \in  \Big(D_{{j}}^\top \big(\omega_{j,\bullet} \odot {[-1,+1]}^{d_j+1}\big)\Big)_l & \mbox{if } l \in \mathcal{A}^\complement_j(\hat \beta),
\end{array} 
\right.
\end{equation*}
where $\mathcal{A}(\hat \beta)$ is the active set of $\hat \beta$, see \eqref{active-set-hat-beta}.
The subgradient $\hat g_{j,\bullet}$ belongs to 
\begin{equation*}
\partial\big(\delta_{j}(\hat \beta_{j,\bullet})\big) = \big\{v\in \R^{d_j+1}: (\hat \beta_{j,\bullet} - \beta_{j,\bullet})^\top v \geq 0 \text{ for all }\beta_{j, \bullet} \text{ such that } n_{j,\bullet}^\top \beta_{j, \bullet}= 0\big\},
\end{equation*}
and $\hat k$ to 
\begin{equation*}
\partial\big(\delta_{\mathscr{B}_{p+d}(R)}(\hat \beta)\big) = \big\{v\in \R^{p+d}: (\hat \beta - \beta)^\top v \geq 0 \text{ for all }\beta\text{ such that } \sum_{j=1}^p \norm{\beta_{j,\bullet}}_\infty \leq R\big\}.
\end{equation*}
From Equation~\eqref{kkt-result}, and considering any vector $\beta \in \mathbb R^{p+d}$, we obtain
\begin{equation}
\label{eq:KKT-corollary}
(\hat \beta - \beta)^\top \nabla{\ell}_n(f_{\hat \beta}) + (\hat \beta - \beta)^\top ({\hat h} + {\hat g} + {\hat k}) = 0,
\end{equation}
and Equation~\eqref{eqn:gradients} gives
\begin{equation*}
(\hat \beta - \beta)^\top \nabla{\ell}(f_{\hat \beta}) + (\hat \beta - \beta)^\top H_n(f_{\hat \beta}) + (\hat \beta - \beta)^\top ({\hat h} + {\hat g} + {\hat k}) = 0.
\end{equation*}
Consider now a vector $\beta \in \mathscr{B}_{p+d}(R)$ such that $n_{j,\bullet}^\top \beta_{j, \bullet} = 0$ for all $j =1, \ldots, p$, and $h \in  \partial\big( \norm{\beta}_{\TV, \omega}\big)$. Then, the monotony of sub-differential mappings (which is an immediate consequence of their definition, see \citet{Roc-70}) gives the result.

$\hfill\square$

\subsubsection{Proof of Lemma~\ref{lemma:self-concordance}} 
Let us consider the function $G: \R \rightarrow \R$ defined by $G(\eta) = \ell(f_1 + \eta f_2)$, i.e.,
\begin{align*}
G(\eta) &= - \frac{1}{n}\sum_{i=1}^n \int_0^\tau (f_1 + \eta f_2)(X_i)Y_i(t) e^{f^\star(X_i)}\lambda_0^\star(t) \dd t\\
&\qquad + \frac{1}{n} \int_0^\tau \log \big\{S^{(0)}_n(f_1 + \eta f_2, t)\big\} S^{(0)}_n(f^\star, t)\lambda_0^{\star}(t) \dd t.
\end{align*} 
By differentiating $G$ with respect to the variable $\eta$, we get
\begin{align*}
G'(\eta) &= - \frac{1}{n}\sum_{i=1}^n \int_0^\tau f_2(X_i)Y_i(t) e^{f^\star(X_i)}\lambda_0^\star(t) \dd t \\
&\qquad+ \frac1n \int_0^\tau\frac{\sum_{i=1}^nf_2 (X_i)Y_i(t) \exp\big(f_1(X_i) + \eta f_2(X_i)\big)}{\sum_{i=1}^nY_i(t) \exp\big(f_1(X_i) + \eta f_2(X_i)\big)}S^{(0)}_n(f^\star, t)\lambda_0^{\star}(t) \dd t,
\end{align*}
and 
\begin{align*}
G^{''}(\eta) &= \frac1n \int_0^\tau\frac{\sum_{i=1}^nf^2_2(X_i)Y_i(t) \exp\big(f_1(X_i) + \eta f_2(X_i)\big)}{\sum_{i=1}^nY_i(t) \exp\big(f_1(X_i) + \eta f_2(X_i)\big)}S^{(0)}_n(f^\star, t)\lambda_0^{\star}(t) \dd t\\
& \qquad- \int_0^\tau \bigg(\frac{\sum_{i=1}^nf_2(X_i)Y_i(t) \exp\big(f_1(X_i) + \eta f_2(X_i)\big)}{\sum_{i=1}^nY_i(t) \exp\big(f_1(X_i) + \eta f_2(X_i)\big)}\bigg)^2S^{(0)}_n(f^\star, t)\lambda_0^{\star}(t) \dd t.
\end{align*}
For a given $t \geq 0$, we now consider the discrete random variable $U_t$ that takes the value $f_2(X_i)$ with probability
\begin{equation*}
 \mathbb P[U_t = f_2(X_i)]  = \pi_{t, f_1, f_2, \eta}(i) = \frac{Y_i(t) \exp\big(f_1(X_i) + \eta f_2(X_i)\big)}{\sum_{i=1}^nY_i(t) \exp\big(f_1(X_i) + \eta f_2(X_i)\big)}.
\end{equation*}
We observe that for all $k \in \N$, one has
\begin{equation*}
\frac{\sum_{i=1}^nf^k_2(X_i)Y_i(t) \exp\big(f_1(X_i) + \eta f_2(X_i)\big)}{\sum_{i=1}^nY_i(t) \exp\big(f_1(X_i) + \eta f_2(X_i)\big)} = \E_{\pi_{t, f_1, f_2, \eta}}[U_t^k].
\end{equation*}
Then
\begin{equation*}
G'(\eta) = - \frac{1}{n}\sum_{i=1}^n \int_0^\tau f_2(X_i)Y_i(t) e^{f^\star(X_i)}\lambda_0^\star(t) \dd t+ \frac1n\int_0^\tau \E_{\pi_{t, f_1, f_2, \eta}}[U_t]S^{(0)}_n(f^\star, t)\lambda_0^{\star}(t) \dd t,
\end{equation*}
and 
\begin{align*}
G^{''}(\eta) &=  \frac1n \int_0^\tau \Big(\E_{\pi_{t, f_1, f_2, \eta}}[U_t^2] - \big(\E_{\pi_{t, f_1, f_2, \eta}}[U_t]\big)^2\Big)S^{(0)}_n(f^\star, t)\lambda_0^{\star}(t) \dd t\\
&= \frac1n \int_0^\tau\mathbb V_{\pi_{t, f_1, f_2, \eta}}[U_t]S^{(0)}_n(f^\star, t)\lambda_0^{\star}(t) \dd t.
\end{align*}
Differentiating again, we obtain
\begin{equation*}
G^{'''}(\eta) = \frac1n \int_0^\tau \E_{\pi_{t, f_1, f_2, \eta}}\Big[\big(U_t - \E_{\pi_{t, f_1, f_2, \eta}}[U_t]\big)^3\Big]S^{(0)}_n(f^\star, t)\lambda_0^{\star}(t) \dd t.
\end{equation*}
Therefore, we have
\begin{align*}
G^{'''}(\eta) &\leq\frac1n \int_0^\tau \E_{\pi_{t, f_1, f_2, \eta}}\Big[\big|U_t - \E_{\pi_{t, f_1, f_2, \eta}}[U_t]\big|^3\Big]S^{(0)}_n(f^\star, t)\lambda_0^{\star}(t) \dd t\\
& \leq \frac1n 2\norm{f_2}_\infty\int_0^\tau \E_{\pi_{t, f_1, f_2, \eta}}\Big[\big(U_t - \E_{\pi_{t, f_1, f_2, \eta}}[U_t]\big)^2\Big]S^{(0)}_n(f^\star, t)\lambda_0^{\star}(t) \dd t\\
&\leq 2\norm{f_2}_\infty G^{''}(\eta),
\end{align*}
where $\norm{f_2}_\infty := \max_{1 \leq i \leq n}|f_2(X_i)|.$
Applying now Lemma 1 in~\cite{bach2010selfconcordance} to $G$, we obtain for all $\eta \geq 0$,
\begin{equation}\label{eqn:selfconcordance}
  {G}^{''}(0)\frac{ \psi(-\norm{f_2}_\infty)}{\norm{f_2}_\infty^2}\leq G(\eta) - G(0) - \eta {G}^{'}(0) \leq {G}^{''}(0) \frac{ \psi(\norm{f_2}_\infty)}{\norm{f_2}_\infty^2}. 
\end{equation}
We will apply inequalities in~\eqref{eqn:selfconcordance} in the  following two situations:
\begin{itemize}
  \item Case \#1: $\eta=1$, $f_1 = f_{\hat \beta}$ and $f_2 = f_{ \beta} - f_{\hat \beta}$.
  \item Case \#2: $\eta=1$, $f_1 = f^\star$ and $f_2 = f_{\beta} - f^\star$.
\end{itemize}
In case \#1, 
\begin{align*}
G'(0) &= -(\beta - \hat \beta)^\top \frac{1}{n}\sum_{i=1}^n \bigg\{ \int_0^\tau X_i^B Y_i(t) e^{f^\star(X_i)} \lambda_0^\star(t) \dd t \\ 
&\qquad\qquad\qquad\qquad\quad- \int_0^\tau X_i^B Y_i(t) e^{f_{\hat \beta}(X_i)} \frac{ S_n^{(0)}(f^\star,t)}{S_n^{(0)}(f_{\hat \beta},t)} \lambda_0^\star(t) \dd t\bigg\} \\
&=(\beta -\hat  \beta)^\top \nabla \ell(f_{\hat \beta}),
\end{align*}
and then
\begin{equation*}
G(1) - G(0) - G^{'}(0) = \ell(f_{\beta}) - \ell(f_{\hat \beta}) + (\hat \beta - \beta)^\top \nabla \ell(f_{\hat \beta}).
\end{equation*}
With the left bound of the self-concordance inequality~\eqref{eqn:selfconcordance}, we obtain~\eqref{eqn:self_concord_1} in Lemma~\ref{lemma:self-concordance}.

In case \# 2, one gets 
\begin{align*}
G^{'} (0) &= 0, \\
\text{ and } \quad G^{''}(0) &= \frac1n \int_0^\tau\frac{\sum_{i=1}^n\big(f_{\beta}(X_i)-f^\star(X_i)\big)^2Y_i(t) e^{f^\star(X_i)}}{\sum_{i=1}^nY_i(t) e^{f^\star(X_i)}}S^{(0)}_n(f^\star, t)\lambda_0^{\star}(t) \dd t\\
   & \qquad- \frac1n \int_0^\tau \bigg(\frac{\sum_{i=1}^n(f_{\beta}(X_i)-f^\star(X_i))Y_i(t) e^{f^\star(X_i)}}{\sum_{i=1}^nY_i(t) e^{f^\star(X_i)}}\bigg)^2S^{(0)}_n(f^\star, t)\lambda_0^{\star}(t) \dd t\\
&= \norm{f^\star - f_\beta}_{n}^2,
\end{align*}
which gives~\eqref{eqn:self_concord_2} in Lemma~\ref{lemma:self-concordance}.

$\hfill\square$

\subsubsection{Proof of Lemma~\ref{compatibility-Sigma-T}}

For any concatenation of index sets $L = [L_1, \ldots, L_p]$, we define 
\begin{equation*}
  \hat\kappa_{\tau}(L) = \inf\limits_{\beta \in \mathscr{C}_{\TV,\omega}(L) \backslash \{\bold 0 \}}\frac{ \sqrt{\beta^\top \hat \Sigma_n(f^\star, \tau) \beta }}{\norm{\beta_{L}}_2}.
\end{equation*}
To prove Lemma~\ref{compatibility-Sigma-T}, we will first establish the following lemma, which asssures us that if Assumption~\ref{assump:compatibility} is fulfilled, our random bound $\hat\kappa_{\tau}(L)$ is bounded away from $0$ with large probability.
\begin{lemma}
\label{lemma-compatibility-in-proba}
Let $L = [L_1, \ldots, L_p]$ be a concatenation of index sets. Then, 
\begin{align*}
\hat \kappa^2_{\tau}(L) \geq \kappa^2_{\tau}&(L)- 4|L|\Big(\frac{8\max_{j}(d_j+1) \max_{j,l}\omega_{jl}}{\min_{j,l}\omega_{j,l}}\Big)^2\\
& \times \Big\{ \big(1 + e^{2f^\star_\infty}\Lambda^\star_0(\tau)\big) \sqrt{2/n\log(2(p+d)^2/\varepsilon)}+ (2e^{2f^\star_\infty}\Lambda^\star_0(\tau)/s^{(0)}(\tau))t^2_{n,p,d,\varepsilon} \Big\}
\end{align*}
holds with probability at least $1 - e^{-ns^{(0)}(\tau)^2/8e^{2f^\star_\infty}} - 3\varepsilon$. 
\end{lemma}

\noindent{\it {Proof of Lemma~\ref{lemma-compatibility-in-proba}.}} The proof is adapted from Theorem 4.1 in~\cite{huang2013}, with the difference that we work here in a fixed design setting. We break down the proof into three steps.

\noindent{\it {Step 1.}}
By replacing $\dd \bar{N}(t)$ by its compensator $n^{-1}S_n^{0}(f^\star,t)\lambda_0^\star(t)\dd t$, an approximation of $\widehat\Sigma_n(f^\star, \tau)$ can be defined by 
\begin{equation*}
\bar{\Sigma}_n(f^\star, \tau) = \frac{1}{n}\sum_{i=1}^n \int_0^{\tau} \big(X_i^B - \breve{X}_n(s)\big)^{\otimes2}Y_i(s)e^{f^\star(X_i)} \lambda_0^\star(s)\dd s.
\end{equation*}
The $(m,m')$th component of 
\begin{equation*}
\sum_{i=1}^n \big(X_i^B - \breve{X}_n(s)\big)^{\otimes2}\frac{Y_i(s)e^{f^\star(X_i)}}{\sum_{i=1}^nY_i(s)e^{f^\star(X_i)}}
\end{equation*}is given by 
\begin{equation*}
  \sum_{i=1}^n \big[(X_i^B)_m - \big(\breve{X}_n(s)\big)_m\big]\big[(X_i^B)_{m'} - (\breve{X}_n(s)\big)_{m'}\big]\frac{Y_i(s)e^{f^\star(X_i)}}{\sum_{i=1}^nY_i(s)e^{f^\star(X_i)}},
\end{equation*}
which is bounded by $4$ in our case. Moreover, we know that 
\begin{equation*}
 \int_0^\tau Y_i(t) \dd N_i(t) \leq 1 \; \text{ for all }\; i = 1,\ldots,n.
 \end{equation*} 
 Thus, Lemma 3.3 in~\cite{huang2013} applies and
\begin{align*}
  \P\big[\big(\widehat\Sigma_n(f^\star, \tau) - \bar{\Sigma}_n(f^\star, \tau)\big)_{m,m'} > 4 x\big] \leq 2e^{-nx^2/2}.
\end{align*}
Next, using an union bound, we get
\begin{align*}
\P\big[\max_{m,m'}\big(\widehat\Sigma_n(f^\star, \tau) - \bar{\Sigma}_n(f^\star, \tau)\big)_{m,m'} > 4 \sqrt{2/n\log\big(2(p+d)^2/\varepsilon\big)}\big] \leq \varepsilon.
\end{align*}
Let
\begin{equation*}
\bar{\kappa}^2_{\tau}(L) = \inf\limits_{\beta \in \mathscr{C}_{\TV,\omega}(L) \backslash \{\bold 0 \}}\frac{\sqrt{\beta^\top \bar{\Sigma}_n(f^\star, \tau)  \beta}}{\norm{\beta_{L}}_2}.
\end{equation*}
Lemma~\ref{lemma:technical_l1_TV} implies that
\begin{align}
\label{relation-hatkappa-barkappa}
\P\Big[\hat \kappa^2_\tau(L) \geq \bar{\kappa}^2_{\tau}(L)  - 4|L|\Big(\frac{8\max_{j}(d_j+1) \max_{j,l}\omega_{jl}}{\min_{j,l}\omega_{j,l}}\Big)^2\sqrt{2/n\log\big(2(p+d)^2/\varepsilon\big)} \Big] \geq 1 - \varepsilon.
\end{align}

\noindent{\it {Step 2.}}
Let 
\begin{equation*}
\widetilde{\Sigma}_n(f^\star, \tau) = \frac{1}{n}\sum_{i=1}^n \int_0^{\tau} \big(X_i^B - \bar{X}_n(s)\big)^{\otimes2}Y_i(s)\ e^{f^\star(X_i)} \lambda_0^\star(s) \dd s
\end{equation*}
and
\begin{equation*}
  \tilde\kappa_{\tau}(L) =  \inf\limits_{\beta \in \mathscr{C}_{\TV,\omega}(L) \backslash \{\bold 0 \}}\frac{\sqrt{\beta^\top \widetilde{\Sigma}_n(f^\star, \tau)  \beta}}{\norm{\beta_{L}}_2}.
\end{equation*}
We will now compare $\bar{\kappa}^2_{\tau}(L)$ and $\tilde\kappa^2_{\tau}(L)$. Straightforward computations lead to the following equality:
\begin{align*}
&\sum_{i=1}^n\big(X_i^B - \bar{X}_n(s)\big)^{\otimes2}Y_i(s) e^{f^\star(X_i)}- \sum_{i=1}^n\big(X_i^B - \breve{X}_n(s)\big)^{\otimes2}Y_i(s) e^{f^\star(X_i)}
\\& \quad = S_n^{(0)}(f^\star, s) \big(\breve{X}_n(s) - \bar{X}_n(s)\big)^{\otimes2}.
\end{align*}
Hence,
\begin{align} 
\label{equivalent-A7huang}
\bar{\Sigma}_n(f^\star, \tau) &= \widetilde{\Sigma}_n(f^\star, \tau)-\frac 1n \int_0^{\tau}S_n^{(0)}(f^\star, s) \big(\breve{X}_n(s) - \bar{X}_n(s)\big)^{\otimes2} \lambda_0^\star(s) \dd s.
\end{align}
We first bound the second term on the right-hand side of~\eqref{equivalent-A7huang}. Let
\begin{align*}
 \Delta_n(s) &= \frac 1n S_n^{(0)}(f^\star, s) \big(\breve{X}_n(s) - \bar{X}_n(s)\big),
\end{align*}{}
so that for each $(m,m')$, we get
\begin{equation*}
\Big( \frac 1n \int_0^{\tau}S_n^{(0)}(f^\star, s) \big(\breve{X}_n(s) - \bar{X}_n(s)\big)^{\otimes2} \lambda_0^\star(s)\dd s \Big)_{m,m'} \leq \Big(\frac{\int_0^{\tau}\Delta_n(s)^{\otimes2}\lambda_0^\star(s)\dd s}{n^{-1}S_n^{(0)}(f^\star, \tau)}\Big)_{m,m'}.
\end{equation*}
In our setting, for each $i$ and all $t\leq \tau$, $Y_i(t)e^{f^\star(X_i)} \leq e^{f^\star_\infty}$. 
By Hoeffding's inequality, we then obtain
\begin{equation*}
\P[\dfrac 1n S_n^{(0)}(f^\star, \tau) < s^{(0)}(\tau)/2] \leq e^{-ns^{(0)}(\tau)^2/8e^{2f^\star_\infty}}.
\end{equation*}
Furthermore, we have 
\begin{align*}
\E[\Delta_n(s)|X] = \frac 1n  \sum_{i=1}^n y_i(s)e^{f^\star(X_i)}\Big(X_i^B - \frac{\sum_{i=1}^n X_i^B y_i(s) e^{f^\star(X_i)}}{\sum_{i=1}^ny_i(s) e^{f^\star(X_i)}}\Big) = \mathbf{0},
\end{align*}
and the $(m,m')$th component of $\Delta_n(s)^{\otimes2}$ is given by 
\begin{align*}
\big(\Delta_n(s)^{\otimes2}\big)_{m,m'} &= \frac1{n^2}\sum_{i=1}^n\sum_{i'=1}^nY_i(s)Y_{i'}(s)e^{f^\star(X_i)} e^{f^\star(X_{i'})}\\
&\qquad\qquad\qquad \times\big[(X_i^B)_m - \big(\bar{X}_n(s)\big)_m\big]\big[(X_{i'}^B)_{m'}- \big(\bar{X}_n(s)\big)_{m'}\big].
\end{align*}
Therefore,
$\int_0^{\tau}\big(\Delta_n(s)^{\otimes2}\big)_{m,m'}\lambda^\star_0(s)\dd s$ is a V-statistic for all $(m,m')$. Moreover, 
\begin{equation*}
\int_0^{\tau}\big|\big(\Delta_n(s)^{\otimes2}\big)_{m,m'}\big|\lambda^\star_0(s)\dd s \leq 4e^{2f^\star_\infty}\Lambda^\star_0(\tau),
\end{equation*}
where $\Lambda^\star_0(\tau)=\int_0^{\tau}\lambda^\star_0(s)\dd s$.
By Lemma 4.2 in~\cite{huang2013}, we obtain that
\begin{equation*}
\P\Big[\max_{1\leq m,m'\leq p+d}\pm\int_0^{\tau}\big|\big(\Delta_n(s)^{\otimes2}\big)_{m,m'}\big|\lambda^\star_0(s)\dd s > 4e^{2f^\star_\infty}\Lambda^\star_0(\tau)x^2\Big] \leq 2.221 (p+d)^2\exp\Big(\frac{-nx^2/2}{1 +x/3}\Big).
\end{equation*}
Thanks to~\eqref{equivalent-A7huang}, Lemma~\ref{lemma:technical_l1_TV}, and the above two probability bounds, we obtain
\begin{equation}
\label{kappabar-kappatilde}
\bar{\kappa}_{\tau}^2(L) \geq \tilde\kappa^2_{\tau}(L)  - 8e^{2f^\star_\infty}\Lambda^\star_0(\tau)|L|\Big(\frac{8\max_{j}(d_j+1) \max_{j,l}\omega_{jl}}{\min_{j,l}\omega_{j,l}}\Big)^2\frac{t^2_{n,p,d,\varepsilon}}{s^{(0)}(\tau)}
\end{equation}
holds with probability $1 - e^{-ns^{(0)}(\tau)^2/8e^{2f^\star_\infty}} - \varepsilon$.\\

\noindent{\it {Step 3.}}
Next, $\widetilde{\Sigma}_n(f^\star, \tau)$ is an average of independent matrices with mean ${\Sigma}_n(f^\star, \tau)$ and $\big(\widetilde{\Sigma}_n(f^\star, \tau)\big)_{m,m'}$ which are uniformly bounded by $4e^{2f^\star_\infty}\Lambda^\star_0(\tau)$, so  Hoeffding's inequality ensures that
\begin{equation*}
\P\big[\max_{m,m'}\big|\big(\widetilde{\Sigma}_n(f^\star, \tau)\big)_{m,m'} - \big({\Sigma}_n(f^\star, \tau)\big)_{m,m'}\big| > 4e^{2f^\star_\infty}\Lambda^\star_0(\tau) x \big] \leq (p+d)^2 e^{-nx^2/2}.
\end{equation*}
Again, Lemma~\ref{lemma:technical_l1_TV} implies that with probability larger than $1-\varepsilon$, one has
\begin{equation}
\label{kappa-and-kappatilde}
\tilde{\kappa}_{\tau}^2(L) \geq {\kappa}^2_{\tau}(L)  - 4e^{2f^\star_\infty}\Lambda^\star_0(\tau)|L|\Big(\frac{8\max_{j}(d_j+1) \max_{j,l}\omega_{jl}}{\min_{j,l}\omega_{j,l}}\Big)^2\sqrt{2/n\log\big(2(p+d)^2/\varepsilon\big)}.
\end{equation}
Finally, the result follows from~\eqref{relation-hatkappa-barkappa},~\eqref{kappabar-kappatilde} and~\eqref{kappa-and-kappatilde}.

$\hfill\square$

Going back to the proof of Lemma~\ref{compatibility-Sigma-T}, following Lemma 5 in \cite{alaya2016}, for any $u$ in 
\begin{equation}
\mathscr{C}_{1, \omega}(K) \stackrel{}{=} \bigg\{u \in \R^d: \sum_{j=1}^p \norm{(u_{j, \bullet})_{K_j^\complement}}_{1,\omega_{j,\bullet}}  \leq 3\sum_{j=1}^p \norm{(u_{j, \bullet})_{K_j}}_{1,\omega_{j,\bullet}}  \bigg\},
\end{equation}
the following holds:
\begin{equation*}
\frac{({\bf{T}}u)^\top \widehat \Sigma_n(f^\star, \tau){\bf{T}}u}{|\norm{u_L\odot\zeta_L}_1 - \norm{u_{L^\complement}\odot\zeta_{L^\complement}}_1|^2} \geq\kappa^2_{\bf{T}, \zeta}(L)\frac{({\bf{T}}u)^\top \widehat \Sigma_n(f^\star, \tau){\bf{T}}u}{({\bf{T}}u)^\top {\bf{T}}u}.
\end{equation*}
Then, note that if $u \in \mathscr{C}_{1, \omega}(K)$,  ${\bf{T}} u \in  \mathscr{C}_{ \TV, \omega}(K).$ Hence, by the definition of $\hat \kappa_\tau(L)$ and Lemma~\ref{lemma-compatibility-in-proba}, we obtain the desired result.

$\hfill\square$

\subsubsection{Proof of Lemma~\ref{lemma:technical_l1_TV}}

First, we have that 
\begin{equation*}
|\beta^\top \tilde{\Sigma}\beta - \beta^\top \Sigma \beta| \leq \norm{\beta}_1^2 \max_{j,l} |\tilde{\Sigma}_{j,l} - \Sigma_{j,l}|.
\end{equation*}
Hence, we get 
\begin{equation*}
\beta^\top \tilde{\Sigma}\beta \geq \beta^\top \Sigma \beta - \norm{\beta}_1^2 \max_{j,l} |\tilde{\Sigma}_{j,l} - \Sigma_{j,l}|.
\end{equation*}
Thus, to obtain the desired result, it is sufficient to control $\norm{\beta}_1$ using the cone $\mathscr{C}_{\TV,\omega}$.
Recall that for all $j =1, \ldots, p$, we have $T_j D_j = I$. Then, for any $\beta$ we have that
\begin{align*}
\norm{\beta}_1 &= \sum_{j=1}^p \norm{T_jD_j \beta_{j,\bullet}} \\
&= \sum_{j=1}^p \sum_{l=1}^{d_j+1} \big|\sum_{r=1}^l (D_j\beta_{j,\bullet})_r\big|\\
&\leq \sum_{j=1}^p (d_j+1) \sum_{l=1}^{d_j+1} \big|(D_j\beta_{j,\bullet})_l\big|\\
&\leq \frac{\max_{j}(d_j+1)}{\min_{j,l}\omega_{j,l}}\sum_{j=1}^p\sum_{l=1}^{d_j+1} \omega_{j,l}\big|(D_j\beta_{j,\bullet})_l\big|\\
&\leq \frac{\max_{j}(d_j+1)}{\min_{j,l}\omega_{j,l}}\sum_{j=1}^p \norm{\beta_{j,\bullet}}_{\TV, \omega_{j,\bullet}}.
\end{align*}
For any concatenation of index subsets $L=[L_1, \ldots, L_p] \subset\{1, \ldots, p+d\}$, we then get
\begin{equation*}
  \norm{\beta}_1 \leq \frac{\max_{j}(d_j+1)}{\min_{j,l}\omega_{j,l}}\Big(\sum_{j=1}^p \norm{(\beta_{j,\bullet})_{L_j}}_{\TV, \omega_{j,\bullet}} + \sum_{j=1}^p \norm{(\beta_{j,\bullet})_{L^\complement_j}}_{\TV, \omega_{j,\bullet}}\Big).
\end{equation*}
Now, if $\beta \in \mathscr{C}_{\TV,\omega}(L)$, we obtain 
\begin{equation*}
\norm{\beta}_1 \leq \frac{4\max_{j}(d_j+1)}{\min_{j,l}\omega_{j,l}}\sum_{j=1}^p \norm{(\beta_{j,\bullet})_{L_j}}_{\TV, \omega_{j,\bullet}}.
\end{equation*}
Further, we have that $\norm{\beta_{j,\bullet}}_{\TV, \omega_{j,\bullet}} \leq 2 \max_{j,l}\omega_{j,l} \norm{\beta_{j,\bullet}}_{1}$. Hence, we obtain 
\begin{align}
\norm{\beta}_1 &\leq \frac{8\max_{j}(d_j+1)}{\min_{j,l}\omega_{j,l}} \max_{j,l}\omega_{j,l} \sum_{j=1}^p \norm{(\beta_{j,\bullet})_{L_j}}_{1} \nonumber\\
&= \frac{8\max_{j}(d_j+1)}{\min_{j,l}\omega_{j,l}} \max_{j,l}\omega_{j,l} \norm{\beta_L}_1 \label{eq:norm1-norm1-onL} \\
&\leq \sqrt{|L|}\frac{8\max_{j}(d_j+1)}{\min_{j,l}\omega_{j,l}} \max_{j,l}\omega_{j,l} \norm{\beta_L}_2. \nonumber
\end{align}

$\hfill\square$

\section{Proof of Theorem~\ref{thm-oracle-estimation}}
\label{proof-thm-estim}

\paragraph{On the definition of $b^\star$.}
Let us first make a remark concerning the choice we made to approximate $f^\star$ using $b^\star$.
Instead of what we did in~\eqref{eq:approx-beta-star} and~\eqref{definition-of-b-star}, it may be tempting to define $b^\star$ such that 
\begin{equation*}
\tilde{f}_{j,\bullet} \in \argmin_{f_{\beta_{j,\bullet}} \in \cP^{\mu_{j,\bullet}}} \norm{f^\star_{j, \bullet} - f_{\beta_{j,\bullet}}}_\mathcal{Q}
\end{equation*}
for all $j = 1, \ldots, p$, with $\cP^{\mu_{j,\bullet}}$ the set of $\mu_{j,\bullet}$-piecewise-constant functions defined on $[0,1]$, and $\mathcal{Q}$ denoting either the Hilbert space over $[0,1]$ endowed by the norm $\norm{f}^2 =\int_0^1f^2(x)\dd x$, or the complete normed vector space of real integrable functions in the Lebesgue sense.
In the first case $\big(\mathcal{Q}=L^2([0,1])\big)$, $\tilde{f}_{j,\bullet}$ could be viewed as an orthogonal projection. However, the resulting approximated vector $b^\star$ would almost surely have a support set relative to the total variation penalty double the size of  $\beta^\star$'s one, which is not intuitive. 
In the second case $\big(\mathcal{Q}=L^1([0,1])\big)$, both $\beta^\star$ and $b^\star$ would have the same cardinality of their respective support sets relative to the total variation penalty. But for a given cut-point $\mu^\star_{j,k}$, the corresponding $b^\star$ cut-point would be $\mu_{j,l^\star_{j,k}-1}$ if $\mu^\star_{j,k}$ was closer to $\mu_{j,l^\star_{j,k}-1}$ than to $\mu_{j,l^\star_{j,k}}$ and vice versa, which would make the writing more cumbersome. To get around this difficulty, we defined $\tilde{f}_{j,\bullet}$ in~\eqref{eq:approx-beta-star} such that the corresponding cut-point is always the right bound of $I_{j,l^\star_{j,k}}$, i.e., $\mu_{j,l^\star_{j,k}}$.

\paragraph{On the approximation bias.}
Let us now state an initial lemma concerning the ``bias'' existing between the true function $f^\star$ and its approximation $f_{b^\star}$ defined in~\eqref{definition-of-b-star}. We state the following result bounding $\norm{f^\star - f_{b^\star}}_{n}^2$ with large probability.  Towards this end, we define 
\begin{equation*}
\hat \pi_{j,k} = \frac{|\{i=1, \ldots, n: X_{i,j} \in \mathcal{I}_{j,k}^\star\} |}{n},
\end{equation*}
where we denote
\begin{equation*}
\mathcal{I}_{j,k}^\star = \big(I^\star_{j,k} \cap I_{j,l^\star_{j,k-1}}\big) \bigcup \big((I^\star_{j,k})^c \cap I_{j,l^\star_{j,k}} \big)
\end{equation*}
for all $j =1, \ldots, n$ and $k = 1, \ldots, K^\star_j+1$.

\begin{lemma}
\label{control-of-bias}
The  inequality 
\begin{align*}
\norm{f^\star - f_{b^\star}}_{n}^2 \leq &\Big\{\sum_{j \in \mathscr A(\beta^\star)}\sum_{k=1}^{K^\star_j+1}|\beta^\star_{j,k}|\frac{n_{j,l^\star_{j,k}}}{n}\Big\}^2 \pi_n + \frac{2\pi_n e^{2f^\star_\infty}}{c_Z}\sum_{j \in \mathscr A(\beta^\star)}\sum_{k=1}^{K^\star_j+1}\hat\pi_{j,k}|\beta^\star_{j,k}|^2
\end{align*}
holds with probability at least $1 - 2 e^{-nc_Z^2/2}.$
\end{lemma}

\noindent{\it {Proof of Lemma~\ref{control-of-bias}.}}
We have
\begin{align*}
\norm{f^\star - f_{b^\star}}_{n}^2 = \int_0^\tau \sum_{i=1}^n \big[(f^\star - f_{b^\star})(X_i) - \big(\bar{f^\star}(t) - \bar{f_{b^\star}}(t)\big)\big]^2\frac{Y_i(t)e^{f^\star(X_i)}}{S_n^{(0)}(f^\star,t)}\dd \bar{N}(t)
\end{align*}
and
\begin{equation*}
  \bar{f^\star}(t) - \bar{f}_{b^\star}(t) = \sum_{i=1}^n(f^\star - f_{b^\star})(X_i)\dfrac{Y_i(t)e^{f^\star(X_i)}}{S_n^{(0)}(f^\star,t)}.
\end{equation*}
It is obvious that 
\begin{align*}
\norm{f^\star - f_{b^\star}}_{n}^2 = \int_0^\tau \sum_{i=1}^n \big((f^\star - f_{b^\star})(X_i)\big)^2 \frac{Y_i(t)e^{f^\star(X_i)}}{S_n^{(0)}(f^\star,t)}\dd \bar{N}(t) - \int_0^\tau \big(\bar{f^\star}(t) - \bar{f_{b^\star}}(t)\big)^2 \dd \bar{N}(t),
\end{align*}
which means that
\begin{align}
\label{first-cntrl}
\norm{f^\star - f_{b^\star}}_{n}^2 \leq \int_0^\tau \sum_{i=1}^n \big((f^\star - f_{b^\star})(X_i)\big)^2\frac{Y_i(t)e^{f^\star(X_i)}}{S_n^{(0)}(f^\star,t)}\dd \bar{N}(t).
\end{align}
Next, we control the right-hand-side of~\eqref{first-cntrl}. 
For all $i = 1, \ldots, n$, we have that 
\begin{align*}
(f^\star_j &- f_{b^\star_{j,\bullet}})(X_i)\\
& = \sum_{k=1}^{K^\star_j+1}\beta^\star_{j,k}\big(\ind{}(X_{i,j} \in I^\star_{j,k}) - \sum_{l=l^\star_{j,k-1}+1}^{l^\star_{j,k}}\ind{}(X_{i,j} \in I_{j,l})\big)  + \sum_{k=1}^{K^\star_j+1}\beta^\star_{j,k} \sum_{l=l^\star_{j,k-1}+1}^{l^\star_{j,k}} \frac{n_{j,l}}{n}\\
&= \sum_{k=1}^{K^\star_j+1}\beta^\star_{j,k}\big\{\ind{}(X_{i,j}\in I^\star_{j,k}\cap I_{j,l^\star_{j,k-1}} - \ind{}(X_{i,j}\in (I^\star_{j,k})^c \cap I_{j,l^\star_{j,k}})\big\} + \sum_{k=1}^{K^\star_j+1}\beta^\star_{j,k} \sum_{l=l^\star_{j,k-1}+1}^{l^\star_{j,k}} \frac{n_{j,l}}{n}.
\end{align*}
Then, we obtain
\begin{align*}
|f^\star(X_i) - f_{b^\star}(X_i)|\leq \sum_{j=1}^p\sum_{k=1}^{K^\star_j+1} \big|\beta^\star_{j,k}| \ind{}(X_{i,j}\in \mathcal{I}_{j,k}^\star) + \Big|\sum_{k=1}^{K^\star_j+1}\beta^\star_{j,k}\sum_{l=l^\star_{j,k-1}+1}^{l^\star_{j,k}} \frac{n_{j,l}}{n}\Big|.
\end{align*}
Let us rewrite  constraint~\eqref{sum-to-zero-constraint} such that
\begin{align*}
  0 = \sum_{k=1}^{K^\star_j+1}\beta^\star_{j,k} n^\star_{j,k} &= \sum_{k=1}^{K^\star_j+1}\beta^\star_{j,k} \big(\sum_{l=l^\star_{j,k-1}+1}^{l^\star_{j,k}-1} n_{j,l} + |\{i: X_{i,j} \in \big(I^\star_{j,k}\cap I_{j,l^\star_{j,k}}\big) \cup \big(I^\star_{j,k}\cap I_{j,l^\star_{j,k-1}}\big)\}|\big) 
\end{align*}
(see Figure~\ref{fig:illustration}) to obtain
\begin{align*}
  \sum_{k=1}^{K^\star_j+1}\beta^\star_{j,k} \sum_{l=l^\star_{j,k-1}+1}^{l^\star_{j,k}} n_{j,l}&= \sum_{k=1}^{K^\star_j+1}\beta^\star_{j,k}\big(\sum_{l=l^\star_{j,k-1}+1}^{l^\star_{j,k}-1} n_{j,l} + n_{j,l^\star_{j,k}}\big)\\
  &= \sum_{k=1}^{K^\star_j+1}\beta^\star_{j,k}\big(n_{j,l^\star_{j,k}} - |\{i: X_{i,j} \in (I^\star_{j,k}\cap I_{j,l^\star_{j,k}}) \cup (I^\star_{j,k}\cap I_{j,l^\star_{j,k-1}})\}|\big).
\end{align*}
Hence, 
\begin{equation*}
\Big|\sum_{k=1}^{K^\star_j+1}\beta^\star_{j,k}\sum_{l=l^\star_{j,k-1}+1}^{l^\star_{j,k}} n_{j,l}\Big| \leq \sum_{k=1}^{K^\star_j+1}|\beta^\star_{j,k}|n_{j,l^\star_{j,k}}
\end{equation*}
and \begin{equation*}
|f^\star(X_i) - f_{b^\star}(X_i)|\leq \sum_{j=1}^p\sum_{k=1}^{K^\star_j+1}  \big|\beta^\star_{j,k}| \Big(\ind{}(X_{i,j}\in \mathcal{I}_{j,k}^\star) + \frac{n_{j,l^\star_{j,k}}}{n}\Big). 
\end{equation*}
Bringing this all together, we have that 
\begin{align*}
&\int_0^\tau \sum_{i=1}^n \big((f^\star - f_{b^\star})(X_i)\big)^2\frac{Y_i(t)e^{f^\star(X_i)}}{S_n^{(0)}(f^\star,t)}\dd \bar{N}(t)\\
&\leq\int_0^\tau \sum_{i=1}^n \Big\{\sum_{j=1}^p\sum_{k=1}^{K^\star_j+1} \big|\beta^\star_{j,k}|\big(\ind{}(X_{i,j}\in \mathcal{I}_{j,k}^\star) + \frac{n_{j,l^\star_{j,k}}}{n}\big)\Big\}^2\frac{Y_i(t)e^{f^\star(X_i)}}{S_n^{(0)}(f^\star,t)}\dd \bar{N}(t)\\
&\leq 2 \underbrace{\int_0^\tau \sum_{i=1}^n \sum_{j=1}^p\sum_{k=1}^{K^\star_j+1}|\beta^\star_{j,k}|^2\ind{}(X_{i,j}\in \mathcal{I}_{j,k}^\star)\frac{Y_i(t)e^{f^\star(X_i)}}{S_n^{(0)}(f^\star,t)}\dd \bar{N}(t)}_{(i)}\\
&\qquad+ 2\underbrace{\int_0^\tau \sum_{i=1}^n\Big\{\sum_{j=1}^p\sum_{k=1}^{K^\star_j+1}|\beta^\star_{j,k}|\frac{n_{j,l^\star_{j,k}}}{n}\Big\}^2\frac{Y_i(t)e^{f^\star(X_i)}}{S_n^{(0)}(f^\star,t)}\dd \bar{N}(t)}_{(ii)},
\end{align*}
where we used the fact that the indicator functions are orthogonal.
On the one hand, we have
\begin{align}
(ii) &= \Big\{\sum_{j \in \mathscr A(\beta^\star)}\sum_{k=1}^{K^\star_j+1}|\beta^\star_{j,k}|\frac{n_{j,l^\star_{j,k}}}{n}\Big\}^2 \pi_n \nonumber \\
&\leq
\frac{\max_{j \in \mathscr A(\beta^\star)} \| \beta_{j,\bullet} \|^2_{\infty} \max_{j \in \mathscr A(\beta^\star)} \| n_{j,\bullet}\|^2_\infty}{ n} \big( | \mathscr A (\beta^\star) |+ K^\star\big)  \pi_n \label{first-contrl-of-bias}.
\end{align}
On the other, using the fact that $e^{f^\star(X_i)} \leq e^{f_\infty}$ and $Y_i(t) \leq 1$ for all $t \in [0,\tau]$, we get
\begin{align*}
(i) &\leq e^{f^\star_\infty}\sum_{j=1}^p\sum_{k=1}^{K^\star_j+1} \dfrac 1n \sum_{i=1}^n\ind{}(X_{i,j}\in \mathcal{I}_{j,k}^\star)|\beta^\star_{j,k}|^2 \int_0^\tau\frac{1}{n^{-1}S_n^{(0)}(f^\star,t)} \dd \bar{N}(t)\\
&\leq \frac{\pi_n e^{f^\star_\infty}}{\inf\limits_{t\in[0,\tau]} n^{-1} S_n^{(0)}(f^\star,t)}\sum_{j=1}^p\sum_{k=1}^{K^\star_j+1}\hat\pi_{j,k}|\beta^\star_{j,k}|^2 \\&\leq \frac{\pi_n e^{f^\star_\infty}}{\inf\limits_{t\in[0,\tau]} n^{-1} S_n^{(0)}(f^\star,t)}
\max_{j \in \mathscr A(\beta^\star)} \| \beta_{j,\bullet} \|^2_{\infty} \max_{j \in \mathscr A(\beta^\star)} \| \hat \pi_{j,\bullet}\|_\infty\big( | \mathscr A (\beta^\star) |+ K^\star).
\end{align*}
Moreover, remember that $n^{-1} S_n^{(0)}(f^\star,t) = n^{-1} \sum_{i=1}^n \ind{}(Z_i\geq t)e^{f^\star(X_i)}$, and observe that for all $t\leq \tau$, we have $\{Z_i \geq \tau\} \subset \{Z_i \geq t\}$. Hence,
\begin{align*}
\frac 1n S_n^{(0)}(f^\star,t) \geq e^{-f^\star_{\infty}} \frac 1n \sum_{i=1}^n \ind{}(Z_i\geq \tau) \text{ for all } t\leq \tau.
\end{align*}
Using the Dvoretzky-Kiefer-Wolfowitz inequality~\citep{massart1990}, we get that:
\begin{align*}
\P\Big[\frac 1n \sum_{i=1}^n &\ind{}(Z_i\geq \tau) \geq  \frac 12 \P[Z_1\geq \tau]\Big] \\
&\geq \P\Big[\sqrt{n}\sup\limits_{t\in[0,\tau]}\Big|\frac 1n \sum_{i=1}^n \ind{}(Z_i\geq t) - \P[Z_1 \geq t]\Big|\geq \frac{\sqrt{n}}{2}\P[Z_1\geq \tau]\Big]\\
& \geq 1 - 2e^{-nc_Z^2/2}.
\end{align*}
Then, we have 
\begin{align}
  \P\Big[\inf\limits_{t\in[0,\tau]}\frac 1n S_n^{(0)}(f^\star,t) \geq e^{-f^\star_{\infty}}\frac{c_Z}2\Big] &\geq \P\Big[\frac 1n \sum_{i=1}^n \ind{}(Z_i\geq \tau) \geq  \frac {c_Z}2\Big]\geq 1 - 2 e^{-nc_Z^2/2}\label{second-contrl-of-bias}.
\end{align}
Combining \eqref{first-contrl-of-bias} and \eqref{second-contrl-of-bias}, we  obtain the desired result.

$\hfill\square$

\paragraph{Proof of Theorem~\ref{thm-oracle-estimation}.}
Using the triangle inequality, we have that 
\begin{align*}
  \norm{f_{b^\star} - f_{\hat \beta}}^2_{n} \leq (\norm{f_{b^\star} - f^\star}_{n}  + \norm{f^\star - f_{\hat \beta}}_{n})^2 \leq 2(\norm{f_{b^\star} - f^\star}^2_{n} + \norm{f^\star - f_{\hat \beta}}^2_{n}).
\end{align*}
Inequality~\eqref{eqn:self_concord_2} in Lemma~\ref{lemma:self-concordance} yields
\begin{align*}
\norm{f^\star - f_{\hat \beta} }^2_{n}\leq \frac{\norm{f^\star-f_{\hat \beta} }^2_\infty}{ {\psi(-\norm{f^\star-f_{\hat \beta} }_\infty)}} {KL_n(f^\star,f_{\hat \beta})} \leq (f^\star_\infty + R + 2) {KL_n(f^\star,f_{\hat \beta})},
\end{align*}
where we use inequality~\eqref{eq:ineq_g}.
The construction of the approximation $f_{b^\star}$ of $f^\star$ gives $|\mathcal A(b^\star)| = K^\star$, so an application of Theorem~\ref{thm-faster-in-approx} to $b^\star$ combined with inequality~\eqref{eqn:self_concord_2} in Lemma~\ref{lemma:self-concordance} ensures that with a probability greater than $1 - 28.55 e^{-c} - e^{-ns^{(0)}(\tau)^2/8e^{2f^\star_\infty}} - 3\varepsilon$,
\begin{align*}
\label{oracle-for-b-star}
KL_n(&f^\star,f_{\hat \beta}) \leq 3KL_n(f^\star,f_{b^\star})+ \frac{1024(f^\star_\infty + R^\star + 2)K^\star\max_{1 \leq j \leq p}\norm{(\omega_{j,\bullet})_{\mathcal{A}_j(b^\star)}}^2_\infty}{{{\kappa}^2_{\tau}\big(\mathcal{A}({b^\star})\big) - \Xi_{\tau}\big(\mathcal{A}({b^\star})\big)}}\\
&\leq 3\norm{f^\star - f_{b^\star}}_{n}^2 \frac{ \psi(f_\infty^\star + R^\star +2 )}{(f_\infty^\star + R^\star+2)^2} + \frac{1024(f^\star_\infty + R^\star + 2)K^\star\max_{1 \leq j \leq p}\norm{(\omega_{j,\bullet})_{\mathcal{A}_j(b^\star)}}^2_\infty}{{{\kappa}^2_{\tau}\big(\mathcal{A}({b^\star})\big) - \Xi_{\tau}\big(\mathcal{A}({b^\star})\big)}},
\end{align*}
where we used the fact that $u\mapsto \psi(u)/u^2$ is increasing.
Therefore, with a probability greater than $1 - 28.55 e^{-c} - e^{-ns^{(0)}(\tau)^2/8e^{2f^\star_\infty}} - 3\varepsilon$, the following holds:
\begin{align*}
\norm{f_{b^\star} - f_{\hat \beta}}^2_{n}
&\leq2 \norm{f_{b^\star} - f^\star}^2_{n} \Big(1 + 3\frac{ \psi(f_\infty^\star + R +2 )}{f_\infty^\star + R+2}\Big)\\
&\qquad + \frac{2048(f^\star_\infty + R + 2)^2K^\star\max_{1 \leq j \leq p}\norm{(\omega_{j,\bullet})_{\mathcal{A}_j(b^\star)}}^2_\infty}{{{\kappa}^2_{\tau}\big(\mathcal{A}({b^\star})\big) - \Xi_{\tau}\big(\mathcal{A}({b^\star})\big)}}.
\end{align*}
By Lemma~\ref{control-of-bias}, we obtain
\begin{equation*}
	\norm{f_{b^\star} - f_{\hat \beta}  }^2_{n} \leq {\bf{I}} + {\bf{II}}
\end{equation*}
with a probability larger than $1 - 28.55 e^{-c} - e^{-ns^{(0)}(\tau)^2/8e^{2f^\star_\infty}} - 3\varepsilon - 2 e^{-nc_Z^2/2}$.
Now using the definitions of $\norm{\cdot}_n$ and $\kappa_{\tau}$ in~\eqref{compatibility-condition}, we have 
\begin{align*}
	\norm{f_{b^\star} - f_{\hat \beta} }^2_{n} = (b^\star - \hat\beta)^\top \widehat\Sigma_n(f^\star, \tau)(b^\star - \hat\beta)
	 \geq \kappa_\tau^2\big(\mathcal{A}(b^\star)\big) \norm{(b^\star - \hat\beta)_{\mathcal{A}(b^\star)}}_2^2.
\end{align*}
We therefore have that
\begin{align*}
	\norm{(\hat\beta - b^\star)_{\mathcal{A}(b^\star)}}_1 \leq \frac{\sqrt{{K^\star(\bf{I}} + {\bf{II})}}}{\kappa_\tau\big(\mathcal{A}(b^\star)\big)},
\end{align*}
with a probability larger than $1 - 28.55 e^{-c} - e^{-ns^{(0)}(\tau)^2/8e^{2f^\star_\infty}} - 3\varepsilon - 2 e^{-nc_Z^2/2}$.

$\hfill\square$

\end{appendices}

\bibliography{biblio}
\bibliographystyle{plainnat}{}
\end{document}